\documentclass{article}
\usepackage[utf8]{inputenc}

\usepackage[margin=1in]{geometry}

\usepackage{microtype}
\usepackage{graphicx}
\usepackage{subfigure}
\usepackage{booktabs}
\usepackage{microtype}
\usepackage{float}
\usepackage[colorlinks,allcolors=purple]{hyperref}
\usepackage{algorithm}
\usepackage{amsmath}
\usepackage[table,xcdraw,dvipsnames]{xcolor}
\usepackage{stmaryrd}
\usepackage{amssymb}
\usepackage{bbm}
\usepackage[natbibapa]{apacite}
\usepackage{alltt}
\usepackage{authblk}
\usepackage[capitalise,nameinlink]{cleveref}
\usepackage{svg}
\usepackage{tikz}
\usetikzlibrary{trees}
\usepackage[most]{tcolorbox}
\usepackage{soul}
\usepackage{enumitem}

\usepackage{fancyhdr}

\usepackage{listings}
\definecolor{githubblue}{RGB}{49,46,138}
\definecolor{schemegreen}{RGB}{15,137,15}

\lstdefinelanguage{Scheme}{
  morekeywords=[1]{define, define-syntax, define-macro, lambda, define-stream, stream-lambda},
  morekeywords=[2]{begin, call-with-current-continuation, call/cc,
    call-with-input-file, call-with-output-file, case, cond, condition,
    do, else, for-each, if,
    let*, let, let-syntax, letrec, letrec-syntax,
    let-values, let*-values,
    and, or, not, delay, force,
    quasiquote, quote, unquote, unquote-splicing,
    map, fold, syntax, syntax-rules, eval, environment, query },
  morekeywords=[3]{import, export},
  alsodigit=!\$\%&*+-./:<=>?@^_~,
  sensitive=true,
  morecomment=[l]{;},
  morecomment=[s]{\#|}{|\#},
  morestring=[b]",
  basicstyle=\small\ttfamily,
  keywordstyle=\bf\ttfamily\color[rgb]{0,.3,.7},
  commentstyle={\color[rgb]{0.24, 0.51, 0.51}},
  stringstyle={\color[rgb]{0.75, 0.49, 0.07}},
  upquote=true,
  breaklines=true,
  breakatwhitespace=true,
  literate=*{`}{{`}}{1},
  showstringspaces=false
}
\lstdefinestyle{churchstyle}{
  backgroundcolor=\color{white},   commentstyle=\color{gray},
  keywordstyle=\color{githubblue},
  numberstyle=\color{black}\tiny,
  stringstyle=\color{red},
  basicstyle=\ttfamily\color{githubblue},
  breakatwhitespace=false,         
  breaklines=true,                 
  captionpos=b,                    
  keepspaces=true,                 
  numbers=none,                    
  numbersep=5pt,                  
  showspaces=false,                
  showstringspaces=false,
  showtabs=false,                  
  tabsize=2,
  literate=*{\{}{{\textcolor{NavyBlue}{\{}}}{1}
        {\}}{{\textcolor{black}{\}}}}{1}
        {[}{{\textcolor{black}{[}}}{1}
        {]}{{\textcolor{black}{]}}}{1}
        {(}{{\textcolor{black}{(}}}{1}
        {)}{{\textcolor{black}{)}}}{1}%
}
\usepackage[T1]{fontenc}
\usepackage{inconsolata}
\definecolor{CodeBackgroundLight}{rgb}{0.975,0.975,0.975}
\definecolor{CodeBackgroundDark}{rgb}{0.025,0.025,0.025}

\usepackage[]{caption}
\usepackage{newfloat}
\newenvironment{code}{\captionsetup{type=listing,skip=0pt}}{}
\DeclareFloatingEnvironment[name=Code Block]{listing}
\crefname{listing}{Code Block}{Code Block}
\Crefname{listing}{Code Block}{Code Block}
\usepackage[framemethod=tikz]{mdframed}
\mdfdefinestyle{dialogBoxStyle}{%
  outerlinewidth=1.0mm,
  leftline=true,
  rightline=false,
  topline=false,
  bottomline=false,
}
\definecolor{colorCondition}{HTML}{3688BF}
\definecolor{colorQuery}{HTML}{44C58D}
\definecolor{colorDefine}{HTML}{875291}
\definecolor{colorRedefine}{HTML}{F79256}
\newenvironment{Dialogue}[1][colorCondition]{
    \begin{mdframed}[style=dialogBoxStyle,linecolor=#1]
}{
    \end{mdframed}
}

\newcommand{\Condition}[1]{%
    \noindent{\color{colorCondition}\textbf{Condition:}} \textit{#1}
}

\newcommand{\Define}[1]{%
    \noindent{\color{colorDefine}\textbf{Define:}} \textit{#1}
}
\newcommand{\Redefine}[1]{%
    \noindent{\color{colorRedefine}\textbf{Redefine:}} \textit{#1}
}

\usepackage[object=vectorian]{pgfornament}
\newcommand{\ornament}{
    \begin{center}
        \pgfornament[width = 2em, color = black]{80}
    \end{center}
}

\fancypagestyle{firstpagefooter}{%
  \fancyhf{}

  \fancyfoot[L]{\footnotesize Code for the examples in this paper is available at: \href{https://github.com/gabegrand/world-models}{\texttt{github.com/gabegrand/world-models}}.\\
  Correspondence: co-primary authors (\texttt{zyzzyva@mit.edu, gg@mit.edu}); co-supervisors (\texttt{jda@mit.edu, jbt@mit.edu}).
  }
}

\title{\vspace{-.5in} \textbf{From Word Models to World Models:}\\Translating from Natural Language to the\\Probabilistic Language of Thought}

\date{}

\author{
    {Lionel Wong}$^{1\star}$, 
    {Gabriel Grand}$^{1 \star}$, 
    {Alexander K. Lew}$^{1}$, 
    {Noah D. Goodman}$^{2}$,
    {Vikash K. Mansinghka}$^{1}$,
    {Jacob Andreas}$^{1}$, 
    {Joshua B. Tenenbaum}$^{1}$ \\
    \small{$^{\star}$\textit{Equal contribution.}} \\
    \vspace{1em}

    \large{
    $^1$MIT,
    $^2$Stanford} \\
}

\begin{document}

\maketitle

\begin{abstract}
How does language inform our downstream thinking? In particular, how do humans make meaning from language---and how can we leverage a theory of linguistic meaning to build machines that think in more human-like ways?
In this paper, we propose \textit{rational meaning construction}, a computational framework for language-informed thinking that combines neural models of language with probabilistic models for rational inference. We frame linguistic meaning as a context-sensitive mapping from natural language into a \textit{probabilistic language of thought} (PLoT)---a general-purpose symbolic substrate for probabilistic, generative world modeling. Our architecture integrates two powerful computational tools that have not previously come together: we model thinking with \textit{probabilistic programs}, an expressive representation for flexible commonsense reasoning; and we model meaning construction with \textit{large language models} (LLMs), which support broad-coverage translation from natural language utterances to code expressions in a probabilistic programming language.
We illustrate our framework in action through examples covering four core domains from  cognitive science: probabilistic reasoning, logical and relational reasoning, visual and physical reasoning, and social reasoning about agents and their plans. In each, we show that LLMs can generate context-sensitive translations that capture pragmatically-appropriate linguistic meanings, while Bayesian inference with the generated programs supports coherent and robust commonsense reasoning. We extend our framework to integrate cognitively-motivated symbolic modules (physics simulators, graphics engines, and goal-directed planning algorithms) to provide a unified commonsense thinking interface from language. Finally, we explore how language can drive the construction of world models themselves.
We hope this work will help to situate contemporary developments in LLMs within a broader cognitive picture of human language and intelligence, providing a roadmap towards AI systems that synthesize the insights of both modern and classical computational perspectives.
\end{abstract}
\vspace{1em}

\thispagestyle{firstpagefooter}

\section{Introduction}\label{sec:introduction}
Language expresses the vast internal landscape of our thoughts. We use language to convey what we believe, what we are uncertain about, and what we do not know. We talk about what we see in the world around us, and what we imagine in real or wholly hypothetical futures. We discuss what we want and what we plan to do, and dissect what others want and what we think they will do. We build and pass on new bodies of knowledge in language---we ask questions and offer explanations, give commands and instructions, and propose and refute theories. Some of these ideas can be expressed in part through other means. But language stands apart for its flexibility and breadth, and its seeming proximity to our thoughts.

What \textit{is} language? How does language get its meaning, and when should we say that a person or machine knows, understands, and can use it? What is the relationship between language and the rest of general cognition---what allows language to inform and support so much of thought? This paper focuses on these questions as they relate to \textit{human} language and thought, in computational terms. What integrated cognitive theory can model how language relates to the other core systems of human cognition? If we seek to build AI systems that emulate how humans talk and think, what architecture can integrate language robustly into systems that support the full scope of our thought? 
\clearpage
Theories of cognition have long considered human language and thinking to be deeply related, but fundamentally distinct. \textit{Thinking}, in many traditional cognitive theories, revolves around goal-directed world modeling, inference, and decision making---constructing mental models of the world that reflect prior beliefs, can be updated from new observations, and support rational prediction and decision making toward's one's goals \citep{morgan1999learning,nersessian2010mental,johnson1980mental,johnson1989mental,gentner2014mental,craik1967nature, lake2017building}. \textit{Language}, in contrast, centers around communicating these thoughts to others, and receiving their thoughts in turn. In most linguistic theories, human languages are mappings between the internal representations of thought and an externalizable symbol system, which might be phonemes, signs, or glyphs \citep{frege_uber_1892,kratzer1998semantics,lewis1976general}. To produce language is to map thoughts into these external symbols, and to understand language is to transduce from these external symbols back into the representations of thought. 

The theoretical distinction between language and thought rests on multiple intersecting lines of evidence. Prior to learning language, infants are born equipped with a powerful toolkit for modeling and thinking about the world, including an understanding of physical objects and events, and the goals and actions of agents \citep{spelke2007core,spelke2022babies}, and  general abilities for learning statistics and structure \citep{saffran2001acquisition,xu2021bayesian}. Building on these foundations, children acquire language from relatively sparse input data, rapidly generalizing beyond the utterances they hear to produce and understand entirely new ones \citep{bloom2002children,landauer1997solution,pinker1998words,gleitman2005hard,gleitman1990structural,smith2008infants}; they then use language to acquire new concepts they would not get merely from direct experience \citep{carey09origin,wellman1992cognitive,gopnik1996scientist}. Language and thought also appear to operate in distinct but interacting brain systems: neuroimaging and neurological studies reveal a  ``language network'' specialized for processing sentences, functionally and anatomically separate from but closely connected to brain networks supporting other aspects of general cognition \citep{fedorenko_language_2016,mahowald2023dissociating}.

`These empirical findings have shaped decades of computational models in cognitive science and AI. To model the expressiveness of human cognition, an influential computational paradigm suggests that humans 
compose and execute mental programs in an internal \textit{language of thought} \citep{fodor_language_1975}, a structured symbolic substrate for representing conceptual knowledge that provides a general interface to algorithms for problem solving and reasoning. These symbolic systems are not merely logic engines; they support our probabilistic inferences, and rich intuitive simulations \citep{russell_norvig_2021,goodman2014concepts,oaksford2007bayesian}.
This paradigm underlies many of the success stories in cognitive science and related applications in AI. It has influenced models that capture how people draw causal and explanatory inferences about facts and observations \citep{pearl2000models,pearl1988probabilistic}, learn and generalize concepts from few examples \citep{lake2017building}; plan actions over long time horizons and under complex conditions \citep{russell_norvig_2021,kaelbling2013integrated}; imagine and predict the physical world \citep{battaglia2013simulation,ullman2017mind}; and reason about other agents with their own beliefs and goals \citep{baker2011bayesian}. Within linguistics and natural language processing, in turn, this paradigm underlies \textit{semantic parsing} systems designed to map from human language into symbolic computational representations. It has yielded AI systems that could follow instructions \citep{tellex2011understanding} and answer natural language queries with respect to structured knowledge representations \citep{steedman2011combinatory,klein2003accurate,wong2007learning,liang2016learning}; as well as cognitive models that capture how human children learn the grammar and meaning of expressions in their native language \citep{perfors2011learnability,chater2006probabilistic,piantadosi2012bootstrapping,frank2009using,goldwater2009bayesian,abend2017bootstrapping,gauthier2018word}. 

Despite this progress, however, modular and symbolic models of language and thought have been dogged by persistent critiques of their scalability and scope. Cognitive and AI researchers over the years have carved off specific domains of world knowledge, constructing bespoke representations to model them without a general account of whether they would generalize to all of human knowledge, or how they could be scalably learned. Semantic parsing systems inherited these critiques, and faced additional challenges in implementing the mapping from sentences into symbolic representations. These mapping functions were either hand-engineered or learned from strong supervision on specific domains of language, limiting them to brittle, imperfect models of the breadth and complexity of real human discourse. 
\ornament
\clearpage

In just the last few years, a serious challenge has emerged to the traditional view of language and thought as distinct but interacting components of the mind, each modeled using structured representations. \textit{Large language models} (LLMs) use a new generation of attention-based deep neural networks to learn the probabilistic distributions of words from vast datasets of human language, generally training on orders of magnitude more data than a human encounters in their lifetime \citep{vaswani2017attention,openai2023gpt4,brown_language_2020,rae2021scaling,bommasani2021opportunities}. The underlying computational objective that drives these models is not itself new. LLMs follow in the tradition of distributional approaches to discovering structure in language \citep{harris_distributional_1954, firth1957synopsis,osgood1952nature}, which seek to extract representations of meaning from statistical patterns in how words are used in context \citep{sahlgren2008distributional,Griffiths2007TopicsIS,mikolov2013word2vec,dumais2004latent}. What is new, however, is the scale and scope of today's distributional vision, which has expanded in stages. A first generation of LLMs, trained specifically to predict words in context, produced such fluent language that they challenged traditional symbolic approaches to modeling language \citep{radford_language_2019,peters1802deep,devlin2018bert}. Their qualitative success, as well as internal representational probes, suggested that linguistic structures sufficient for grammatically coherent language could be learned entirely from modeling the statistics of words \citep{piantadosi2023modern,tenney2019bert}. By scaling to even larger datasets and neural networks, LLMs appeared to learn not only the structure of language, but capacities for some kinds of thinking; they could learn new words in context, and extract patterns in language from a few examples that they could generalize locally to similar cases \citep{brown_language_2020}. The most recent LLMs have been trained not only to model the statistics of language but explicitly to reason, with targeted  supervision on instruction following, writing code, and other forms of human dialog and feedback in conversational contexts \citep{ouyang2022instructgpt,openai2023gpt4,openai2023chatgpt,chen2021evaluating}.  They produce such fluent language on a wide variety of tasks that many have begun to ask whether merely more training of this sort, with increasing scale, could
learn representations sufficient for general intelligence \citep{bubeck2023sparks}. Proponents of the most extreme ``scaling hypothesis'' have argued that because language is used to express so much of human thought, a sufficiently large and performant predictive language model would effectively \textit{have} to construct an internal model of all of cognition \citep{scalinghypothesis2022}.

This theoretical vision has sparked both excitement and controversy, but proponents and critics agree that it raises its own questions about its long-term scalability---most significantly, what will be required to close the outstanding gaps between today's LLMs and general cognitive models that reason systematically and consistently about the language they receive or produce. Current LLMs can produce impressive results on a set of linguistic inputs and then fail completely on others that make trivial alterations to the same underlying domain \citep{ullman2023large}; they mix confident answers to complex questions with equally confident, hallucinated language that does not reflect a consistent, calibrated notion of truth or belief \citep{openai2023gpt4,bubeck2023sparks}. These issues make it difficult to evaluate whether LLMs have acquired cognitive capacities such as social reasoning and theory of mind \citep{ullman2023large}, or to compare different kinds of world modeling and planning tasks \citep{valmeekam2023planning}. One approach to solving these problems is through additional data. Perhaps fully robust, systematic reasoning will finally emerge if models are trained on still more language, or supervised more explicitly on data from complex reasoning tasks. This scaling route raises practical questions about whether it will be possible to acquire enough data to train such a model, as well as theoretical questions whether more data and more parameters alone \textit{will} in fact yield robust systems for thought.  Another strategy in recent work seeks to build more robust cognitive capacities by augmenting LLMs with various external tools for structured representation and symbolic reasoning,  
such as calculators \citep{cobbe2021training}, logic engines \citep{weirDynamicGenerationInterpretable2022}, databases \citep{borgeaudImprovingLanguageModels2022, thoppilanLaMDALanguageModels2022, alonNeuroSymbolicLanguageModeling2022, izacardFewshotLearningRetrieval2022}, physics simulators \citep{liu2022mind}, planners \citep{liu2023llm+}, and APIs for executing arbitrary code \citep{karpasMRKLSystemsModular2022, schick2023toolformer, openai2023gpt4}. But these new hybrid approaches resurrect many of the same long-term scalablity challenges that confronted earlier semantic parsing and knowledge representation systems, by designing a menagerie of bespoke representations and tools without a broader account of how they will scale towards general models of language and thought.

In this paper, we consider a different approach to integrating the strengths of modern language models and classic symbolic architectures, one that draws on but also runs counter to recent trends in AI, in a sense flipping these scaling questions on their head.  Instead of trying to turn models trained to predict language into models that might genuinely think---filling each gap in reasoning we discover through yet more data, new kinds of language training or linguistic prompting tricks, or by plugging in yet another external tool---we ask: what are the prospects for a unifying computational framework guided by the study of thought and language in the human mind and brain, as well as what we have learned from multiple eras of AI? Can we build intelligent architectures that use, learn and understand language as people do, informed by neuroscience constraints and developmental trajectories?  That is, can we build models in which language is learned efficiently within one relatively small, modular computational system, which interfaces generally with other systems dedicated to robust world modeling and reasoning? What architecture lets language build on pre-existing capacities for symbolic world modeling and inference, while also allowing linguistic meanings and world knowledge to scaffold and bootstrap each other, as a learner's experiences and competences grow?

\ornament

\begin{figure}[t!]
    \centering
    \includegraphics[width=\textwidth]{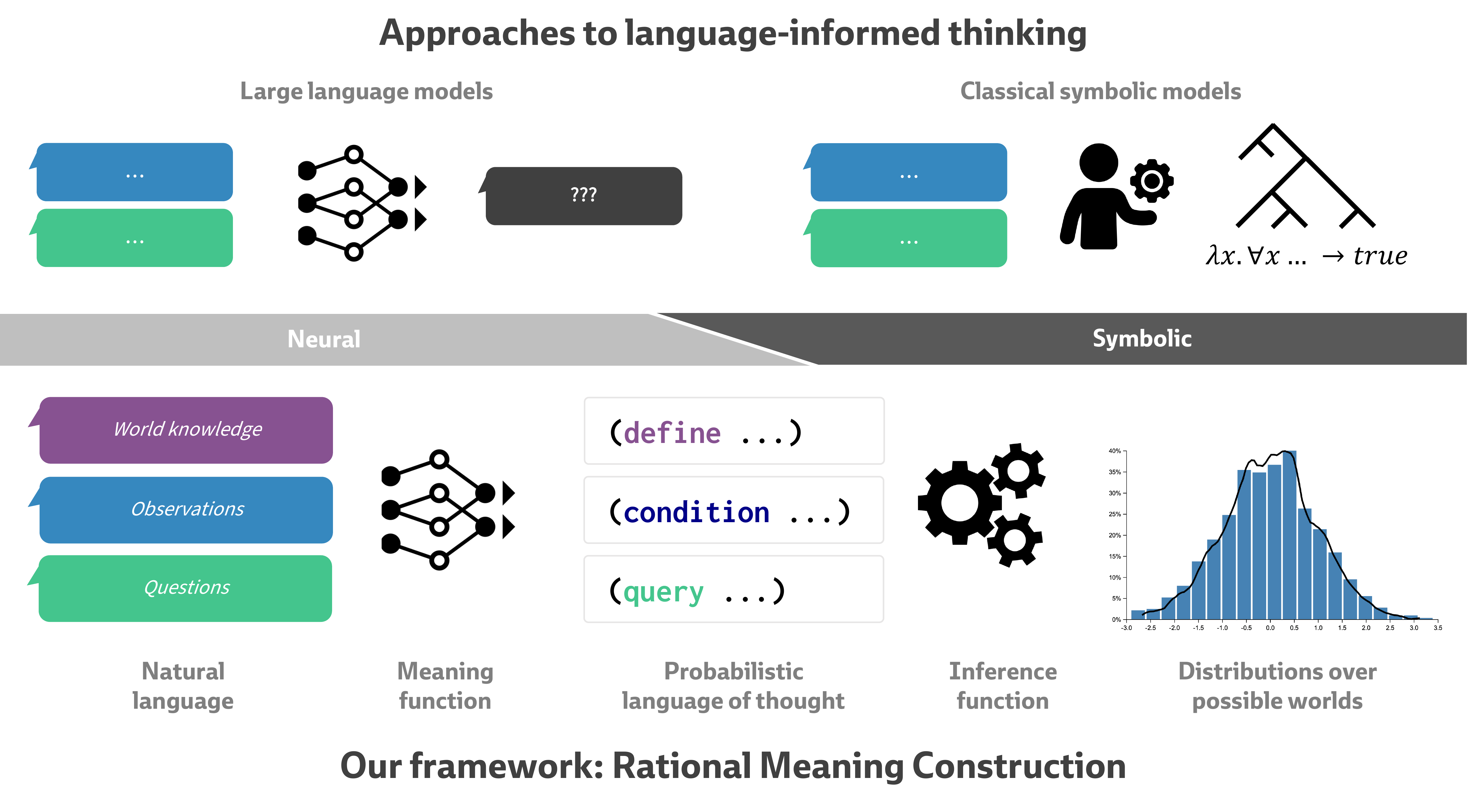}
    \caption{Human language understanding supports flexible inferences in a process we term \textit{language-informed thinking}. Computational approaches to language-informed thinking sit on a neurosymbolic continuum: On one side, classical symbolic models (top right) yield systematic, structured inferences, but are typically limited to narrow linguistic domains and often require hand-engineering. On the other side, large language models (LLMs; top left) achieve remarkable facility with open-domain natural language, but struggle to ground reasoning in a consistent world state that supports coherent inferences, predictions and plans. Our \textit{rational meaning construction} framework decomposes language-informed thinking into two modules: (1) A \textit{meaning function} translates natural language into probabilistic programming language (PPL) statements that represent linguistic meaning with respect to a symbolic world model. (2) An \textit{inference function} computes probabilities over the space of possible worlds consistent with and conditioned on information in the language. In the rest of this paper, we illustrate how our framework can combine the strengths of LLMs and PPLs, affording both broad coverage of natural language and a principled treatment of reasoning about uncertain events, outcomes, and scenarios.}
    \label{fig:intro-architecture}
\end{figure}

This paper attempts to show what such a model might look like---and how it can build theoretically and practically on the insights from both classical paradigms for language and thought, \textit{and} the recent successes of statistical learning made by large language models. We propose a framework for intelligent computational architectures that reason about and learn from language, but we begin with a proposal for what it means to \textit{think}. As in the traditional cognitive view, thinking at its core is constructing general-purpose representations for  modeling the entities and events in the world, sufficient to support rational, coherent inferences under uncertainty and planning actions that achieve our goals. We then consider how language relates to this architecture to support \textit{language-informed thinking}---how language sets up world modeling and inference, to guide, constrain, and drive our downstream thought, and grow new thinking capacities.

Our proposal, which we call \textbf{rational meaning construction}, rests on the integration of two computational components, each which we suggest can be instantiated using modern computational tools---a \textit{probabilistic language of thought} for constructing structured models of arbitrary situations, which supports supports coherent belief updating and inferences over them; and a general mechanism for taking natural language and \textit{constructing meaning} from it, represented as distributions over expressions in this language of thought (\cref{fig:intro-architecture}). We propose that \textit{probabilistic programs} can formally instantiate the first component. They offer a structured representation for expressing novel situations and arbitrary problems with respect to a meaningful model over possible world states, a coherent notion of conditional belief updating, and a systematic framework for inferences with respect to queries and goals. We propose, in turn, that \textit{meaning construction} can be modeled as \textit{translation} from utterances in language to expressions in a general probabilistic programming language. Theoretical and empirical results have long suggested that human languages implement locally compositional, efficiently learnable mappings between symbolic representations of thought and external symbol systems. We therefore propose that code-trained large language models can be viewed as in-principle implementations of broad, context-sensitive, and \textit{resource-rational meaning functions}, in that they can be used to efficiently infer distributions between language and programs from stored, prior patterns in the background distribution of language and code. By integrating these two components, we propose that this paradigm suggests a general framework by which language can meaningfully relate to many fundamental aspects of cognition, modeling how we might condition on language to systematically update our beliefs, pose new questions and goals in language, and convey structured background information or even define new relevant concepts about a situation or about the world.

In \cref{sec:overview}, we give an overview of this framework, describing the overall structure and more detailed rationale behind the computational components we build on in the remainder of this paper. We then describe a concrete but minimal implementation of this framework using contemporary probabilistic programming and language modeling tools, intended to demonstrate the basic computational components of this approach and elucidate the scope and scalability of the broader proposal.

\begin{figure}[ptbh!]
   \centering
   \includegraphics[width=\textwidth]{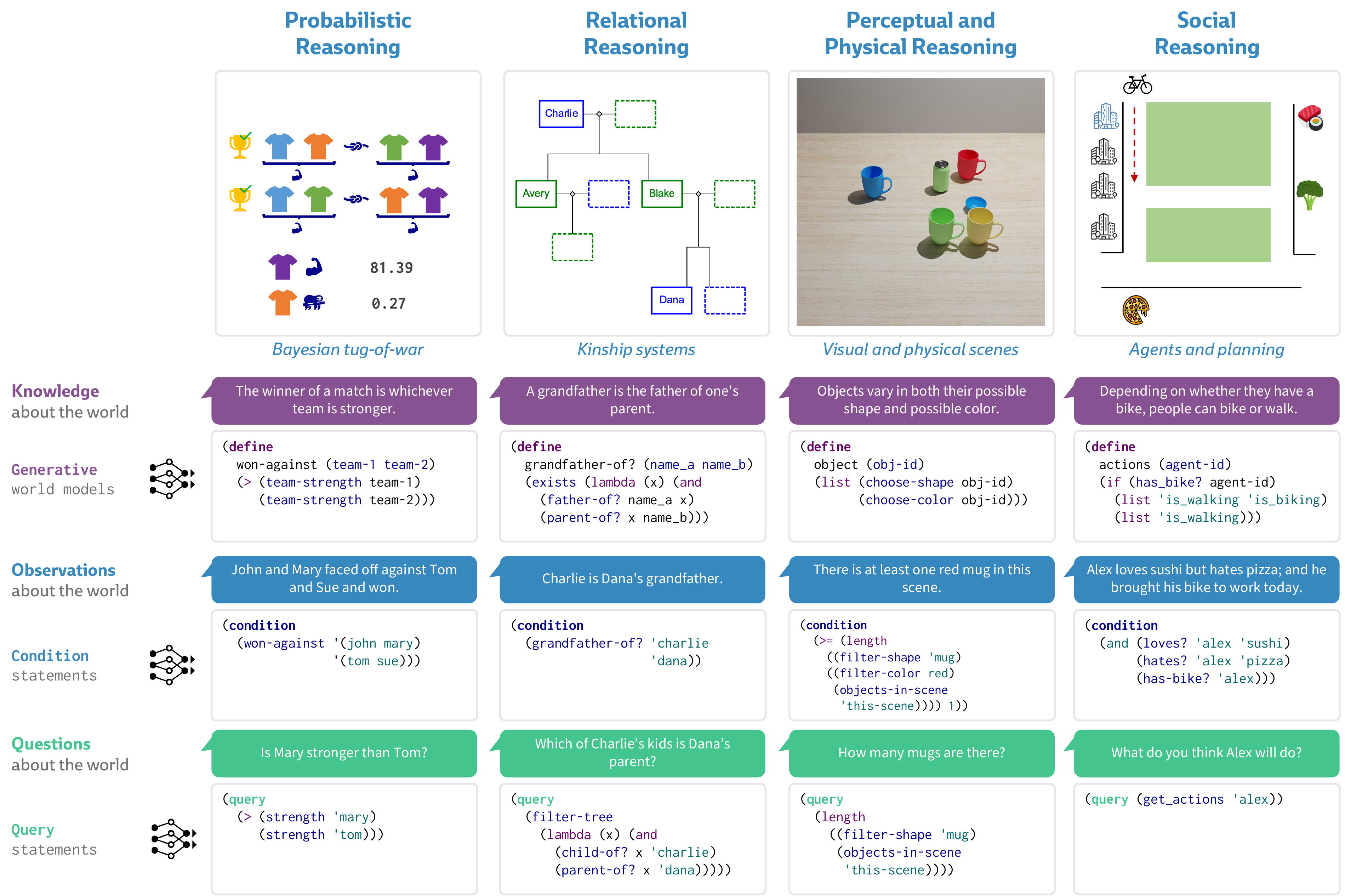}
    \caption{Understanding language in four domains of reasoning that form the core of this paper.  \textbf{Probabilistic reasoning} requires integrating sparse evidence to predict the outcomes of uncertain events, like the winners of tug-of-war matches. \textbf{Relational reasoning} involves maintaining and updating coherent beliefs about structured domains, like family trees, based on relational information. \textbf{Perceptual and physical reasoning} links language to our sensory and intuitive physical knowledge of objects in the external world, such as kitchen items on a tabletop. \textbf{Social reasoning} involves reasoning about the minds of other intelligent agents, such as how their goals, preferences, and circumstances shape their actions as they navigate in the world. Across all the domains, we present a unified framework that translates language into code in a probabilistic programming language to facilitate human-like reasoning.}
   \label{fig:splash}
\end{figure}

Given this general paradigm, we first illustrate the potential breadth of this approach for integrating \textbf{meaning construction and reasoning}, showing how it might address a core set of computational and cognitive domains that we communicate about in language (\cref{fig:splash}). Each of these examples uses minimal pedagogical examples intended to suggest how this approach integrates language with important bodies of work from computational cognitive science and artificial intelligence.  We first show how this framework can condition on language in order to describe and reason about \textit{uncertain situations} with respect to an ongoing discourse (\cref{sec:models-probabilistic-reasoning}), then show how this approach can be extended to reason about \textit{relational systems} (\cref{sec:models-relational-reasoning}), \textit{physical and perceptual scenes} (\cref{sec:models-reference-and-grounding}), and social situations involving agents with \textit{goals and plans} (\cref{sec:models-agents-and-planning}). 

We then turn to how this approach might begin to address core scalability challenges that confront traditional approaches to modeling thinking as symbol processing, whether logical or probabilistic.  In \cref{sec:growing-and-constructing}, we show how language can support growing knowledge autonomously, without hand engineering, by using the rational meaning construction framework to construct a broad range of \textbf{new concepts} in existing models and even whole \textbf{new world models}, which in turn support coherent downstream reasoning.

Ultimately, this paper is a prospective one, and the examples presented here are intended to convey a sufficiently concrete proposal to suggest avenues for future work. In \cref{sec:open-questions}, we outline what we see as some of the most significant open questions and future directions raised by this framework. These include theoretical questions that relate our approach to classical models of language, open cognitive directions for extending this approach to model language acquisition and production, and important engineering directions necessary for scaling inference, robust translation, and learning under this general paradigm. Finally, in \cref{sec:conclusion}, we conclude by looking ahead to the longer-term implications of this proposal for modeling intelligent systems that use, understand, and think about language as we do.

\clearpage
\section{Overview of the key ideas}\label{sec:overview}
The central goal of this paper is to propose a new computational framework, \textit{rational meaning construction}, which relates language to thought. This framework licenses a concrete class of computational architectures for building intelligent systems that use language, which we propose can be implemented using modern AI tools. In this section, we briefly overview the key ideas that form the basis of this proposal. We draw on three observations from a rational, probabilistic perspective on biological intelligence and human language:

\paragraph{A rational perspective on intelligent thought.} Biological intelligence encompasses many computational capacities. The foundational notion of thought we focus on here centers on \textit{rational inference and decision making} in service of one's goals \citep{anderson1990adaptive,chater1999ten}. Under this perspective, thought comprises systems for modeling the world. These internal world models allow us to infer the particulars of a situation from whatever information is at hand, evaluate alternative world states and imagine possible future ones, and decide on actions that might bring one towards valuable future states in the world. Following extensive work in computational cognitive science, we view the world models that support biological intelligence as structured and probabilistic \citep{griffiths2010probabilistic,lake2017building,goodman2014concepts}, designed to integrate the noisy evidence an agent receives into causal, explanatory models that allow them to maintain coherent beliefs about the world and generalizably infer consistent, useful predictions and plans. This basic, underlying view of intelligent thought draws on empirical evidence from essentially every species with a brain, from bees \citep{biernaskie2009bumblebees,wang2002human}, to zebrafish \cite{johnson2020probabilistic,bolton2019elements}, mice \citep{english2023bayesian}, birds \citep{isomura2019bayesian}, and primates \citep{khalvati2021bayesian}. Informally, a rational view of thought can be summarized as the ability to \textit{solve useful problems given our internal models of the world}, ranging from navigation and foraging to physical prediction and social reasoning. Against this overarching picture of thought, human intelligence further stands out for its flexibility and expressiveness. We \textit{invent} our own problems along with new approaches to solving them, rather than sticking to a limited set of largely innate goals and strategies \citep{tomasello2022evolution}. A few other species, non-human primates, dolphins, and some birds, are creative problem-solvers and problem-creators, but none come close to the range of goals humans can adopt \citep{chu2023praise}. Uniquely in the natural world, humans think about and come to understand problems far beyond the narrow range necessary for our immediate survival, considering goals and questions that draw on abstract, culturally constructed, and even entirely hypothetical systems for modeling and conceptualizing the world \citep{dennett2017bacteria}. 

\paragraph{A rational perspective on language.} As with thought, language also encompasses many systems and capacities. This paper focuses on the class of problems we refer to as \textit{language-informed thinking}, the general means by which language informs the inferences and decisions of an intelligent agent. We take a broadly rational perspective on language---we consider language to be a system of \textit{goal-directed actions for externalizing and communicating thoughts} to other intelligent beings \citep{chater2006probabilistic,goodman2016pragmatic,gibson2014language}. In this context, we frame the problem of deriving \textit{meaning} as inferring the mappings between a language’s system of external communicative signals into the representations of rational thought. It is worth highlighting that thought does not require language and is distinct from language in the human brain \citep{mahowald2023dissociating}. Non-human species, and pre-verbal infants \citep{spelke2022babies}, are clearly capable of modeling the world towards their inferences and goals without language. But for humans, language clearly plays a profound role in determining the problems we think about, and how we think about them. Our natural languages allow us to communicate an extraordinarily broad range of our thoughts about the problems we pose and solve, including our abstract and general world knowledge, our specific beliefs about a situation, the particular questions or goals we have or want to pose to others, and our approaches to reasoning about them. 

\paragraph{A resource-rational perspective on language and thought.} Finally, our integrated computational approach to language and thought builds on extensive evidence that humans are \textit{resource-rational} thinkers---under finite constraints of time and memory, we rationally allocate computational resources in order to make useful inferences and plans \citep{lieder2019resource,gershman2015computational}. Resource rational agents \textit{amortize} computational effort across prior experience and problems, storing and reusing prior computation towards similar new problems that we encounter in the future \citep{gershman2014amortized,le2017inference}. Certain domains of inferences share more structure than others, and evidence suggests that we therefore heavily amortize them. Prior work, for instance, suggests that computations involved in basic perceptual activities \citep{fodor1983modularity,dasgupta2021memory,wilson2023why}, such as object recognition under common lighting conditions, are highly amortizable from reusable patterns in computation that are learnable and shared across a background distribution of perceptual instances. This view suggests why fast, bottom-up pattern recognition models have made great advances in modeling perception in recent years, while it has proved much more challenging to amortize the wide range of flexible inferences required for arbitrary problem solving. 

We propose an analogous resource-rational perspective on the kinds of computation implicated in language-informed thought. Under almost every theoretical and empirical account of linguistic structure and semantics, the mappings between language and meanings should be highly amortizable across the background distribution of language---there are structured, systematic, and learnable patterns in how units of language map onto units of thought. The idea that meaning construction should be highly amortizable follows from our view on language itself as an efficient communicative system. Extensive empirical evidence suggests that communicative pressures shape how language maps onto meanings at every level of linguistic structure, from  individual morphemes \citep{bybee1985morphology} to patterns in how common syntactic frames communicate meaning \citep{grimshaw1981form,gleitman1990structural}, and even reusable pragmatic implications present across common discourse situations \citep{white2020learning}. But while we take the view that a resource-rational agent should intelligently learn and reuse prior computation when possible, we do not view language-informed thinking, or thinking in general, as solely a matter of learning and interpolating over statistical patterns from prior experience. When we think, including when we think \textit{about} the meanings we recover from language---to update our beliefs, to follow instructions, or to answer questions posed in language---we must be able to flexibly model arbitrary situations and support capacities for general problem solving, including inference, planning, and simulation, under a wide range of new and unencountered circumstances.

The efficient learnability of human language also highlights that, in many senses, the computational relationship between language and thought in humans is almost the inverse of that in today’s LLMs. For humans, language could be characterized as an emergent property of \textit{thinking}. Infants can model the world and draw inferences well before they know language \citep{gopnik1996scientist,spelke2022babies}, and reliably acquire complete linguistic capabilities from exposure to relatively tiny amounts of language \citep{brown1973first}. Congenitally-Deaf humans born with \textit{no} language input spontaneously develop languages to communicate their thoughts, with the same basic hallmarks of mature natural languages \citep{goldin201226,pyers2010evidence,senghas2004children}. This paper seeks to understand and model the cognitive and computational structures underlying \textit{this} human scaling route to intelligence and language use — one that begins with robust capacities for thought, and scaffolds language efficiently on top of them to then offer a powerful tool for driving and constructing new thought.

\subsection[A rational meaning construction framework]{Our proposal: A framework for modeling rational meaning construction}
The perspective we offer above draws from theoretical and empirical work that precedes this paper. Our core contribution in this paper is to propose a new computational framework in light of these observations, that seeks to unify prior symbolic, probabilistic inference and statistical learning traditions and to take advantage of the clear computational advances made by modern LLMs as learned statistical models of language. We describe a framework for \textbf{rational meaning construction} in which linguistic meaning is formalized as a context-sensitive mapping from natural language to a distribution over expressions in a probabilistic language of thought (PLoT) for rational world modeling and inference. Under this framework, we then propose that large language models trained on language and code can be \textit{used to implement meaning functions in a resource-rational architecture} -- they can implement learned, broad-coverage mappings between language and code; and they can be understood as part of a human-like, resource-rational system that efficiently infers these mappings using stored patterns amortized from the prior joint distribution over language and code. This motivates the concrete architecture we propose and illustrate throughout the remainder of this paper, and its two main components for modeling thinking and modeling language relative to thinking---or how language informs thinking. 

\subsubsection{Modeling thinking} 
We propose implementing \textbf{thinking using probabilistic programs} as a general representational substrate for building world models and specifying rational inferences over them. This proposal builds on prior work in cognitive science and AI formalizing how a broad class of problems can be expressed as probabilistic programs \citep{chater2006probabilistic,goodman2014concepts}, following a generic \textit{inference query} motif \citep{goodman2012church} — a probabilistic program that combines a \textit{generative world model} that models abstract, causal beliefs about probable world states; specific evidence that an agent \textit{conditions} on; and a particular {query} being posed as the question or goal for thinking. Inference to solve a problem consists of formally computing or sampling from a probability distribution over answers to this question, specified by the world model and conditions. This computational proposal forms the backbone of the \textit{probabilistic language of thought} model of general human cognition \citep{goodman2014concepts}, and has been used empirically to model a wide range of human inferences, including those that draw on visual perception \citep{mansinghka2013approximate}, physical simulation \citep{battaglia2013simulation}, and social reasoning \citep{baker2011bayesian}. It is designed explicitly to formalize a central property of human thought — the capacity to expressively and flexibly pose problems involving entirely novel situations and goals, and to solve them relative to a computable representation of the world and internal belief.

\subsubsection{Modeling language relative to thought} 
Given this model for thought, we propose formalizing \textbf{rational meaning construction} as a broad-coverage, contextual translation function that maps language into a distribution over expressions in a probabilistic language of thought. This proposal builds most closely on and 
draws inspiration from efforts to articulate a \textit{probable world semantics} for natural language in prior work \citep{goodman2015probabilistic}, in order to express how language could compactly convey uncertain propositions and vague meanings with respect to a formal probabilistic generative model. It also builds on the longer history of symbolic semantic theories we overview in the introduction, including formal semantics theories that model language as mapping into formal propositions over possible worlds (eg. \cite{kratzer1998semantics,lewis1976general}), and semantic parsing systems (eg. \cite{klein2003accurate,liang2016learning,steedman2001syntactic,wong2007learning,abend2017bootstrapping}) that map language into formally executable program expressions.

Our goal, however, is to broaden and generalize these framings to suggest a general framework for modeling how language can interface with and inform such a broad swatch of human cognition. By positing that meaning is a general mapping between sentences and expressions in a probabilistic language of thought, we believe that a rational meaning construction approach can elaborate on and concretely model core desiderata of a coherent theory of linguistic meaning – modeling how meanings drive inferences about what is true and probable; formalizing how language can pose propositions and queries that are then evaluated with respect to an internal model over probable worlds; and relating meaning to the general computational systems for representing, thinking about, and receiving new information about the world from broader cognition.

This proposal suggests a wide class of possible architectures that map from language into probabilistic programs---in principle, any general mapping function that expresses a distribution over programs conditioned on sentences in context. Under this umbrella of possible implementations, we propose finally that \textbf{large language-to-code models} can be used to generally instantiate these meaning functions. Unlike prior semantic parsers or attempts to hand implement mappings between language and code, LLMs offer a concrete means of instantiating far more broad-coverage mappings between human sentences and meanings than have been previously possible. They are also context-sensitive, in that they can construct meanings for an utterance that condition both on the general distribution of language and thought and a local linguistic and thinking context. They can condition translation on a local discourse context, when prompted with prior utterances, and on a local problem under consideration, when prompted with existing code in a probabilistic program. 
 
By using LLMs to map between language and code, this proposal is also closely related to the recent lines of work we review in the introduction that seek to augment and connect LLMs with various structured and symbolic reasoning tools---both domain-specific reasoning engines like planners and physics engines (eg. \cite{liu2023llm+,liu2022mind}), and more general APIs for code execution (eg. \cite{openai2023plugins,schick2023toolformer,karpasMRKLSystemsModular2022}). As we demonstrate throughout this paper, however, we propose that the probabilistic language of thought can offer a cognitively-motivated, unifying symbolic substrate for interfacing between language and many core aspects associated with general cognition. It provides a general motif for structuring and constructing generative world models, which can nest calls to other domain-specific systems (such as planners and physics engines); and an overarching framework for modeling how diverse kinds of observations can be used to update these models and answer new queries, framed as Bayesian conditioning and inference. With respect to the more general landscape of large statistical language models, this proposal finally suggests one way to situate the strengths of LLMs into a more human-like, modular framework for language-informed thinking. Rather than look to statistical patterns to capture all of the ways we think, plan, and reason about language, this resource-rational approach seeks to ground distributional aspects of language into a framework that can leverage learned prior patterns when they are useful---while also modeling how language can construct and relate to coherent world models and algorithms for explicit, novel decision making and inference.

\subsubsection{Illustrating the architecture by example} 

This general architecture is best explained through concrete implemented examples, which we give in the next sections.  For each of the four domains of reasoning shown in \cref{fig:splash}, we work through a representative dialog between a speaker of English and our language-informed thinking computational architecture, which could stand in for how we model another human being's understanding and thinking about the speaker's language, or the ways we hope a human-like AI system would similarly respond.  

For pedagogical reasons, we have chosen to implement these examples using one particular probabilistic programming language and one particular language-to-code model. These particular tools are not necessarily the most performant or scalable AI solutions; nor the best accounts we have of the corresponding components of human architecture. Nevertheless, they are familiar and simple, and provide the most direct route we know to illustrate our ideas in ways others can also experiment with. To elaborate on these choices:

\begin{itemize}
 \item The probabilistic language of thought we use to express inference problems is \textit{Church} \citep{goodman2012church}, a Turing-universal probabilistic programming language constructed on top of the functional programming language Scheme. We have used the WebChurch dialect which implements several general inference procedures, but we have chosen the simplest and most general---and least efficient---approach based on rejection sampling: Inference is based on drawing samples from the prior over world states described by the generative model, and rejecting those that fail satisfy the constraints of any observation conditions.  The samples that remain constitute a posterior sample over possible worlds consistent with the observed information, sufficient to answer the queries under consideration in the language discourse. Other similarly functional PPLs such as WebPPL or Gen could have been chosen instead. In \cref{sec:open-questions}, we discuss future directions for extending and scaling inference beyond these simple illustrative implementations. 
 
 \item The language-to-code model we use to amortize meaning construction over programs is \texttt{Codex} model \citep{chen2021evaluating}, a GPT-3-based language model fine-tuned on source code, which provides pairings between natural language and code with comments, drawn from programs on GitHub and other sources. Since the release of Codex, many other language-to-code models have been developed, and more recent versions of GPT-based language models are now routinely trained on large amounts of source code; we believe these could be used to similar effect. In \cref{sec:open-questions}, we also discuss future directions for more cognitively plausible training and updating of neural models that amortize inference in joint distributions over natural language and probabilistic languages of thought.
\end{itemize}

Finally, before turning to the examples, we want to add an important note about our intentions and goals. The examples are designed to be illustrative and pedagogical---we choose them for their simplicity and clarity, and to show how prior empirical and computational work from cognitive science can be related under this general framework to language. Each example gestures at a larger domain of reasoning, but, of course, each domain is much broader than what we can implement here. Each example is also representative of a wide class of computational cognitive models that can be instantiated in a probabilistic language of thought, and which we propose can be integrated with natural language inputs and outputs under a rational meaning construction framework. In each section we therefore also discuss how this framework might be scaled, and what more work may be necessary, to scale from these examples towards a richer model of language in relation to those domains. 

We also hope that these examples, and other variations that elaborate on them and on the core domains of reasoning we discuss here, will offer useful starting points for more rigorous, systematic, cognitively-oriented evaluation and interpretation of the reasoning processes emergent in large language models and other language-based AI systems. In our own preliminary evaluations of these domains, we find that current large language models show many of the properties we discuss in the introduction. In some cases they appear to approximate implicitly the representations and algorithms we seek to model explicitly.  In others, particularly with more complex modifications beyond these simple examples, we find that large language models left to their own devices produce outputs that diverge from our intuitions. We seek here to model the representations with which \textit{people} make meaning from language in relation to all of these domains, but hope that these frameworks will be useful for understanding other computational systems that use language as well, including interpreting the representations that large language models already learn or should seek to acquire.

\vspace{1em}
\begin{tcolorbox}[title=\textbf{Graphical conventions}, colback=white,colframe=white!50!black]
\textit{Throughout the examples presented in this paper:}\\

\raisebox{-0.75\fontcharht\font`\B}{\includegraphics[height=2.5\fontcharht\font`\B]{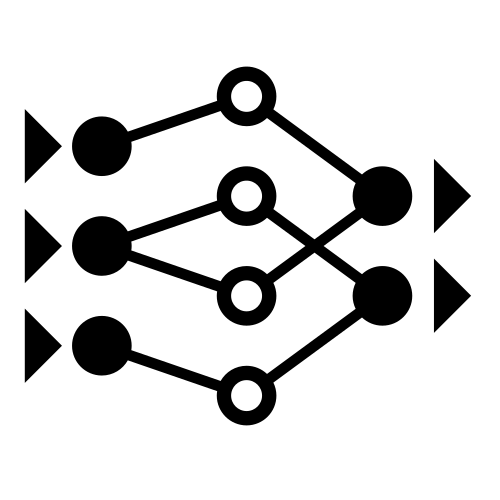}}
\textbf{Translations} mapping from language into probabilistic programs, produced by \texttt{Codex}, are indicated by a neural network icon.\\

\raisebox{-0.75\fontcharht\font`\B}
{\includegraphics[height=2.5\fontcharht\font`\B]{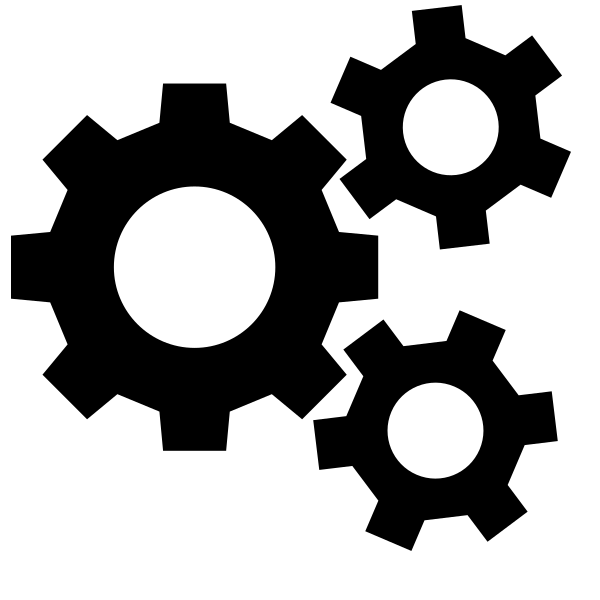}} 
\textbf{Probabilistic inferences}, performed by \texttt{Church}, are indicated by a cog icon.
\end{tcolorbox}

\clearpage
\subsection{Understanding language with probabilistic reasoning}\label{sec:models-probabilistic-reasoning}
To illustrate our framework, let's consider a concrete scenario that involves reasoning from language in the face of uncertainty. Suppose a friend is telling you about a tug-of-war tournament that took place the prior weekend in which the authors were participating:

\begin{quote}
\textit{Right off the bat, Josh won against Lio. He then proceeded to claim victory against Alex. Even working as a team, Lio and Alex still could not beat Josh!}
\end{quote}

\noindent In order to understand this story, it is useful to construct a little mental model: there are different players, they face each other solo or in teams, and based on his track record, Josh appears to be particularly strong. Now, suppose your friend tells you about a newcomer: \textit{In a huge upset, Gabe managed to best Josh in the fourth round.} Maybe Gabe is even stronger than Josh! Or, perhaps Josh was simply feeling lazy in the last match, in which case, Gabe might not actually be so strong. To clarify, you might ask a question, \textit{Who is stronger: Gabe or Josh?} Your friend's answer, which might itself express uncertainty, will nevertheless provide further information for you to incorporate into your understanding.

In making meaning from language about a scenario like the above, you are engaging in \textit{probabilistic reasoning}: integrating over different possibilities in order to infer likely explanations. People are remarkably proficient at making inferences from exactly this kind of sparse evidence. Sometimes, we acquire this evidence through direct experience---by watching the tournament, for instance---but often, this kind of information comes to us through language that cues us to update our beliefs accordingly. Critically, in order to reason \textit{consistently}, we need to represent core aspects of the situation: who are the different actors, what events took place, and what inferences have we already made? To this end, it is extremely useful to have a \textit{world model}, which we defined earlier as a probabilistic generative model that encapsulates the key mechanics of a domain and facilitates coherent, causal explanations of events. In this section, our aim is to further formalize what exactly {we} mean by world models and how large-scale neural models might serve as an interface between natural language and these kinds of cognitive representations.

\begin{figure}[hbtp]
    \centering
    \includegraphics[width=\textwidth]{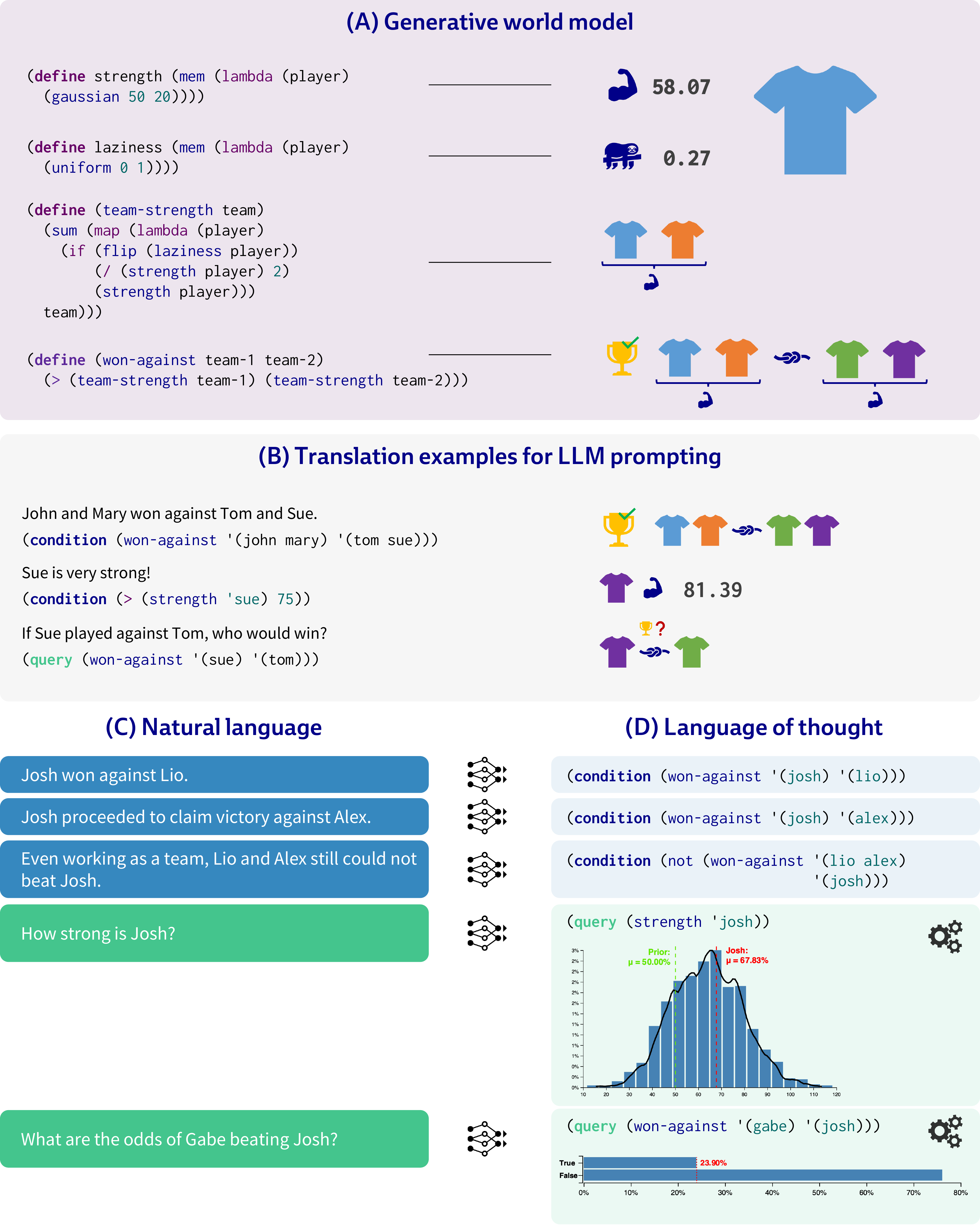}
    \caption{Illustration of probabilistic reasoning via language-to-code translation in the tug-of-war domain. (A) The generative model defines two latent traits, \texttt{strength} and \texttt{laziness}, and specifies how these interact to determine \texttt{team-strength}. By combining (A) and (B), we can few-shot prompt an LLM to translate open-ended natural language (C) into Church statements (D) that capture linguistic meaning with respect to the domain. The resulting probabilistic inferences transparently represent the model's beliefs and naturally capture human-like intuitions about players' latent traits.}
    \label{fig:tug-of-war-schematic}
\end{figure}

\paragraph{World models as generative programs.} The core of each example in this paper is a probabilistic generative model that defines the mechanics of a domain. For the purposes of this demonstration, and throughout \cref{sec:language-and-world-models}, we focus on reasoning from language given a pre-specified world model. Later, in \cref{sec:growing-and-constructing}, we show how language can be used to grow out and construct new world models. 

As a playground for this initial demonstration, we consider the ``Bayesian tug-of-war,'' a classic experimental domain in cognitive science that requires making inferences about the latent traits of individuals from sparse evidence. Prior work establishes that Bayesian inference in a probabilistic generative model closely captures people's predictions about scenarios in the tug-of-war \citep{gerstenberg2012ping, goodman2014concepts}, and that simple sentences can be mapped onto queries in this model~\citep{goodman2015probabilistic}. Here, we build on this work to give an account for how people might turn open-ended natural language into statements in the probabilistic language-of-thought. 

In tug-of-war, we start with a generative model of a tournament in which players of varying strengths compete in a series of matches, facing off either solo or as part of fluid teams (\cref{fig:tug-of-war-schematic}A). Each player has a latent strength value randomly sampled from a Gaussian distribution (with parameters arbitrarily chosen as $\mu=50$ and $\sigma=20$). As an observer, our goal is to infer the latent strength of each individual based on their win/loss record. However, players sometimes don't pull at their full strength and each player has a different intrinsic ``laziness'' value (uniformly sampled from the interval $[0,1]$) that describes how likely they are to be lethargic in a given match. The full Church code for the tug-of-war is given in \cref{appendix-sec:probabilistic-generative}.

\paragraph{Linguistic meanings as probabilistic program expressions.}
While the generative model defines the generic mechanics of the domain, we want to be able to talk about \textit{specific} people and events. In our framework, we focus on two kinds of linguistic utterances:

\textbf{Observations} provide information about people, objects, and events in the world; e.g., ``Josh faced off against Lio and won.'' In our framework, we translate observations into \lstinline[language=Lisp]{condition} statements in Church, which update the state of the world model to reflect new facts. Note that \lstinline{condition} statements have no return value; instead, they constrain the world model such that downstream inferences must be consistent with respect to the conditioning statement.

\textbf{Questions} seek information in the face of uncertainty about the world; e.g., ``Would Josh beat Gabe if they played again?'' In our framework, we translate questions into \lstinline{query} statements in Church, which evaluate the quantity of interest. Calling \lstinline{query} triggers a probabilistic computation that simulates possible worlds under the model, constrained by any observations so far. The query expression is evaluated in each simulated world, yielding multiple samples that form a posterior distribution over the value of interest.

Throughout the examples in this work, we freely interleave \lstinline{query} and \lstinline{condition} statements, much as questions might occasionally arise between statements of fact in a natural dialogue. Implementationally, this behavior is achieved through a read-evaluate-print loop (REPL) inspired by Venture's~\citep{mansinghka2014venture}, that evaluates queries against all \lstinline{condition} statements that have appeared up to that point in the dialogue history. In our model, we assume that the user specifies whether each utterance is a \lstinline{condition} or a \lstinline{query}, but LLMs could likely classify unannotated utterances accurately.

\paragraph{Translating from natural language to program expressions.} Inspired by the work of \cite{goodman2015probabilistic}, if we had some way to translate linguistic utterances into probabilistic program statements, we could perform a wide variety of probabilistic inferences from plain English. Up until recently, however, it was unclear how to construct a \textit{meaning function} sufficiently general to translate open-ended natural language into highly structured expressions compatible with a Church model. Our core observation is that language-code LLMs have many of the properties necessary to serve as a useful meaning function: broad-coverage exposure to natural language, a robust capacity to model joint language-code text distributions, and the ability to quickly grasp domain-specific syntax and semantics from a few examples.

In this work, we leverage the few-shot prompting capabilities of one such LLM, the Codex model from OpenAI, to induce a translation model from English to Church code. As it turns out, we only need to provide a small handful of \textit{example translations} (represented in \cref{fig:tug-of-war-schematic}B) to achieve a variety of interesting behaviors. To translate a new language utterance to Church, we simply concatenate the generative model (full text in \cref{appendix-sec:probabilistic-generative}) and the translation examples (full text in \cref{appendix-sec:probabilistic-example-translations}) into a prompt whose final line is the utterance. We then generate from Codex, which, based on the comment-code pattern in the prompt, infers that the completion should be written in Church, using the function definitions and constructs provided in the prompt.

Notice the high degree of variation in phrasing and lexical choice in \cref{fig:tug-of-war-schematic}C; none of the utterances contain ``won'' or ``against,'' yet Codex still maps these to the \lstinline{won-against} function. Here, we start to see some of the advantages of using an LLM over more traditional semantic parsing techniques like CCG parsers \citep{artzi-zettlemoyer:2013:TACL,artzi-lee-zettlemoyer:2015:EMNLP}. Because the model is pre-trained on a vast amount of linguistic data, it fluently handles many different kinds of linguistic variation. However, by including the Church generative model in the prompt, we can effectively constrain the output space; the model infers that the generated code should use the functions defined in the generative model. 

As a semantic parsing tool, this combination of pre-training and prompting manages to achieve broad invariance to spurious linguistic variation while remaining sensitive to wording choices that might affect meaning. We can see this tradeoff at work in \cref{fig:tug-of-war-schematic}C, where the translation uses a negation, closely reflecting the structure of ``Lio and Alex still could not beat Josh.'' Of course, there are multiple aspects of the utterance that this translation does \textit{not} capture (e.g., ``Even working as a team...'' suggests that Lio and Alex's efforts were well-coordinated; as opposed to something like, ``Stepping on each other's toes the whole match...,'' which would imply the opposite). Our point is not that the LLM translation perfectly captures all aspects of the utterance meaning, but rather, that it encodes those that are relevant to and compatible with the domain model so as to facilitate downstream reasoning.

\paragraph{Reasoning about scenarios with probabilistic inference.} So far, we've illustrated how we might \lstinline{condition}  a PLoT model on natural language, but what about reasoning? After hearing the information in \cref{fig:tug-of-war-schematic}C, we might assume that the player named Josh is quite strong. Exactly how strong is Josh, though? And how likely is it that he would beat another player who isn't Lio or Alex? Just as we used Codex to translate facts into \lstinline{condition} statements, we can use it to translate questions into \lstinline{query} statements in Church. The Church inference engine then automatically simulates scenarios (in this case, 1000 times) that are consistent with the given \lstinline{condition} statements in order to produce an approximate posterior distribution over each query.

By offloading reasoning from the LLM to the PLoT, we can obtain a much richer picture of our model's beliefs about the world (\cref{fig:tug-of-war-schematic}D). While the LLM alone can only respond with textual statements like ``Josh is very strong,'' Church gives us an entire probability density over Josh's strength (on expectation, he is a little less than one standard deviation above the average \lstinline{strength = 50}). Likewise, we can easily obtain a distribution over the outcomes of a Gabe-Josh match (given Josh's strong track record, our model finds Gabe's chances slim, at 23.90\%). Critically, Church is doing much of the heavy lifting of inference in the background in order to produce these posterior distributions.

\ornament

\begin{figure}[ht!]
    \centering
    \includegraphics[width=\textwidth]{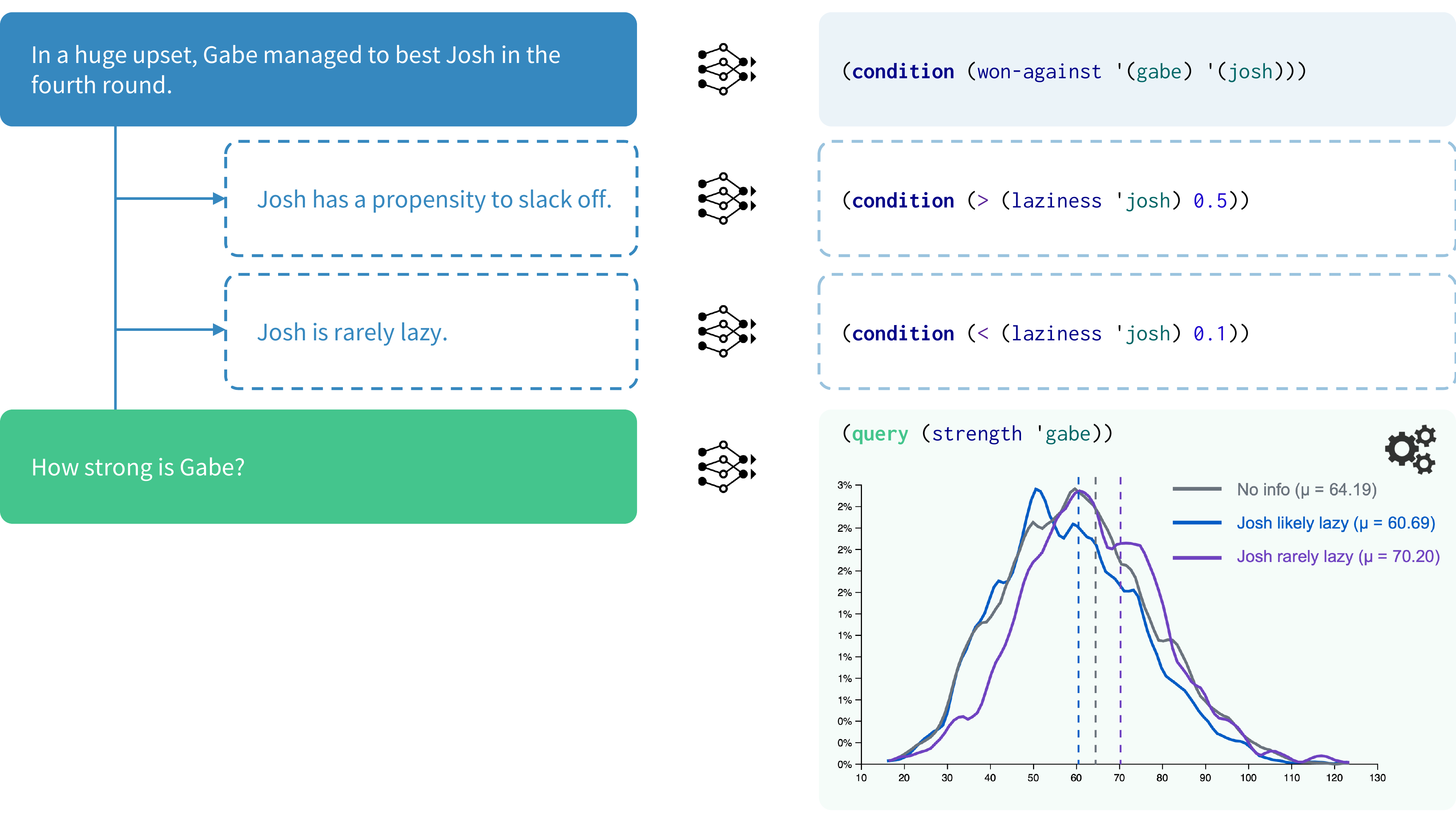}
    \caption{Reasoning about a pair of hypothetical scenarios with language-code translation. In a world where Josh is often lazy, Gabe's win is counteracted by a high likelihood that Josh threw the match. Conversely, in a world where Josh is rarely lazy, Gabe's win is surprising and suggests a high strength value. Rational meaning construction with an LLM appropriately resolves the linguistic meaning of these two scenarios, selecting reasonable probability parameters for the conditioning statements. Meanwhile, probabilistic inference about Gabe's strength is finely sensitive to the implications of these competing hypotheses.}
    \label{fig:tug-of-war-advanced-example}
\end{figure}

In addition to providing useful interpretability, reasoning in Church models is sensitive to each new piece of information. Much like human learners, Church models can flexibly update their beliefs when presented with low-probability or unanticipated events. Picking up our tug-of-war saga, consider the plot twist in \cref{fig:tug-of-war-advanced-example}:

\begin{quote}
\textit{In a huge upset, Gabe managed to best Josh in the fourth round.}
\end{quote}

\noindent How might this new information shape our interpretation of the match 4 outcome? If Josh is likely to be lazy, then it's possible that Gabe simply got lucky and wasn't so strong after all. If, on the other hand, Josh is rarely lazy, we might start to regard Gabe as particularly strong. In \cref{fig:tug-of-war-advanced-example}, we can observe how Church reasons about these two possibilities, shifting the probability density over Gabe's strength left if Josh is likely lazy and right if Josh is rarely lazy.

Note how, in order to translate a phrase like ``Josh has a propensity to slack off,'' Codex must choose a particular probability threshold. This choice is arbitrary and, while there is no ``correct'' answer, we see that Codex is able to choose valid probability values between [0, 1] that feel appropriate to the wording: a ``propensity to slack off'' doesn't necessarily imply that someone slacks off \textit{all the time}, while, in contrast, ``rarely lazy'' offers more certainty. Indeed, across many different contexts, we observe that Codex is able to pick reasonable parameter values that respect both the language and the parametrization of defined distributions. We consider these inferences to represent a form of ``amortized pragmatics'' \citep{goodman2015probabilistic}, which we will revisit in \cref{sec:open-questions}.

\begin{figure}[ptbh]
    \centering
    \includegraphics[width=\textwidth]{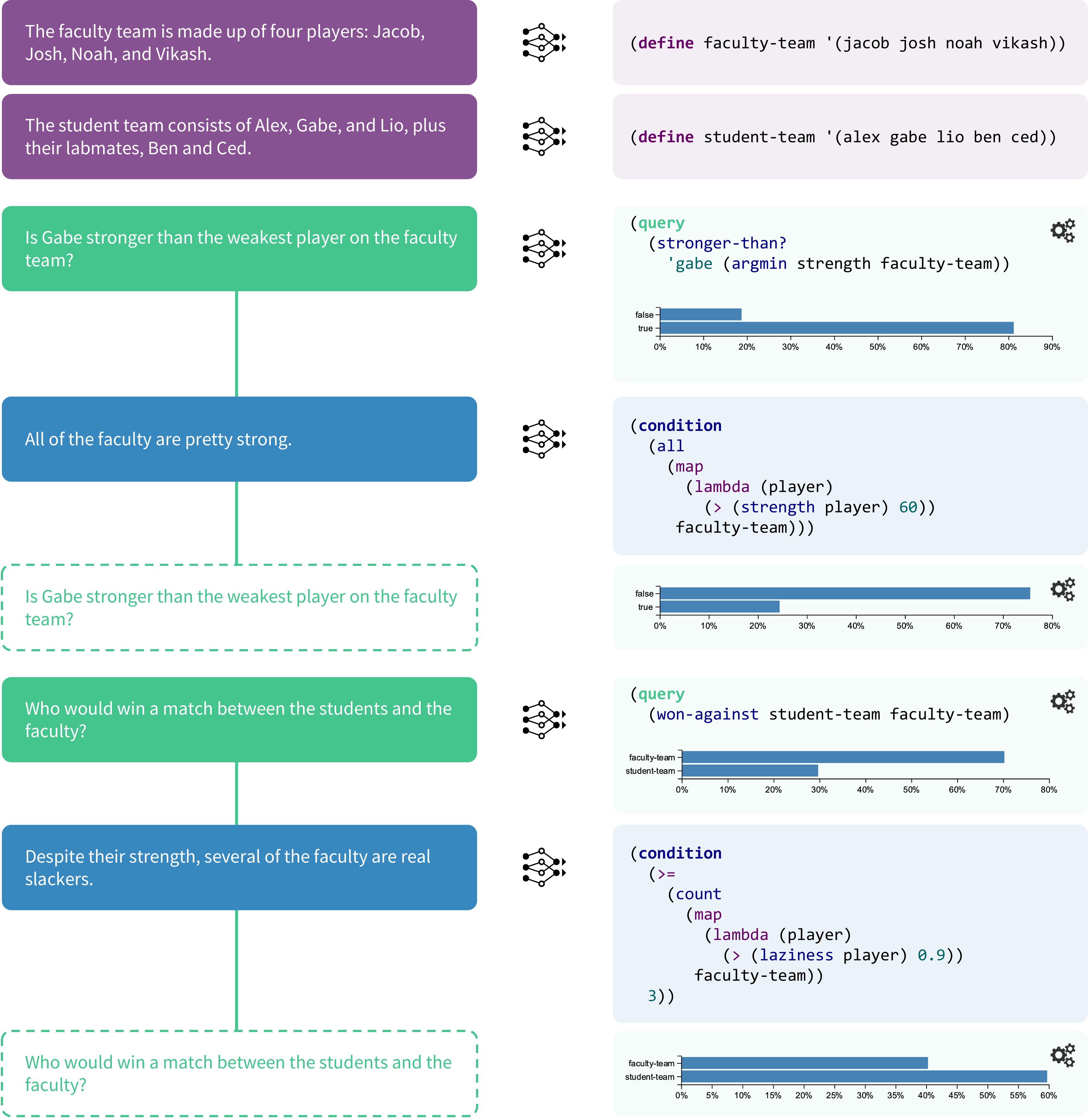}
    \caption{In this final tug-of-war dialogue, natural language plays three interlaced roles in interfacing with the language-of-thought. Definitions (purple) introduce new concepts, such as specific player-teams, that can later be referenced symbolically. Observations (blue) translate into condition statements that probabilistically constrain the world state, sometimes amortizing the resolution of linguistic ambiguity (e.g., ``pretty strong'' or ``real slackers''). Finally, questions (green) translate into queries that trigger inference by probabilistic simulation over possible worlds that is both sensitive to and consistent with prior definitions and observations.}
    \label{fig:tug-of-war-final-example}
\end{figure}

\paragraph{Putting it together: the power of probabilistic reasoning.} We conclude this section with a final example that underscores the flexibility of our framework to model complex reasoning from language and foreshadows multiple themes that we will revisit later in the paper. Consider the dialogue in \cref{fig:tug-of-war-final-example}, in which the students and faculty team up to face one another. The interlocutor poses two questions: ``Is Gabe stronger than the weakest player on the faculty team?'' and ``Who would win in a match between the students and the faculty?'' As we saw in the prior tug-of-war examples, the answers to both of these questions are expressed as probability distributions derived from simulation of the generative tug-of-war model. Moreover, in both cases, the introduction of new information flips the model's belief state in a way that aligns with human intuitions. In this way, the PLoT framework is natively capable of \textit{defeasible inference}---a phenomenon of human reasoning that was of great interest to early AI pioneers of non-monotonic logics \citep{mccarthy1980circumscription, ginsberg1987readings}. 

A key advantage of our framework is that achieving these kinds of defeasible and flexible inferences from natural language reduces to grounding utterances into appropriate \lstinline{condition} and \lstinline{query} statements. While the observations and questions in \cref{fig:tug-of-war-final-example} are semantically more complex than those that appeared in the prior examples, and though there are many degrees of freedom involved in the translation problem, we confirm that an appropriately-prompted LLM can produce translations that intuitively capture the meaning of each utterance with respect to the tug-of-war domain. Moreover, as we saw in \cref{fig:tug-of-war-advanced-example},  Codex is able to \textit{amortize} certain pragmatic inferences in resolving ``pretty strong'' to a threshold of \lstinline{strength > 60}, ``real slackers'' to a threshold of \lstinline{laziness > 0.9}, and ``several of the faculty'' to \lstinline{count >= 3}. How far can we go with these kinds of amortizations? Throughout \cref{sec:language-and-world-models} and \cref{sec:growing-and-constructing}, we will see examples of context-driven amortizations across different domains; and in \cref{sec:open-questions}, we will regroup to discuss how these different examples of amortization might inform our theories of language understanding and pragmatics.

In this dialogue, we also give a preview of \lstinline{define}, a powerful construct in our framework that is discussed in depth in \cref{sec:growing-and-constructing}. Just as people come up with terms like ``20th-century pragmatists'' or ``Meatless Monday'' to pick out entire hierarchies of people, things, and events, a core feature of the probabilistic LoT is the ability to \lstinline{define} new concepts that can later be referenced symbolically. In the \cref{fig:tug-of-war-final-example} dialogue, language about team structures defines two new concepts, \lstinline{faculty-team} and \lstinline{student-team}, that facilitate concise translation of language like, ``Is Gabe stronger than the weakest player on the faculty team?'' Moreover, while \lstinline{faculty-team} is a static list, other defined concepts can ground out in \textit{functions} that take arguments. In fact, \lstinline{stronger-than?}, which is defined in the prompt (\cref{appendix-sec:probabilistic-example-translations}), is one such example, illustrating how programming languages are well-suited to capture the infinite productivity of language that arises through structured composition. Through this lens, we can start to imagine how our tug-of-war world model might be expanded to ground many new kinds of language:

\begin{itemize}
    \item The tug-of-war tournament is organized into three leagues for novices, amateurs, and professionals. In order to be considered a professional, a player must win 20 one-on-one matches against other professionals.
    \item Players often get increasingly tired over the course of a tournament, though some players have more stamina than others.
    \item The tournament has an entry fee of \$20 per contestant and a grand prize of \$10,000 for the winning team.
\end{itemize}

\noindent How can we grow our world models to incorporate new language, or even construct new world models entirely from scratch? In \cref{sec:growing-and-constructing}, we revisit the tug-of-war domain with an eye to precisely these questions.

\paragraph{Conclusions.} As an introductory example, the tug-of-war domain serves as a minimal illustration of the kind of reasoning from language that our framework is concerned with. Our goal here was to build intuition for our general approach: by translating natural language into \lstinline{condition} and \lstinline{query} statements as inputs to a probabilistic inference engine, we can achieve forms of reasoning from language that are consistent with respect to a mental model of the world. Nonetheless, in scaling this approach beyond the toy domain of tug-of-war, many questions arise. How does probabilistic inference relate to models of relational and deductive reasoning of the sort that classical AI approaches excel at? How do we ground linguistic meaning in the visual and physical world? And how does language understanding inform our actions and interactions with other agents through goal-directed planning? In \cref{sec:language-and-world-models}, we will progressively expand our scope to touch on each of these questions and show that, in each case, new kinds of language understanding and reasoning can be naturally incorporated into our framework.

\clearpage
\section[World models]{Understanding and reasoning about language with world models}\label{sec:language-and-world-models}
In this section, we illustrate how the general framework we propose in \cref{sec:overview} can be applied and extended to integrate natural language with core domains of human-like thought. In each, we build on the idea that language that conveys observations and questions about uncertain situations, constructing meanings from a generative world modeling program that supports \textit{probabilistic reasoning}. In \cref{sec:models-relational-reasoning}, we show how this approach can be integrated to understand language that conveys structured, logical lexical relations. In \cref{sec:models-reference-and-grounding}, we show how generative programs that support perceptual and physical simulation can be used to ground language about scenes into visual world. Finally, in \cref{sec:models-agents-and-planning}, we consider language about agents with preferences and goals, and show how we can make meaning from sentences with respect to a generative program that supports planning.

\subsection{Language for logical and relational reasoning}\label{sec:models-relational-reasoning}
In the previous section, we examined how translation from natural language into the probabilistic language of thought naturally captures a certain form of reasoning in which uncertainty plays a key role. How does this framework relate to earlier computational theories of reasoning, such as classical AI approaches to logical and relational reasoning \citep{russell_norvig_2021}? Historically, systems like Prolog \citep{philippe1972definition, colmerauer1972systeme} were designed for similar goals to ours here, to allow people to directly interact with computers via natural language (French, originally), specifying only the background knowledge and goals for computation without the algorithmic details \citep{colmerauer1996}. In this section, we demonstrate how the PLoT not only fully supports the style of deductive, logical reasoning characteristic of classical AI, but extends it to support \textit{inductive} inferences as well. Moreover, we argue that many kinds of real-world reasoning problems that are traditionally modeled using structured logic-based approaches actually require a mix of both symbolic and probabilistic reasoning. In doing so, we aim to illustrate how our approach of translating from natural language to the PLoT fluidly integrates both kinds of reasoning in a way that comes naturally to people, but that has proved elusive for both traditional deductive programming systems and purely statistical language models.

\paragraph{Language about kinship relations.} Suppose you are again with your friend from \cref{sec:models-probabilistic-reasoning}, who is telling you about a part of their extended family. ``Avery has a sister named Blake, and their father is named Charlie,'' your friend says. Immediately, you start to sketch a picture in your mind of this family, which you can update on-the-fly as you get more information: ``Charlie is the grandfather of Dana.'' At this point, you can infer that one of Charlie's kids is also Dana's parent, but which one? In the absence of additional information, it's a toss-up between Avery and Blake, with some outside chance that there could be another, unmentioned sibling who is Dana's parent. Hearing that ``Blake has two kids'' might initially shift your beliefs towards Blake. However, upon learning that ``Dana is an only child,'' you'd have to rule Blake out entirely! This kind of relational reasoning, which freely intermixes deductive and inductive inferences, comes quite naturally to people. How do we make such rich inferences from a relatively sparse sequence of words?

In this section, our domain of interest will be \textbf{kinship}: relationships between people in a family. The kinship domain provides fertile ground for the study of logical reasoning for several reasons. First, during development, one of the first domains where we learn about logical relations is in describing families \citep{piaget1951judgment, elkind1962children}. Language has evolved to describe family structures in highly economical terms that naturally express composition (e.g., my mother's father is my \textit{grandfather}) and symmetry (e.g., if Avery is my cousin, then I am Avery's cousin; together, we are \textit{cousins}). Nevertheless, while certain kinship references are relatively straightforward (e.g., ``Blake's mother''), others involve ambiguity (e.g., ``Blake's uncle'' could refer to the brother of \textit{either} of Blake's parents; or even, perhaps, a close older male who is not related by blood or marriage). Finally, kinship reasoning freely intermixes deductive and inductive inferences: for instance, ``Charlie has a grandson named Dana'' \textit{deductively} implies the existence of a child of Charlie who is also a parent to Dana; and it \textit{inductively} implies that Charlie was possibly partnered at some point, such that Dana might have another grandparent in the picture. Traditional logical accounts of reasoning in this domain capture the deductive inferences but not the inductive inferences in cases like this. People, in contrast, routinely make statements such as ``This is Kendall, the partner of Avery's niece'' with the expectation that others will draw roughly the same inferences they would in building a mental model of this family: Avery has a brother or sister, and that sibling has a female child, and Kendall is that person's partner. In sum, the kinship domain offers a rich set of relations and possible inferences, and comes equipped with an extensive natural language vocabulary, making it an ideal playground to explore our translation hypothesis.

\paragraph{World models of kinship as probabilistic generative programs.} Owing to the richness of the domain, recent years have seen a steady interest in computational cognitive models of various aspects of kinship, ranging from development and acquisition of kinship terms across cultures \citep{mollica_logical_2022, mitchell_ontogeny_2021}, tradeoffs in communicative efficiency in natural \citep{jones_human_2010, kemp_kinship_2012} and artificial \citep{smith_simple_2020} kinship systems, and probabilistic inferences about kinship relations from sparse evidence \citep{katz_modeling_2008}. In this work, our primary interest is in how people represent and reason about kinship relations conditioned on language. Following \cite{katz_modeling_2008}, we construct an intuitive domain theory of kinship using a probabilistic generative model and a small number of rules that form a conceptual system.

\begin{figure}[t!]
    \centering
    \includegraphics[width=\textwidth]{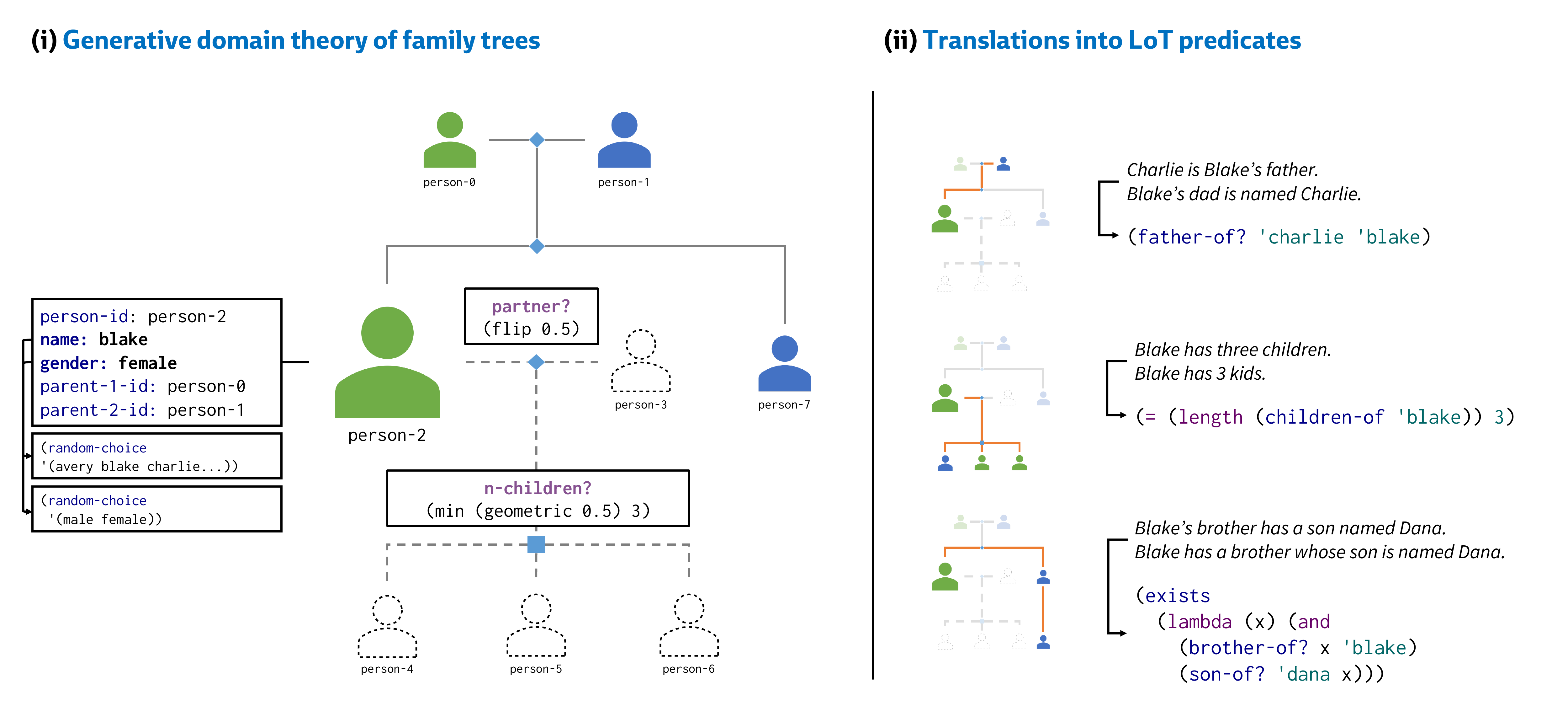}
    \caption{Illustration of a simple kinship domain theory and conceptual system implemented in Church. (i) The generative model specifies a process by which individuals form couples and have children to form family trees. Each tree represents a ``possible world'' in which certain relationships hold. (ii) These relationships are expressed using predicates in a conceptual system that supports quantificational logic and composition, giving rise to an expressive domain semantics that aligns well with natural language.}
    \label{fig:relational-reasoning-kinship-schematic}
\end{figure}

As in \cref{sec:models-probabilistic-reasoning}, our kinship domain theory is expressed as a generative model in Church. In the Bayesian tug-of-war, the generative model consisted of random variables over continuous quantities like \texttt{strength} and \texttt{laziness}. In contrast, in this section, our generative model specifies a series of discrete random choices that describe events in a family's genealogy: people are born, find partners, have children, and the process repeats. All of these events involve random choices that shape the makeup of the family tree.

\cref{fig:relational-reasoning-kinship-schematic} (i) shows a schematic of the kinship generative domain theory. When a person is born, they are assigned a unique \lstinline{person-id}, a name\footnote{For simplicity, names uniquely reference individuals in the tree, so as to avoid confusing scenarios like ``Avery is the mother of Avery.'' Additionally, for efficiency of inference, the only names that are assigned are ones that are used in the conversational context.} sampled from a list of gender-neutral names, and a gender sampled from \{male, female\}. Next, with fixed $p=0.5$, the person partners with a new individual from outside the family. Finally, if partnered, the couple has $n=\{0, 1, 2, 3\}$ children, with the number of kids drawn from a geometric distribution ($p=0.5$). This process repeats recursively until a full family tree is generated. To support efficient inference using Church's generic sampling algorithms, we cap the trees at $3$ generations and limit each couple to $3$ children. Further implementation details of the generative model can be found in  \cref{appendix-sec:kinship-generative}.

As with any computational model of a social phenomenon, this toy kinship model is reductive of many important nuances of identities and relationships. For instance, while the model includes both same- and opposite-gender couples, these couples never split, so step-relations aren't well-captured. While these kinds of compromises are designed to keep inference tractable, still others stem from limitations of the language itself. For example, many colloquial English kinship terms are gender-binary (e.g., mother, grandfather, daughter), so instantiating them as truth-conditional predicates coerces the generative model towards traditional gender assignments. Similarly, many English names carry strong gender associations, which NLP systems trained on large linguistic corpora pick up on \citep{caliskan2017, grand2022semantic}. In our examples, we intentionally select gender-neutral names (e.g., Avery, Blake, Charlie, Dana) to emphasize that these naming-based gender inferences are deliberately not part of the reasoning task. 

To summarize, language both reflects and constrains our intuitive theories of complex domains like kinship (\citealp{sapir1929status, whorf1956language}; c.f. \citealp{gentner2003whither} for a review of contemporary perspectives on linguistic relativity), and these tradeoffs manifest concretely in the toy model presented in this section. Fortunately, where this initial ``off-the-shelf'' kinship model lacks social and cultural nuance, our framework offers opportunities to extend and modify these areas. In section \cref{sec:growing-world-model}, we look at ways of growing our kinship model to include concepts from non-English-speaking cultures and more inclusive concepts of gender.

\begin{figure}[htbp!]
    \centering
    \includegraphics[width=\textwidth]{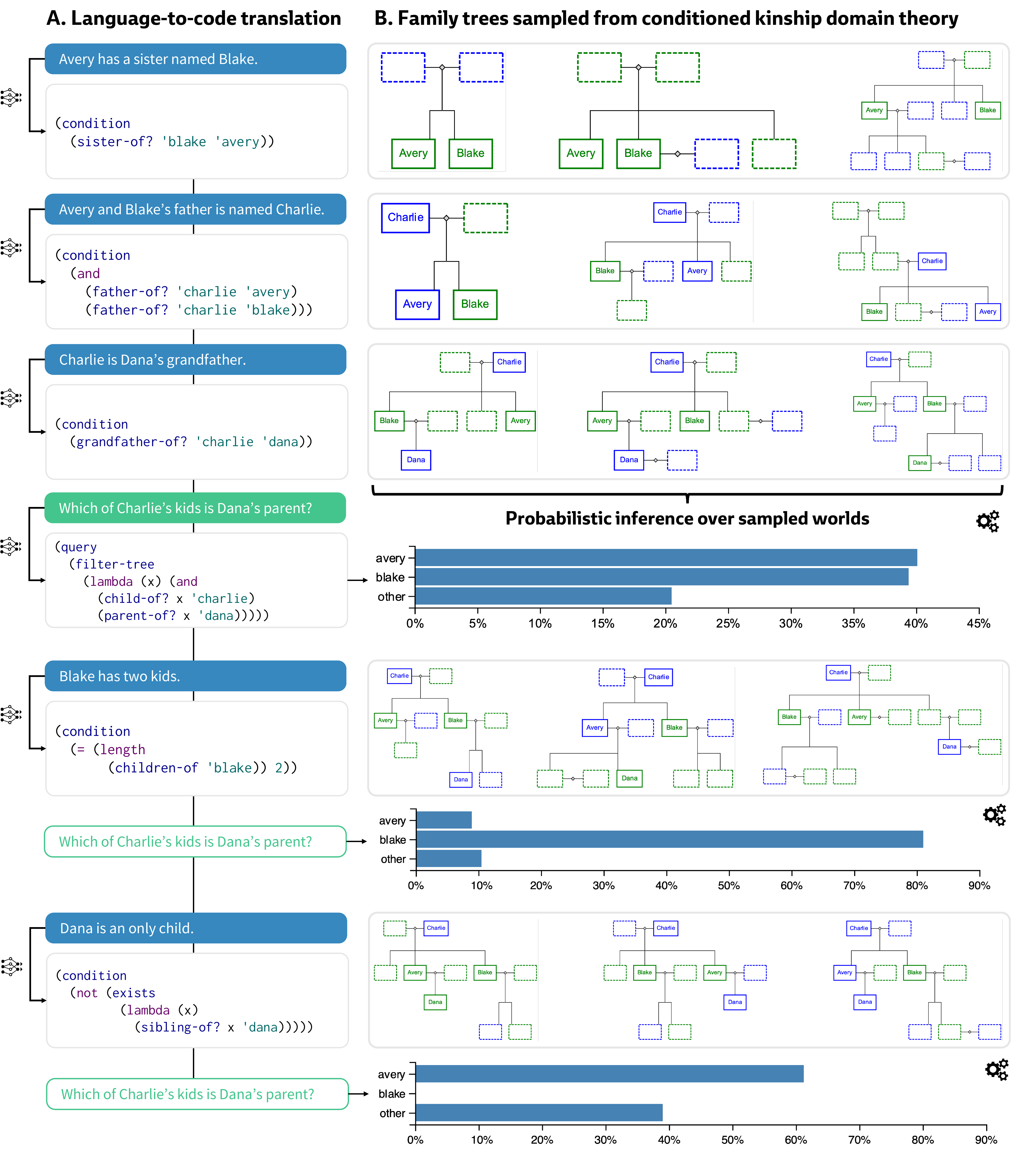}
    \caption{Kinship reasoning from natural language, backed by a domain theory model in the probabilistic language of thought. (A) Natural language utterances about a particular family are readily translated into Church conditioning statements by a LLM. (B) Samples from the conditioned generative domain model are possible family trees that adhere to the growing set of constraints (conditioning statements are cumulative). Reasoning about unknown kinship relations is accomplished through posterior inference against a translated query. With each new piece of information, the model's beliefs reflect both deductive and inductive inferences. }
    \label{fig:relational-reasoning-kinship-sequence-1}
\end{figure}

\paragraph{Relational meanings as program statements.} Given a generative model of family trees, we can define a rich conceptual system to make statements about relationships between individuals. Our conceptual system consists primarily of a dozen-odd \textit{derived predicates} that are binary operators over pairs of names; e.g., \lstinline{(father-of? 'charlie 'blake)} is true iff Charlie is the father of Blake in a particular tree instance.\footnote{Note that because our model includes same-gender couples, Blake may have one father, two fathers, or no fathers. Blake also may not exist in the tree in the first place! Crucially, these aspects of the generative model don't matter to the derived predicate, which simply evaluates whether the relationship in question holds somewhere in the tree.} These derived predicates build on a small number of low-level accessor functions that operate directly on nodes in the tree data structure. For instance, \lstinline{(children-of 'blake)} returns a list of names corresponding to the children of Blake in the tree. Finally, our conceptual system includes several higher-order functions, like \lstinline{map-tree}, \lstinline{filter-tree}, and \lstinline{exists} that take custom predicates as inputs and return a boolean. These functions facilitate the expression of a rich compositional semantics by allowing for compound predicates containing conjunctions and disjunctions. \cref{fig:relational-reasoning-kinship-schematic} (ii) illustrates several examples of the kinds of statements that can be made using combinations of derived predicates, low-level accessors, and higher-order functions. The full set of definitions making up the conceptual system is given in \cref{appendix-sec:kinship-conceptual-system}.

\paragraph{Translating from language to program expressions.} As in \cref{sec:models-probabilistic-reasoning}, we use a handful of paired natural language / code examples (\cref{appendix-sec:kinship-translation}) to induce a meaning function via Codex. Because the prompt also includes the generative model source code and the full set of derived predicates, the LLM is able to resolve statements like ``Blake has two kids'' to the appropriate function (in this case, \lstinline{children-of}) using the available definitions. Moreover, we observe zero-shot generalization to linguistic constructs that are not explicitly defined in the prompt, such as the concept of an ``only child'' (\cref{fig:relational-reasoning-kinship-sequence-1}).

\paragraph{Putting it together: Reasoning from language about kinship relations.} What is the purpose of all of this domain-specific machinery that we have now built up? The answer is two-fold. First, the \textit{generative domain theory} compactly captures the key dynamics of our domain, allowing us to reason about a combinatorially vast space of possible family trees. Meanwhile, the \textit{conceptual system} serves as a higher-level program interface, defining certain relationships that we would like to be able to talk about. Finally, the \textit{large language model} bridges the domain model with natural language, providing a flexible and context-aware way to ground language into conditioning and query statements.

In \cref{fig:relational-reasoning-kinship-sequence-1}, we can see how these components come together to facilitate naturalistic reasoning from language about kinship relations. Each natural language utterance translates to a \lstinline{condition} statement in Church that serves as a constraint on family trees. With each successive \lstinline{condition}, our uncertainty decreases and our picture of the family tree in question starts to crystallize. Samples from the conditioned domain theory model therefore serve as hypotheses about possible worlds that are consistent with the information provided through language. Furthermore, the \textit{distribution over conditioned samples} provides a principled way to reason about queries, such as \textit{Which of Charlie's kids is the parent of Dana?} Posterior inference (in this case, accomplished via rejection sampling) faithfully reflects various possible configurations and their relative probabilities. For instance, in \cref{fig:relational-reasoning-kinship-sequence-1}, after conditioning on \textit{Blake has two kids}, the model puts $>80\%$ probability on Blake being Dana's parent, but also factors in low-probability possible worlds where Avery or a third unnamed sibling is Dana's parent. Yet, despite this confident answer, the model can correctly infer that this same probability drops to $0\%$ in the face of the contradictory information that \textit{Dana is an only child.} Note that the distributional parser plays a crucial role in this inference by providing a correct interpretation of this utterance. Meanwhile, the Church inference engine does the heavy lifting of representing possible worlds and reasoning about them in a principled manner.

\paragraph{Future directions: Logical and relational reasoning with language models.} Significant recent attention has been directed towards studying reasoning in LLMs. Typical approaches involve engineering prompts so as to induce  structured generations in text space that approximate ``step-by-step'' reasoning \citep{nye2021show, wei2022chain, kojima2022large}. Nevertheless, current evaluations find that even with such methods, LLMs are prone to producing unfaithful reasoning chains in which conclusions do not follow logically from the premises \citep{golovneva2022roscoe, lyu2023faithful, ribeiro2023street, liu2023evaluating}. These issues of consistency have motivated several systems that connect LLMs to external symbolic inference engines that perform deductive inference using Prolog-style backwards chaining \citep{weirDynamicGenerationInterpretable2022, pan2023logic, dalvi2022towards}. We see this work as closely-related in spirit to our approach, but fundamentally limited to deductive reasoning. (See \cref{apx:why-not-prolog} for a technical explanation of these limitations.) Of course, we make no claim that Church or its derivatives are the \textit{only} languages that can capture human-like relational reasoning. For instance, ProbLog \citep{suster2021mapping, dries2017solving, problog}, a probabilistic extension of Prolog in which deduction rules can be annotated with probabilities, offers a compelling alternative. Indeed,  interfacing ProbLog with a natural language via an LLM-backed meaning function would constitute a promising instantiation of our rational meaning construction framework. Our core assertion here, and in the rest of this paper, is that \textit{representing probabilistic, generative models over possible worlds} is critical to reasoning coherently about a structured domains.

\clearpage
\subsection{Language for visual and physical reasoning}\label{sec:models-reference-and-grounding}
Sensory detail and physical knowledge pervade our everyday language. We can describe and imagine highly visual objects and scenes---\textit{a few red mugs on a tabletop}, a \textit{tall stack of blue plates}, \textit{a heavy box}, and objects that \textit{move}, \textit{bounce}, and \textit{collide}. We flexibly make predictions about physical events (\textit{what will happen if a kid crashes into that table stacked with plates?}), or infer the underlying physical properties of the world (\textit{how heavy is that box that no one can lift?}), based on situations described entirely in words. As with the other domains we have considered thus far, understanding this language requires integrating over the uncertainty inherent to language, like the possible heights picked out by \textit{tall} and motions picked out by a \textit{bounce}, as well as the uncertainty inherent to how we imagine the physical world itself.

How can we so flexibly relate language to our more general perceptual and physical reasoning? In this section, we illustrate how our overarching framework for language understanding can be modularly extended to capture both of these capabilities. We begin with perception, extending our framework to integrate a \textbf{graphics rendering engine} to relate linguistic meanings to visual knowledge (\cref{sec:models-static-scenes}). We then build on this approach to integrate a \textbf{physics simulation engine} to further interface between language and intuitive, probabilistic physical reasoning (\cref{sec:models-dynamic-scenes}). By incorporating these external engines, these sections blueprint how computational models that ground linguistic meaning in a PLoT can interface with other cognitive modules for perception and physical reasoning.

\subsubsection{Language about visual scenes}\label{sec:models-static-scenes}
\begin{figure}[ht!]
    \centering
    \includegraphics[width=\textwidth]{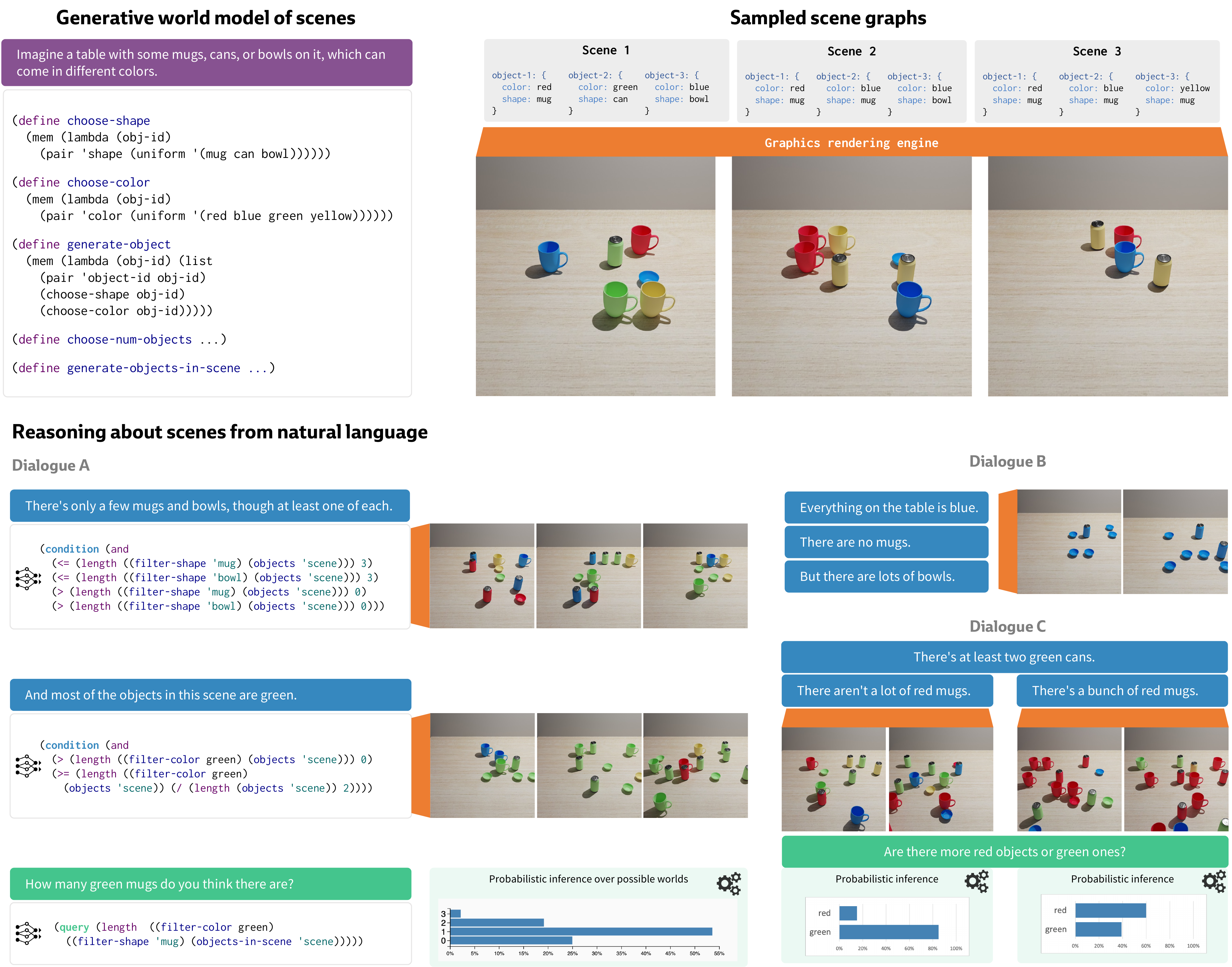}
    \caption{Human language understanding draws on our structured knowledge of the visual world. (\textit{Top}) A probabilistic generative model describes a prior over tabletop scenes with varying configurations of colored mugs, cans, and bowls. Sampled world states describe a scene based on symbolic object concepts. Interfacing this world model with a \textit{graphics rendering engine} models visual imagination of a given scene.
    (\textit{Bottom}) Language about particular visual scenes can now be translated as before into conditions (\textit{blue}) and queries (\textit{green}) on the distribution over scenes, which can be rendered into visual scenes that reflect language.}
    \label{fig:overview-scenes}
\end{figure}
To illustrate the structured relationship between language and visual knowledge, imagine how we might talk about a very simple domain of scenes (\cref{fig:overview-scenes}, top)---tables on which someone was placing some household objects (\textit{mugs}, \textit{cans}, or \textit{bowls}) that come in different colors (\textit{red}, \textit{green}, \textit{yellow}, or \textit{blue}.) 

Given descriptions of particular scenes (\cref{fig:overview-scenes}, bottom), most of us can easily picture tabletop scenes that fit these descriptions, updating what we imagine to incorporate arbitrary new information, like that \textit{everything on the table is blue}, and also that \textit{there are no mugs}, and \textit{lots of bowls}. We can do this despite uncertainty in the language itself---a phrase like \textit{lots of bowls} leaves open just how many bowls there are, though we have general intuitions that there should be more than one or even two bowls on our imagined table. You can also draw a host of fundamentally probabilistic inferences to answer many arbitrary questions about the scenes you imagine, like \textit{how many green mugs} there might be, or whether there are \textit{more red objects or green ones}. The set of scenes you imagine, and the way you answer these questions, is structured and compositional at the level of individual objects and their properties (\textit{a mug}, \textit{a green mug}, \textit{a bunch of green mugs}), and over successive sentences (like \textit{there are many red objects on the table}, \textit{there are just a few green mugs}, and \textit{there are also at least three green bowls}.)
The way we talk about scenes like these suggests the level of abstraction with which we mentally represent them. We describe and reason over object categories, lexical properties, numeric quantities, and set relations, and we can easily visualize scenes from these abstract, linguistic descriptions. In contrast, recent evaluations of current multimodal models---large language models fine tuned on corpora of images \citep{ramesh2021zero,ramesh2022hierarchical}---suggest that even large models struggle with just these kinds of simple but abstract relational concepts in language, such as producing images consistent with quantifiers like \textit{more red things than green things}, or relations like \textit{a plate on top of a cup} \citep{radford_language_2019,marcus2022very,conwell2022testing}. 

In this section, we propose that the basic motif outlined in our framework also suggests an alternate approach for relating language and visual reasoning. Our architecture draws on the traditions of viewing perception as ``analysis by synthesis'' or ``vision as inverse graphics'' from cognitive science and classic computer vision \citep{lee2003hierarchical,yuille2006vision,battaglia2013simulation,wu2015galileo,
kulkarni2015picture,gothoskar20213dp3}.
This approach frames visual imagination and visual scene understanding as two sides of the same coin, modeling visualization in a mental \textbf{graphics rendering engine} over internal scene representations and perception as probabilistic inference to \textbf{invert the renderer} and thereby recover the physical content of scenes from vision. In this section, we show how this general approach to modeling human perception can integrate cleanly into the framework we have sketched so far, augmenting the probabilistic language of thought with an interface to a rendering engine so it can serve as a general, flexible intermediary for relating language, world models, and visual scenes. 

\paragraph{Integrating the probabilistic generative model over scenes with a rendering engine.} To model the domain of tabletop scenes, we begin with a probabilistic generative model like those in the preceding sections. The generative program excerpted at the top of \cref{fig:overview-scenes} (\textit{purple}) describes a prior over the number of objects in a given scene, and the shape and color of each object. This program is similar in many ways to the kinship model in \cref{sec:models-relational-reasoning}, which generates possible family trees as a collection of entities and stochastic choices about each one. Similarly, the generative model in this domain generates a particular \textit{scene} by making stochastic choices over the number of objects in the scene (\lstinline{choose-num-objects}), then generates each individual object (\lstinline{generate-object}) based on stochastic choices over its possible properties (e.g \lstinline{choose-shape} and \lstinline{choose-color}). This basic structure can be augmented in many ways to model more complex scenes, with more variation over possible properties like \textit{size} or \textit{material}, hierarchical classes of object categories like \textit{dishware}, \textit{cups}, and \textit{mugs}, or hierarchical object structures like a \textit{stack of plates}.

Each sample from the generative model in \cref{fig:overview-scenes} is a structured, symbolic representation of a particular scene state, represented in our particular implementation as a list of object dictionaries that map between attribute kinds (like \lstinline{object-shape}) and values (like \lstinline{'mug}). These scene states are very simple instances of the many symbolic scene representations used throughout computer graphics and computational models of human scene understanding, data structures which model the abstract and semantic contents of scenes \citep{clark1976hierarchical,bar2003scenegraphs,johnson2015image,johnson2017inferring,armeni20193d,gothoskar20213dp3,zinberg2019structured}.

We can now extend this probabilistic generative program so that it expresses not just a distribution over possible scene states, but over the visual percepts of each scene. We do so by extending our base probabilistic programming language with a new function, \lstinline{render}, that takes in scene graphs as inputs and calls out to Blender, a 3D computer graphics engine.\footnote{\url{https://www.blender.org/}} Our \lstinline{render} implementation builds on the basic capabilities of any programmable graphics engine. It defines how symbolic object entities with the properties defined in our model (shapes like \textit{mug}) are rendered and colored into 3D CAD shapes, and can forward render any sampled scene graph into a visual scene with the requisite object types and colors, and overall structure (\cref{fig:overview-scenes}, top, \textit{Rendered possible worlds}). Collectively, this generative model and rendering interface unites the underlying belief distribution over possible scene states with how each of these scenes might look.

More broadly, this implementation is intended as a simple, illustrative example of how our framework could be integrated to model many, complex relationships between the objects we talk about in a scene and how they look --- recent work in scene understanding, for instance, models variation in lighting, viewer angle and distance from the scene, stereo depth sensing, and sources of noise in perception (such as from a viewer who only looks briefly at an image, or an imperfect, non-idealized visual sensor) (e.g., in \cite{gothoskar20213dp3,zinberg2019structured,kulkarni2015picture,mansinghka2013approximate,deng2021generative,hughes2022hydra}).

\begin{figure}[ht!]
    \centering
    \includegraphics[width=\textwidth]{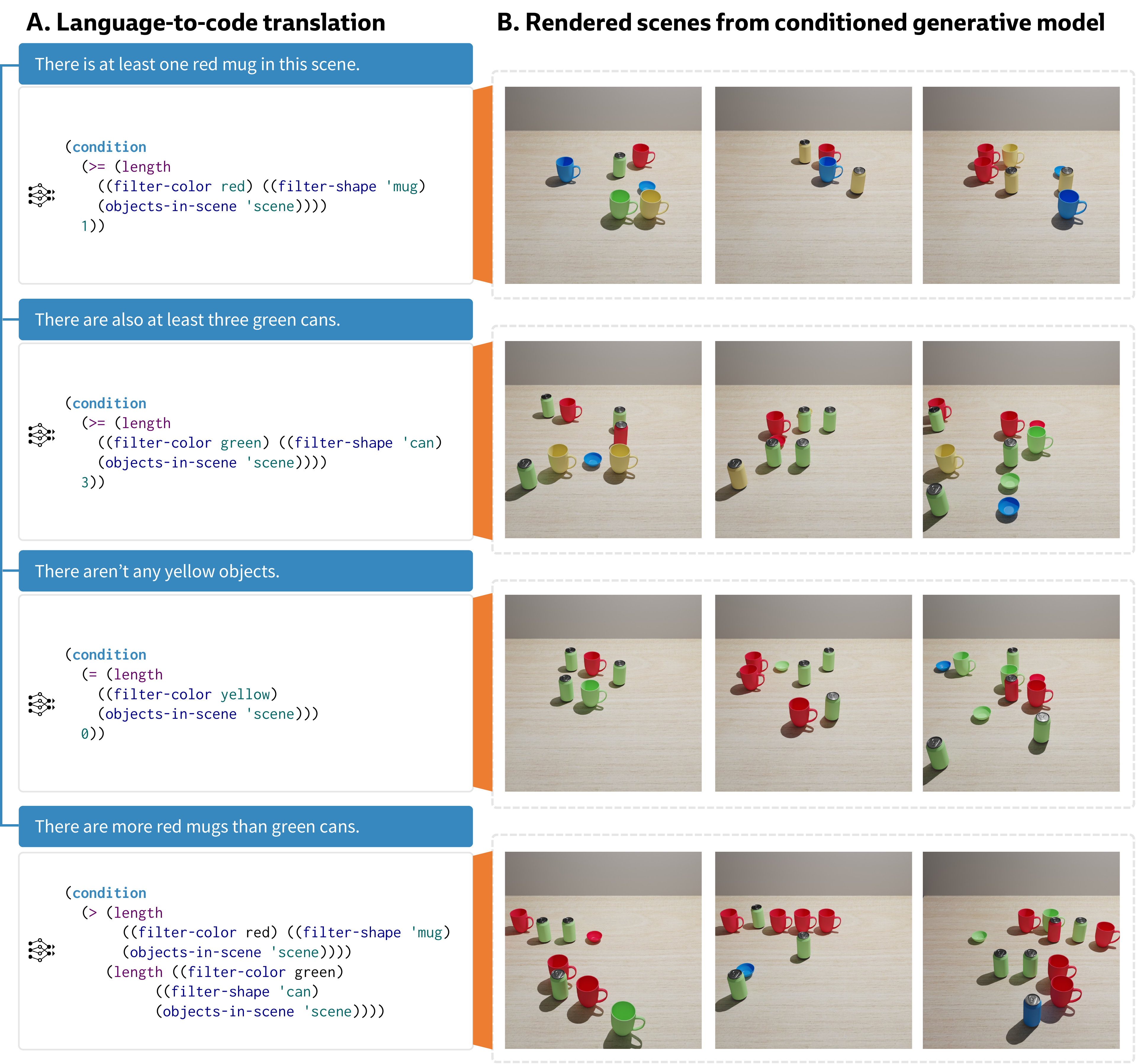}
    \caption{Each sentence in this sequence (\textit{left}) translates into a separate, composable condition expressions that updates the underlying generative model over scene states. After each sentence, sampling symbolic scene states from the updated distribution and \lstinline{render}-ing them (\textit{right}) yields images that reflect the prior over scenes and are consistent with the information in all successive sentences.}
    \label{fig:ref-grounding-translation}
\end{figure}

\paragraph{Grounded meanings as program expressions.} By augmenting probabilistic generative models with a graphics rendering engine, we have now extended our framework to allow \textit{language} that describes and asks questions about scenes to interface with visual depictions of those scenes.

In our simple tabletops scene domain, for instance, we can ground linguistic descriptions of the number, kinds, and colors of objects in a scene (\cref{fig:overview-scenes}, \textit{blue}) like \textit{there's at least two green cans} or \textit{a few mugs and bowls} into probabilistic program condition statements on scene states in the generative model. As with preceding sections, the translations shown in \cref{fig:overview-scenes} are quite straightforward and interpretable, because the generative model we have defined expresses compositional predicates on object properties at the grain of language. Constraints on objects of specific types, like \textit{green cans} are translated into a sequence of conditions on the relevant properties of object entities, successively filtering on the set of objects that are green (\lstinline{filter-color green}) and then further filtering to the set of objects that are also cans (\lstinline{filter-shape 'can}).\footnote{In our implementation, which can be found in \cref{appendix-sec:models-static-scenes}, we derive named color predicates like \lstinline{green} over the base generative model, which samples color properties over a continuous space of RGB values. This implementation suggests a more general point---that any number of lexical concepts, such as many more arbitrary color names over the underlying color space, can be derived as symbolic predicates over a richer continuous space reflected in the generative model. A similar approach could be taken for other lexical terms that carve up continuous spaces, such as prepositions like \textit{left}, \textit{center}, or \textit{near} over geometric space.}

Sampling scene states from the conditioned generative model, and rendering these scenes into images with the  \lstinline{render} interface, then produces visual depictions that are consistent with any sequence of observations made in language. This approach disentangles reasoning, as probabilistic inference over a structured generative model, from the perceptual properties of scenes. As with before, we can translate questions like \textit{How many green mugs do you think there are?} into probabilistic query expressions. Our approach reasons about these questions as inferences over the distribution of possible scenes, adapting beliefs about the scenes to condition systematically and coherently on sequences of new statements made in language.

\paragraph{Translating from language to program expressions.} As with the previous sections, we can now translate actual descriptions and questions about scenes, by using a large language-to-code model conditioned on the generative domain model and a few example pairs of language and code (see \cref{appendix-sec:models-static-scenes} for the full prompt we provide to condition the language-program model).

The translations in \cref{fig:overview-scenes} and \cref{fig:ref-grounding-translation} generally showcase the local generalizability and flexibility we illustrate in the other sections---the translation is robust to conjunction and syntactic variation, differing numbers of object predicates (\textit{yellow object}, \textit{red mug}), compositions of object predicates (eg. \textit{a few mugs and bowls}), negations over set quantity (\textit{there aren't any}), and comparatives over object sets (\textit{more red mugs than green cans}).

Even on this relatively simple domain, \cref{fig:overview-scenes} and \cref{fig:ref-grounding-translation} also showcase ways in which the LLM can represent conditional inferences from language to program expressions that go beyond simple, literal semantic meanings. These examples build on what we already find in \cref{sec:models-probabilistic-reasoning}, in which the LLM can contextually interpret vague language like \textit{very strong} as thresholds on continuous variables in the generative world model.

In this domain, find the LLM can translate vague quantifiers (like \textit{few}, \textit{most}, \textit{aren't a lot}, \textit{a bunch}, or \textit{aren't many}) without explicit program predicates defining each lexical term---the model can directly translate these terms into reasonable, interpretable quantities over sets of objects (such as translating \textit{only a few} to \lstinline{(<= 3)} objects). We also find that sampling from the distribution over meanings further supports the idea that the LLM represents a broader distribution over intended meanings, including acceptable semantic \textit{variation} in the interpretation of vague lexical terms. Sampling from the distribution at higher temperatures, for instance, we find that our implementation variously translates \textit{most} into program expressions that interpret this as \textit{more than half}, or \textit{more than 80\%}, or other greater fractions, of the set of overall objects in the scene. These translations draw on the language-to-code model's background prior on language itself (we do not prompt it with examples of these particular phrases), its amortized understanding of how these phrases relate to continuous, named variables in code (like \lstinline{length} of a set of objects), and the particular context of the generative world model itself (which defines the prior over numbers of objects that determines the context-specific scale of these graded quantifiers.)

Translations of vague quantifiers like these have been handled in classical semantics and recent accounts as explicit, pragmatic and probabilistic inferences based on context-specific priors---the acceptable quantity most people would infer for \textit{many} mugs on a table is intuitively very different from the quantity intended by \textit{many} grains of sand \citep{edgington1992validity,edgington1997vagueness,graff2000shifting,lassiter2017adjectival}. The results we show here provide further evidence that LLMs can often amortize many of these inferences, to directly predict common interpretations from language. As we discuss in \cref{sec:open-questions}, future work might explore more fluid, joint integrations of these approaches to inferring meanings, trading off between the amortized interpretations the LLM can produce and more explicit probabilistic inference, such as conditioning on other information in language. Learning that \textit{Sally is a wholesale porcelain supplier who owns thousands of mugs in a nearby warehouse} might lead you to infer an updated meaning of \textit{Sally has many mugs}, but is a complex inference that we might not expect to be amortized in an LLM from the background distribution of language and commented code.

\paragraph{Putting it together: Reasoning from language about visual scenes.} Taken together, the examples in \cref{fig:overview-scenes} show how this approach naturally extends the components of this framework---the ability to describe possible worlds in language, flexibly updating a background distribution of beliefs within a conditioned generative model, and query this model to draw probabilistic inferences---to also ground out in visual scenes. The more extended example in \cref{fig:ref-grounding-translation} highlights the more granular, intuitive way in which the distribution over scenes changes to reflect successive new sentences, updating a flexible distribution over scenes that still remains consistent with all of the previous observations.

\subsubsection{Language about dynamic physical scenes}\label{sec:models-dynamic-scenes}

\begin{figure}[ht!]
    \centering
    \includegraphics[width=\textwidth]{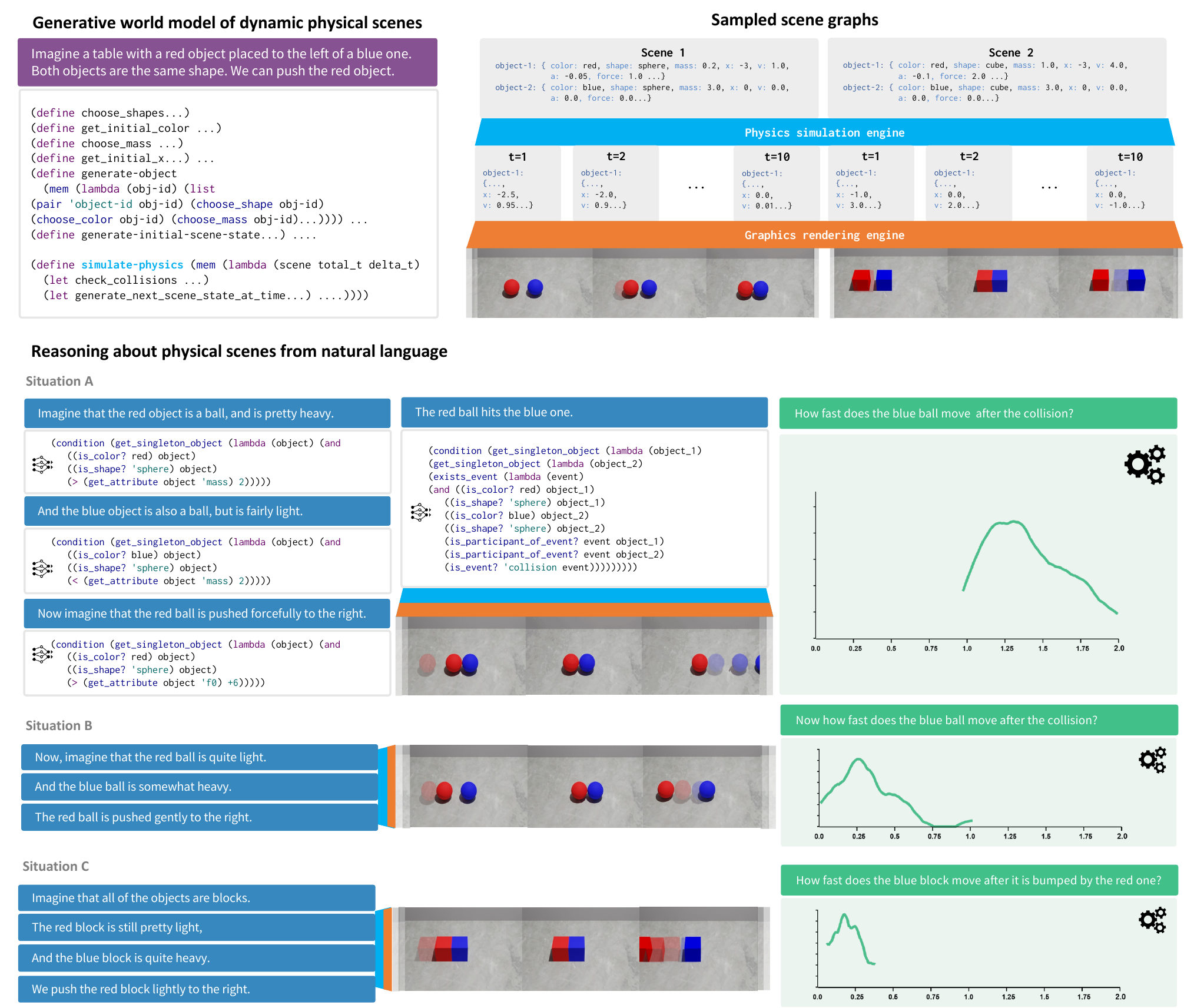}
    \caption{The way we talk about the world also draws on our intuitive physical knowledge. (\textit{Top}) A probabilistic generative model describes a prior over tabletop scenes with a red and blue object placed side by side, of varying mass and shape. Integrating a \textit{physics simulation engine} into this generative model allows this model to express a prior over dynamic scenes, modeling how each possible scene unfolds over time as differing initial forces are applied to the red object. (\textit{Bottom}) Language about possible physical scenes can again be translated into conditions (\textit{blue}) and queries (\textit{green}) on the distribution over dynamic scenes. Rendering these scenes produces scenes that reflect these conditions, and inference over the simulations allows the framework to answer queries contingent on the properties of the objects described in language.}
    \label{fig:physics-overview}
\end{figure}
When we talk about a scene, however, we describe more than just the colors and shapes of objects sitting on a table. We talk in verbs, describing events unfolding in the changing, physical world around us. Consider, for instance, descriptions of another set of tabletop scenes---ones that just involve a red object placed to the left of a blue one on a table (\cref{fig:physics-overview}). These scenes are initially even simpler than our tables of cans and dishware, but still afford a range of dynamic and physics-specific descriptions.

You can easily imagine, for instance, what would happen if someone \textit{pushed the red ball gently to the right}---you might say that it would \textit{bump into the blue ball}, and you could likely imagine \textit{how fast the blue ball would be moving} as a result. You can infer how these scenes would change if someone \textit{pushed the red ball much harder}, as if shooting a billiard ball, or \textit{tapped it even more gently}, nudging it forward with their finger, so that perhaps it wouldn't collide with the blue ball at all. These inferences are sensitive to many other properties of the scene, and of these objects, that we could describe in language, like whether \textit{the red ball is really heavy}, or \textit{the blue ball is very light}, or at least \textit{much lighter than the red one}. If we changed the objects in the scene, and now \textit{placed a red block to the left of the blue one}, your intuitive understanding of how different shapes relate to different degrees of friction would again change how you might see these scenes play out in your mind, and how you might answer questions about their collision and motion.\\

As adults, we have a deep, general understanding of how physical objects move and behave, and extensive developmental evidence suggests that well before we acquire language, we understand many core physical principles that govern our world \citep{spelke1990principles,spelke1995development,baillargeon2004infants,hespos2008young,rips2015divisions,teglas_pure_2011}. A productive line of computational cognitive models, in turn, has modeled human physical understanding as probabilistic inferences over a \textbf{mental physics engine}, modeled as programmable physics 
 simulation engines \citep{battaglia2013simulation,de2018end,yi2019clevrer,lake2017building,ullman2017mind} like those used in video games, computer animation, and robotics \citep{coumans2016pybullet,todorov2012mujoco,erez2015simulation}. 

As with the previous example on visual scenes, our goal in this section will be to illustrate how the overarching framework we have described in this paper can integrate language with other domains of human reasoning---perception and visual imagination, or intuitive physical reasoning. By translating language into a probabilistic language of thought, we can relate the semantics of language to these other, well-studied computational and cognitive modeling approaches, using probabilistic programs as the underlying interface between language, inference, and these engines for perception and physical simulation.

This approach is closely related to other recent work from the AI literature, most notably \cite{liu2022mind}, which also extends large language models with physics engine to ground natural language in physical simulation. By incorporating an interface to physics within a general probabilistic programming language, we show here how these approaches can model the commonsense, probabilistic judgements we make about everyday physical language---including with respect to uncertainty and vagueness in language about the underlying world state, or combined with inputs from visual reasoning, as discussed in the prior section.

\paragraph{Integrating the probabilistic generative model over scenes with a physics engine.} To model language about the example scenes we described here---red and blue balls, or blocks, placed on a tabletop (\cref{fig:physics-overview})---we implement a probabilistic generative model that is similar by design to the previous visual scenes domain (a very short excerpt appears in \cref{fig:physics-overview}, and the full model appears in \cref{appendix-sec:models-dynamic-scenes}). This generative program describes a prior over the possible properties of the objects initially set on a table, modeling scenes as a collection of objects in which each individual object is again generated (\lstinline{generate-object}) based on stochastic choices over its possible properties (e.g \lstinline{choose_shapes}). In this domain, however, we also model an explicit prior over the physical properties of each object, such as its mass (\lstinline{choose_mass}), and the relationship between shape and friction (as a simple \lstinline{get_friction_constants} function returns different constants, with a higher constant for blocks than spheres). 

As with the visual scenes example, each sample from this generative model again returns a structured, symbolic representation of a possible initial scene state, as a list of object entities that represents each object as a dictionary-like mapping from attribute kinds. This dictionary also stores each object's initial kinematic state, such as its \textit{position}, \textit{velocity}, \textit{acceleration}, and any \textit{forces} applied to it. To model the various ways we can \textit{push} the objects around, our generative model over scene also implements a stochastic function over possible initial forces (\lstinline{choose_initial_forces}) applied to an object. 

To model how each possible world unfolds as a dynamic scene over time, we implement a \lstinline{simulate_physics} function (\cref{fig:physics-overview}) that integrates the basic functionality of any programmable physics engine into the probabilistic model---this function takes in a scene state that specifies the relevant physical properties of objects, and returns a sequence of scene states forward simulated in time under the laws of physics. In our implementation, this sequence is a list of scene states at each timestep, each which contains its own set of the objects and their properties with relevant kinematic properties (like \textit{position}, \textit{velocity}, \textit{acceleration}) updated at each timestep. The physics model we use in our example is simple enough that we implement it fully within the body of the probabilistic program itself (see \cref{appendix-sec:models-dynamic-scenes}) for illustrative purposes---our \lstinline{simulate_physics} updates each object at each timestep under the basic kinematic laws of Newtonian mechanics, includes a simple implementation of static and kinetic friction under gravity, and models simple collisions as impulse exchanges in momentum. 

The rendered simulations we show in \cref{fig:physics-overview} also showcase the interplay between these modular, API-like interfaces integrated into a probabilistic language of thought---combined with the \lstinline{render} interface from the previous section, we can not only simulate underlying physical scene states, but visualize them by rendering each individual scene state in the sequence over time. Collectively, this model now captures a prior over tabletop scenes, models how any given scene in the distribution unfolds dynamically under physics, and captures how each scene appears visually over time.

\paragraph{Grounding physical language in program expressions.} By extending the underlying probabilistic world model to interface with a physics engine, we can ground the semantics of language about the physical world in intuitive, human-like physical reasoning modeled by the physics simulation engine over world states.

Descriptions of the physical properties of objects, for instance, like \textit{the blue ball is not very heavy} (\cref{fig:physics-overview}) translate into conditions on the \lstinline{mass} property of an object in the world state, and maintain uncertainty inherent to language---phrases like \textit{very heavy} translate into conditions that threshold a continuous distribution of possible masses. As in the visual scene example, sampling from the conditioned generative model produces dynamic scene simulations that reflect language. Descriptions of \textit{heavy blue balls}, or \textit{red blocks that are relatively light}, or scenes in which a \textit{red ball is pushed forcefully}, or in which \textit{a red block bumps into a blue one}, all connote sets of scenes that model explicit, physical simulation. In turn, queries about distributions over physical scenes (like \textit{how fast a heavy blue ball will move after it is bumped}) reflect probabilistic inferences that condition on all of these relevant descriptions in language, estimated by sampling and running physical simulations over the possible world states.

In this example, we highlight an approach to translating verbs and descriptions of physical events (\textit{the red ball pushed forcefully to the right}, \textit{the red ball hits the blue ball}) that grounds them directly over continuous variables in our world model. In \cref{fig:physics-overview}, for example, our implementation translates \textit{pushed forcefully to the right} into a condition expression that picks out a distribution of initial forces, over a space of continuous force vectors with direction and magnitude, as the meaning of \textit{push} in a physical 
world. Similarly, we translate \textit{hits} with respect to collisions simulated by the physics engine between the two object entities.

In our appendix, however, we also implement and show how a discrete event semantics can also be constructed over variables in the physics engine, to highlight potential future connections between our implementation and more classical event semantics representations. Neo-Davidsonian event semantics and related approaches \citep{parsons1990events,davidson1967logical}, for instance, have long modeled events in language with discrete \lstinline{event} entities and lexical event predicates (eg. \lstinline{is_hitting}) that describe particular categories of events in time.  Prior work in classical event semantics has also considered how discrete event representations relate to underlying physical forces \citep{talmy1988force}, with particularly close connections to lexical semantics approaches \citep{talmy1988force,levin1993english,schuler2005verbnet,pinker1984language,jackendoff1985semantics} that realize verb meanings into cognitively-grounded physical concepts of motion and forces. 

Our implementation concretely realizes these semantic events and predicates as functions derived entirely on top of a fully realized, continuous world state modeled in a physics engine---\lstinline{is_hitting}, for instance, is an event derived on top of the collision mechanics in the underlying physics engine. Other event predicates, like \lstinline{is_moving}, or \lstinline{is_resting}, can be similarly as thresholds on continuous kinematic properties (here, \textit{velocity}) represented in the world state.  Our broader goal is to show all of these can be broadly constructed over a probabilistic language of thought, which grounds out concretetly with respect to states in an implementable physics engine.

\paragraph{Translating from language to program expressions.} As with our visual scenes example, the translations we show in \cref{fig:physics-overview} are again chosen to illustrate the generalization and amortized inferences that the language-to-code LLM can make. Much like vague quantifiers, we find that the context-conditioned LLM can directly infer reasonable meanings for graded terms that pick out thresholds over a numeric distribution---translating phrases like \textit{not very heavy}, \textit{pretty heavy}, and \textit{pretty light} directly into reasonable, context-specific thresholds on continuous masses, or \textit{pushed gently} and \textit{pushed forcefully} into thresholds on forces. Again, we see interesting future grounds for further integrating these kinds of amortized inferences with more explicit, probabilistic inference mechanisms for deriving them---such as to integrate inferences over language with new contextual observations from other modalities, such as perceptual or motor observations from seeing or actually moving these objects that might update one's background beliefs over the distribution of masses.

\paragraph{Putting it together: Probabilistic inference and physics simulation from language.} The examples in \cref{fig:physics-overview} show how this approach can capture the nuanced relationships between language and physical reasoning. Language that modulates any of the physical properties in our introduction to this section, from the masses of objects, their shapes and corresponding friction when moving, and the forces they receive, changes the distribution over internally simulated scenes, and is reflected in updated inferences about downstream events. 

\paragraph{Future directions: Perception as inverse rendering and complex physical reasoning as intuitive phsyics.} As with all of our other examples, its important to emphasize that our \lstinline{simulate_physics} interface is almost the simplest possible world model we might construct over physical scenes. The approach we take here is inspired by, but much simpler than, many other probabilistic generative models \citep{battaglia2013simulation,wu2015galileo,ullman2017mind,allen2020rapid,xu2021bayesian} of more complex object configurations in more complex environments (such as \textit{ramps, stacks of objects}), many other properties that we can describe about objects themselves (such as their \textit{material}), and arbitrary forces (like \textit{bumping} the table or \textit{dropping} objects from above). 

Our approach in these sections also suggests a rich line of future work for reasoning jointly about observations in language, and from perception. While we do not implement a perceptual module in our example, the framework we sketch here can be directly integrated with the broad body of work on \textit{inverse graphics}, which frames scene understanding as inference from observed visual inputs to recover structured representations of a scene's contents \citep{kersten1996introduction,kersten2004object,yuille2006vision,lee2003hierarchical,wuGalileoPerceivingPhysical2015,yildirimEfficientInverseGraphics,wu2017neural,yi2018neural}. Our framework suggests a particularly fruitful integration between language and the growing body of work that combines probabilistic programs and graphics rendering engines \citep{kulkarni2015picture,mansinghka2013approximate,gothoskar20213dp3,zinberg2019structured}. To draw inferences about visual scenes from perceptual inputs, models like these incorporate convolutional neural networks to make fast, amortized proposals about the scene state from vision, but with respect to a generative program that defines the underlying scene state and guide inferences about particular scenes, such as to reason about occlusion. 

Integrated with the approach we describe here, this framework could ground linguistic queries directly into vision, allowing structured inferences for visual question answering (e.g., \textit{counting the number of unique colors of the dishes in a scene}). Moreover, it could enable more complex, joint inferences that integrate visual observation with linguistic information about latent physical properties of objects in a scene (e.g., mass, friction) or the presence or identity of occluded objects. Such multimodal integration holds the potential to shed further light on the ways that linguistic knowledge can shape our understanding of and reasoning about physical scenes.

\clearpage
\subsection{Language for reasoning about agents and plans}
\label{sec:models-agents-and-planning}
One of the most deeply human things we can talk about is other people. To conclude this section, we turn to language about other social beings---agents who want things, chase goals, and plan how to act in the world around them.

\begin{figure}[h!]
    \centering
    \includegraphics[width=\textwidth]{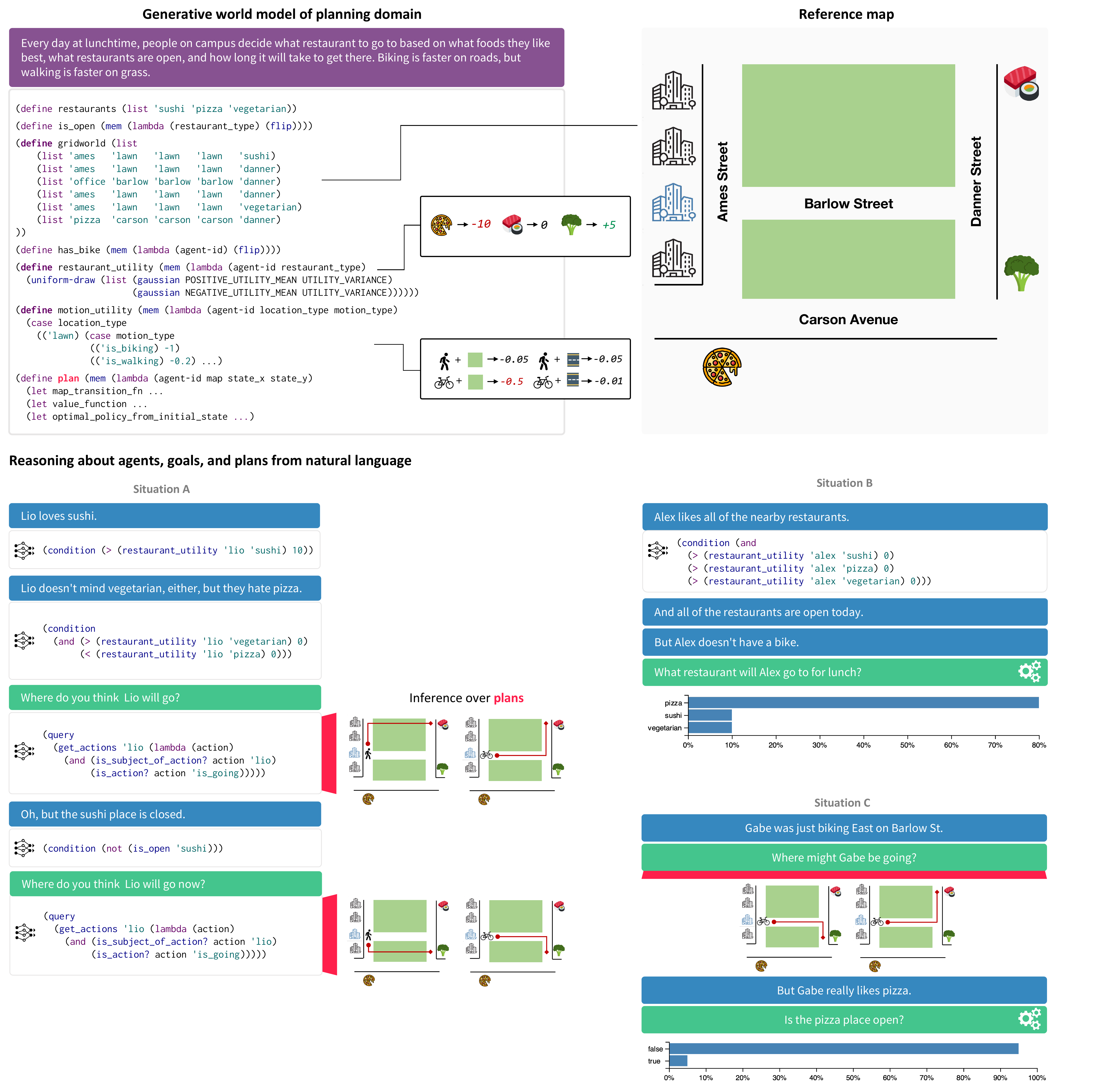}
    \caption{Our language about other people builds on our intuitions about how agents act on their preferences and goals. (\textit{Top}) Our example probabilistic generative model describes a prior over agents with different preferences for the nearby restaurants shown on the map, as well as the relative cost of getting to each one on bike or on foot. Integrating a \textit{model-based planner} into this generative model allows it to express a prior on how agents will actually act based on their desires, balancing these preferences against whether restaurants are open, and whether or not they have a bike. (\textit{Bottom}) Observations and queries about the agents, their goals, and about the world itself updates a unified belief distribution, reflecting how agents plan in the world and how observing their actions drives inferences about the latent state in the world.}
    \label{fig:overview-planning}
\end{figure}

As an illustrative example, we consider a domain (\cref{fig:overview-planning}) inspired by \cite{baker2007goal}, which evaluated commonsense social inferences about agents with different preferences and goals. In our slightly modified example, we consider a set of people with varying food preferences who are making plans for lunch. Based on the map shown in \cref{fig:overview-planning}, we’ll imagine which restaurant they might go to based on what foods they like, how far each restaurant is from their office (shown in blue), and whether restaurants happen to be open or closed. We’ll also note that students can bike or walk to any restaurant, but include the intuitive fact that biking is faster on roads, but slower than walking on the lawns. 

The original experiments in \cite{baker2007goal} used visual stimuli to depict agents’ paths and plans, but language is a particularly natural and nuanced way to communicate information about other agents. Consider the range of situations we can describe in this simple example. We might leverage our wide vocabulary for describing the spectrum of someone’s preferences and desires---whether they \textit{crave pizza} or \textit{hate vegetables}, or whether they \textit{love sushi} rather than merely \textit{liking it}. We might describe their more concrete, discrete goals, like \textit{getting to a pizza place} or generally \textit{getting to the closest restaurant to the office}. The inferences we draw from language also depend on our intuitions about agents themselves. All else being equal, we expect people to minimize the effort it takes to act, while trying to maximize the value they gain from acting. We might generally expect someone to \textit{walk down Ames Street} if they wanted to go to the pizza place rather than taking an unnecessarily convoluted path, or to \textit{jump on a bike} if they owned one, rather than taking a slower walk there. We also understand, of course, that people need to accommodate the world itself in their plans, and might not go to the pizza place no matter how much they love pizza, if they were told that \textit{the pizza place was closed.}

Perhaps more subtly, but equally importantly, what we know about agents also informs the wide range of inferences we can draw from language about their \textit{actions}. Consider, for instance, what you can infer from being told that someone had started at the office, and was now \textit{walking across the southern lawn}. Because they’re on a direct route towards the vegetarian place, you might infer that they are more likely to prefer vegetarian food, and that they either know or at least believe that the vegetarian place is open. Because they are walking on foot, you might also suspect that they do not own a bike, which would have allowed them to get to the restaurant more quickly. All of these inferences build on a cohesive picture of agents as a whole---our expectations about agents as goal-directed, efficient actors inform how we think about any given action. 

As with visual and physical reasoning, this section builds more generally on extensive work in cognitive science and artificial intelligence on \textit{social reasoning}, and seeks to integrate this broader literature into our framework for language. Developmental evidence suggests that we have a core conceptual understanding of agents as goal-directed actors from a very young age \citep{csibra2003one,csibra2008goal,scott2013infants,spelke2007core}. Computational cognitive models, and the broader AI planning literature, have long approached social inferences like those we describe here under under a unifying model of \textbf{planning} and \textbf{inverse planning} \citep{baker2007goal,baker2009action,baker2011bayesian,jara2020naive,cusumano2017probabilistic,seaman2018nested}. This framing couples the forward planning of actions that achieve goals or maximize utilities, to the inverse problem of inferring latent variables about the agent or the world from observations of their actions.

The example below shows how we can extend the modeling motif from our previous discussion of visual and physical reasoning, which shows how our framework can relate language to other core cognitive modules via interfaces implemented in a general probabilistic language of thought. In this section, we introduce \textit{model-based planners} as another such core computational module, which can be integrated into this framework to support a wide range of probabilistic forward and inverse inferences about agents and their actions as they are referenced in language.

\paragraph{Integrating a probabilistic generative model over agents with a planner.} As a concrete example, the generative program excerpted in \cref{fig:overview-planning} (shown in full in \cref{appendix-sec:agents}) illustrates how an integrated probabilistic modeling and planning language can describe the agents and restaurants domain.

To model background states of this environment, our implementation represents the spatial structure of the campus (as a simple \lstinline{gridworld} map), and stochastic Boolean variables that model whether someone owns a bike (\lstinline{has_bike}) and whether a given restaurant is open (\lstinline{is_open}).

We then introduce a generic, utility-based formulation derived from the classical AI planning literature to model the varying preferences of any given person and the effort of taking particular actions (see \cite{russell_norvig_2021} for a review). Incorporated into a probabilistic generative model, we can express the distribution of preferences any particular agent could have, and the way these preferences interact with the stochastic mechanics of any given world. In our implementation, we model these varying preferences as a stochastic utility function associated with particular agents and restaurants (\lstinline{restaurant_utility}). Our implementation shows a bimodal Gaussian distribution, in which people tend to have distinctly negative or positive preferences for any given restaurant, but any other formulation would be easily expressible. We also model how these preferences interact with other aspects of the world---we condition the value an agent derives from actually arriving at a restaurant (\lstinline{utility_at_restaurant_state}) on whether or not it is open. These utilities interact with possible actions an agent can take to get to different restaurants. We model the distribution of possible actions an agent might take (our \lstinline{available_actions} function conditions on whether an agent \lstinline{has_bike}), and the varying effort of individual actions. Our \lstinline{motion_utility} conditions on the type of action and the state in which it is used, to model the greater effort of biking on grass and the relative ease of biking on the road.

Up to this point, the generative model simply expresses a general prior over world states that includes agent preferences. Now, to model how an agent actually decides on a course of action conditioned on the world state, we can finally introduce a \lstinline{plan} interface (\cref{fig:overview-scenes}) that calls out to a model-based planner. Our implementation, while simple, implements the basic functionality core to an AI planner---it computes a sequence of actions that achieves a goal or maximizes a value function, subject to an agent's underlying preferences, available actions, and the conditions of the environment. As with the physics interface in our previous section, our example planner implementation is simple and generic enough that we also implement it fully within the body of the probabilistic program itself (see \cref{appendix-sec:agents}) for illustrative purposes. Our implementation here uses a simple value-iteration algorithm, which computes an optimal policy of action to trade off between the value an agent derives from particular restaurants, and the cost of taking actions (walking or biking in any given direction, from any location in the map) towards them.

\paragraph{Language about agents as program expressions.} By augmenting the probabilistic generative model with a planner, we can now ground many of the basic ways we talk about agents themselves in probabilistic conditions and queries to this model.

Language about what people want and prefer, like whether someone \textit{wants}, \textit{likes}, \textit{loves}, \textit{doesn't mind}, or \textit{hates} a given restaurant in our example domain, can construct formal conditions over underlying utility variables, that in turn drive inferences about how the agent will act. In the examples shown in \cref{fig:overview-scenes}, we illustrate the semantics of these terms as conditions constructed directly over the continuous utility variables defined in this domain. We could also derive a more explicit set of predicates (like a Boolean \lstinline{likes?} predicate, defined over the underlying utilities), but as in several previous sections, we show these more transparent semantics (like translating \textit{likes} into a \lstinline{> 0} threshold on utilities) to illustrate how language relates to our model of agents, and to demonstrate the amortized inferences that a language-to-code model can make in directly inferring these threshold values in context, and for new preference words.

Observations about relevant aspects of the environment, like whether the \textit{sushi place is closed} or \textit{Alex has a bike}, are translated as in previous sections into conditions on the generative world model. In this integrated framework, these observations now support downstream inferences about how agents might change their behavior with respect to what we are told about the world.

Finally, of course, explicit observations and queries about someone's goals, plans, and individual actions (\textit{Gabe was biking East on Barlow Street}, or \textit{What restaurant will Alex go to for lunch?}) can be interpreted with respect to the underlying, model-based planner, to drive inferences about forward planning agents choosing actions in the world, and inverse inferences over the many latent variables in the world that collectively explain language about someone's actions.

\paragraph{Translating language using language-program distributions.} We showcase several distinct examples (\cref{fig:overview-planning}) of context-sensitive, pragmatic inferences derived using a language-to-code meaning function conditioned on language and this generative world model.

As in previous sections, we find that the LLM can directly ground vague, graded terms in context-specific thresholds over particular continuous variables in the probabilistic world model. Here, this approach grounds preference terms (\textit{doesn't mind}, \textit{loves}) into reasonable thresholds over the utility variables in the world model (\cref{fig:overview-planning}). We find that the LLM can both infer reasonable utility thresholds and generalize to words not explicitly given as example translations: we prompt the model with a handful of examples pairs, such as a translation that maps the word \textit{likes} to a \lstinline{> 0} threshold on utilities, and the LLM successively generalizes this parse to ground other preference terms like \textit{hate} and \textit{love}, presumably based on the comparative valences of these preference terms in the broader distribution of language.

We also find that the LLM can directly translate quantifiers over contextual sets in this domain---like \textit{likes all of the nearby restaurants}---into a conjunction over the set of restaurant literals in this domain, by conditioning on the generative world model during parsing. More concretely, this means the LLM identifies the relevant \lstinline{restaurants} list (shown in the excerpted generative world model in \cref{fig:overview-planning}), and conditions on it to directly produce the unrolled conjunction over the list contents (\lstinline{(and (is_open 'sushi) (is_open 'pizza)...)} intended by \textit{all restaurants}, amortizing the computation that would have otherwise been necessary over a more literal semantics like \lstinline{(all restaurants)}. Together with the previous sections, these examples suggest how our framework might jointly support explicit inferences from language into various expressions in a language of thought, and learned patterns from the large language-to-code model that amortize some of these inferences over time, which we discuss directly as grounds for future work in \cref{sec:open-questions}.

\paragraph{Putting it together: Probabilistic inference and planning over language.} The several example dialogues shown in \cref{sec:overview} show how this approach captures the integrated social inferences we make about agents in language. We can now query the plans and goals of agents, deriving inferences with respect to the \textit{forward planning} module incorporated into the underlying generative model, conditioning flexibly on arbitrary information in context, and updating expectations about where agents will go, and how they will change their plans based on new observations about the world. In turn, we can derive \textit{inverse planning} inferences, like whether \textit{the pizza place is open}, based on relatively tangential information about someone's actions---knowing that an agent really likes pizza, but is seen taking a path that wouldn't efficiently lead them there. All of these inferences fall out of the same underlying generative model, which unifies these distinct observations about people and the world in language with respect to a formal model of how agents tend to behave.

\paragraph{Future directions: Scaling integrated world models for planning and inference.} The \lstinline{plan} function in our example implements a very simple but \textit{model-based} planner---it computes actions based on an underlying, structured model of the world. In comparison to the other domains in this paper, linguistic planning and social reasoning have received perhaps the attention in recent work, in part because complex reasoning about other agents \citep{ullman2023large} and precise general planning tasks \citep{bubeck2023sparks,valmeekam2023planning} appear to pose outstanding challenges for even the largest current language models. Recent work has sought to interface large language models with classical planning languages and symbolic planners (eg. \cite{liu2023llm+,ding2023task,xie2023translating,collinsStructuredFlexibleRobust2022}), as well as general purpose programming languages used to express code-based policies \citep{wang2023voyager}. All of these approaches suggest directions for scaling the simple planning implementation we show here---our goal is to show how classical planning approaches can be nested within and integrated into probabilistic generative models to support a range of complex reasoning about other agents to infer their goals and actions from information in language.

Collectively, the broader cognitive science and AI planning literature suggests many directions for scaling up this model towards more of the nuance in human social reasoning, each which would in turn allow this paradigm to ground richer and more nuanced descriptions of agents, and the inferences we draw from this language. Some of the most important ones include planners and planning languages that designed to express explicit and discrete \textit{goals}, like \textit{wanting to be at the highest rated pizza place within a half-mile radius} or \textit{trying to get a plate of sushi for under ten dollars}, rather than continuous values and utilities \citep{davidson2022creativity,fikes1971strips,pednault1989adl,mcdermott1982temporal,mcdermott20001998}; planners that model explicit uncertainty about the world itself, like agents who \textit{don't know whether a restaurant is open or closed until they get there} \citep{baker2011bayesian,zhi2020online,kaelbling2013integrated}; hierarchical planners that recursively turn goals into more specific subgoals to account for plans over longer timescales, at differing levels of abstraction \cite{kaelbling2011hierarchical}; and recursive models of agents who are themselves thinking about other agents, such as models of \textit{two people trying to meet up} at a restaurant that they think will satisfy both of them, or where they might be most likely to find the other \citep{baker2011bayesian,krafft2016modeling,wu2021too}. Each of these could allow this paradigm to ground richer and more nuanced descriptions of agents, and the inferences we draw from this language.

\paragraph{Conclusions.} Together with the previous sections on vision and physics, our approach to grounding language about social agents highlights the more general computational account suggested by our framework. By translating language into probabilistic programs, language can construct, describe, and drive inferences over our internal world models. These may in turn incorporate many more specific computational engines---modeling how scenes are visualized, how physics unfolds in the world, or how agents plan towards their goals--- as modular interfaces that can be called upon in a general probabilistic language of thought.

\clearpage
\section[Growing world models]{Growing and constructing world models from language}\label{sec:growing-and-constructing}
In \cref{sec:language-and-world-models}, we illustrated how domain theories expressed in a probabilistic language-of-thought can provide flexible and powerful scaffolding for language understanding. In each, generative world modeling programs provided a unified substrate for defining a structured domain model and representing the meanings of sentences. But where do these world models come from? If we want our PLoT account of language understanding to scale beyond the knowledge that can be hand-coded by a programmer, we need to provided some account of how such a system might acquire new concepts and domain theories.

One of the hallmarks of human communication is our ability to \textit{teach} each other fundamentally new concepts in language. We coin new words, define interrelated conceptual systems, and describe entirely new world models, explaining the abstract underlying structure of whole domains. Because language spans so many aspects of human thought, it is perhaps a uniquely powerful tool for structuring learning. In language, we can define new concepts and domains that are integrated into our inferences, relational reasoning, understanding of the visual and physical world, and goals and plans.

How do we learn new concepts and world models from language? And how can we build computational systems that can be taught in language as we teach each other? In this section, we showcase the extensibility of the framework we have proposed as a unified model for relating language to thinking. Because world models in a PPL are expressed as programs, the same core computational components can be used to extend and construct world models themselves from language. In \cref{sec:growing-world-model}, we show how we can extend an existing domain model with new lexical concepts. Then, in \cref{sec:constructing-domain-theories}, we turn to language that communicates an entire background domain model from scratch. Through these simple explorations, we aim to point towards a near-term horizon where systems might construct rich and nuanced probabilistic models to make sense of their linguistic environments and the broader world around them.

\subsection{Growing a world model from language}\label{sec:growing-world-model}
How can we enrich our world models with concepts learned from language? Let's consider, for instance, the kinship domain model used in the relational reasoning example in \cref{sec:models-relational-reasoning}. The probabilistic program used in this example described a basic generative model over family trees, and then defined a handful of primitives, such as concepts for \lstinline{grandparent} and \lstinline{sibling}. But most people know and can talk about many more kinship relations than those included in that simple example, such as uncles, aunts, and cousins. What happens when we use language that invokes one of these undefined concepts?

\begin{Dialogue}

\Condition{Avery is Blake's uncle.}
\begin{lstlisting}
(condition
  (exists (lambda (x) (and
                        (sibling-of? x 'avery)
                        (parent-of? x 'blake)))))
\end{lstlisting}
\end{Dialogue}

\noindent The LLM proposes an initial translation that includes some of the important components in the concept of an ``uncle.'' However, several key details are not quite right: an uncle should be the \textit{brother} of Avery's parent, not just a generic sibling. Moreover, an uncle can come from outside the bloodline, in which case this definition would not fit. Much like a person learning English, the LLM has a partial notion of this concept, but could benefit from more explicit instruction from a knowledgeable teacher. In this section, we introduce a new \lstinline{define} construct that does just this by prompting the LLM to generate a new definition from language.

\begin{Dialogue}[colorDefine]
\Define{An uncle is the brother of one's father or mother, or the husband of one's aunt.}
\begin{lstlisting}
(define (uncle-of? name_a name_b)
  (or (exists (lambda (x) (and
                            (brother-of? name_a x)
                            (parent-of? x name_b))))

                            
      (exists (lambda (x) (and
                            (husband-of? name_a x)
                            (aunt-of? x name_b)
                            )))))
\end{lstlisting}
\end{Dialogue}

\noindent We've used \lstinline{define} to fill in a bit of common knowledge that was missing from our conceptual system. But the mental frameworks we use to reason about the world are constantly undergoing conceptual change, both at an individual and a societal level \citep{carey1999sources, posner1982toward}. For instance, shifts in cultural notions of gender and identity have introduced new kinship terms into English. One of the hallmarks of language is the ease with which we can coin and communicate new concepts, like the following:
\begin{quote}
   \textit{``Pibling'' is a gender-neutral term for ``aunt'' or ``uncle'' that refers to the sibling of one’s parent.}
\end{quote}
\noindent Finally, as we touched on in \cref{sec:models-relational-reasoning}, kinship systems vary widely; certain cultures have kinship concepts that are more granular than those found in English. For instance:
\begin{quote}
    \textit{In the language of the Northern Paiute, a group of peoples indigenous to the Great Basin region of the US, ``pāan'i'' refers specifically to the sister of one's father.}\footnote{At the time of writing, a Google search for the term ``pāan'i'' yielded zero results. The term itself was pulled from a non-searchable table in a century-old manuscript \citep{lowie1930kinship}. As far as real-world kinship terms go, it is comparatively unlikely---though not impossible---that ``pāan'i'' was part of Codex's pretraining data.}
\end{quote}
From this definition, we can incorporate the concept of a \textit{pāan'i} into our growing set of kinship concepts. Our framework elegantly captures this ability to learn new concepts in language that we can then use productively to construct new sentences and reason about coherently against the background of our existing world knowledge. Here, we walk concretely through how the basic components of our framework are combined to grow the original kinship model with new concepts.

\begin{figure}[htbp!]
    \centering
    \includegraphics[width=\textwidth]{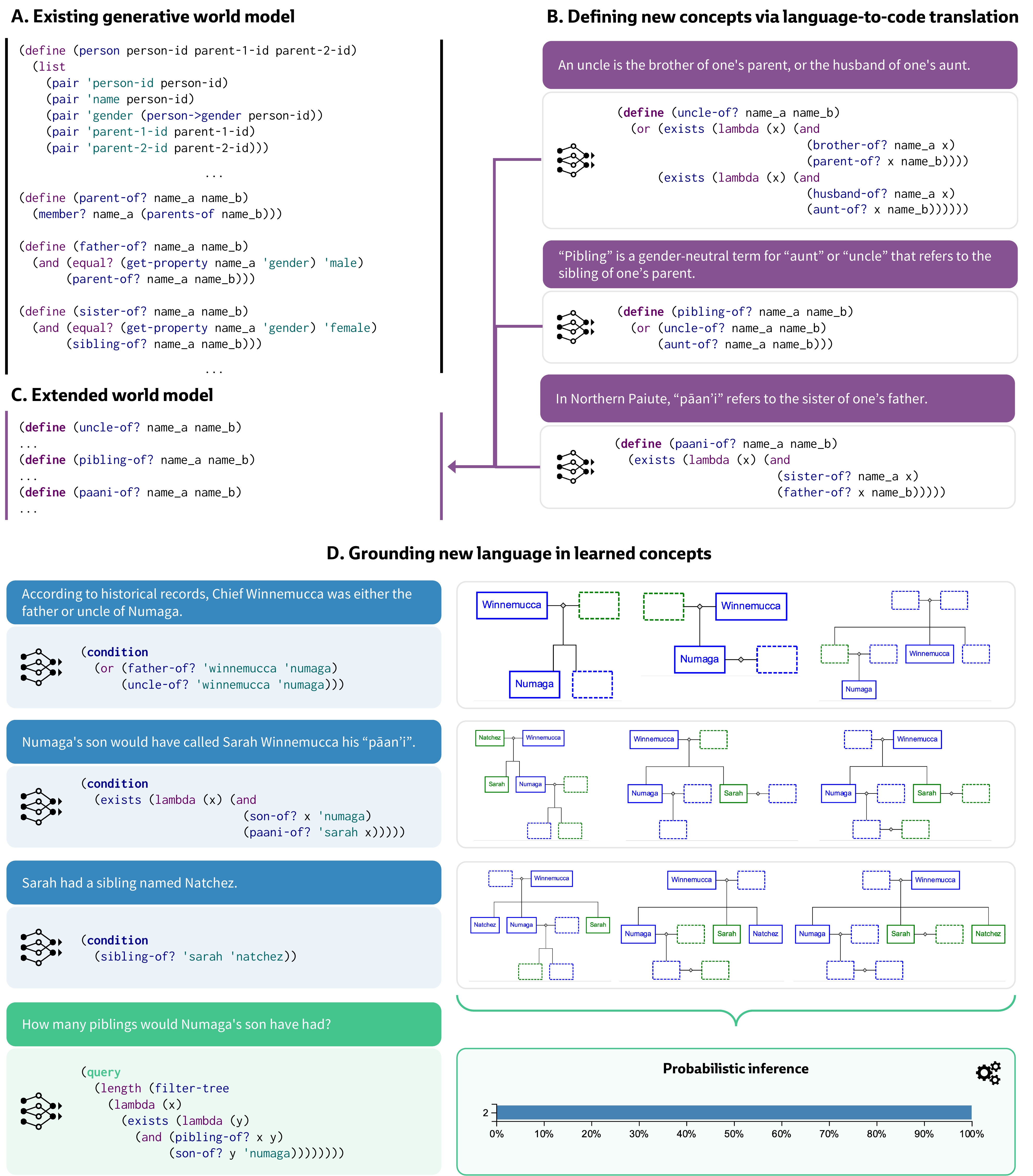}
    \caption{Extending the kinship world model with linguistic descriptions of kinship relations drawn from contemporary English and a low-resource language (Northern Paiute). A language-to-code LLM is prompted with (A) the existing generative model code and (B) language describing novel kinship relations to produce new concept definitions in Church. The extended world model (C) now supports probabilistic reasoning from language that contains these new concepts (D).}
    \label{fig:growing-translations}
\end{figure}

\paragraph{Linguistic meanings as program expressions.} Much as we interpreted observations as program expressions that conditioned an existing world model, and questions as program expressions that queried it, a sentence like, \textit{The term ``pāan'i'' refers to the sister of one's father}, can be modeled as a program expression that \textit{defines} a new such primitive relation, \lstinline{paani-of?}. The examples in \cref{fig:growing-translations} show how the semantics of this sentence, along with the other kinship concepts introduced in the introduction to this section, can be similarly understood as expressions that define new conceptual primitives. These expressions are particularly interesting because they are defined in terms of other concepts, like \lstinline{sister-of?} and \lstinline{father-of?}, that make up this conceptual system. In this way, our treatment of concept learning is closely linked to the idea of a \textit{conceptual role semantics} \citep{field1977logic,harman1982conceptual,greenberg2005conceptual,block1998conceptual}, in which concepts (including lexical concepts) derive meaning from their interrelated roles and relationships to other concepts. In these examples, interpreting these sentences as program expressions defined over the base generative model showcases the flexible role that the generative modeling program can play, in relation to language about the domain.
While our example showcases simple relational definitions over the underlying world model, it is worth noting that these are not the only kinds of functional definitions that we could learn to extend a world model from language. This general approach can be used to make meaning from sentences that grow an underlying world model in other ways, such as by defining new random variables (like phenotypic eye colors or other inherited traits) that extend the probabilistic generative model.

\paragraph{Translating with a language-program distribution.} While the meanings of these sentences play a different role in our framework---they extend the world modeling program, rather than condition or query it---they are still program expressions. Therefore, with minor adjustments, we can use the same language-to-code LLM approach to ground these new concepts in our world model. To derive each of the translations shown in \cref{fig:growing-translations}, we feed the LLM the same prompt as in \cref{sec:language-and-world-models}, which includes the existing generative model and example translations. The final line of the prompt begins with \texttt{Define:} and contains the language describing the new concept definition. Each sentence is then then translated into the new \texttt{define} statements which construct new conceptual kinship primitives. In sum, linguistic definitions are simply another kind of program expression we can translate into from language.

\paragraph{Growing the domain model with new program expressions.} Finally, by incorporating the meanings of sentences like \textit{The term ``pāan'i'' refers to the sister of one's father} back into the domain model itself, we have formalized a simple approach for enriching a world models with concepts learned from language. Each sentence shown in \cref{fig:growing-translations} is translated into a program expression that defines a new relational function which extends the set of conceptual primitives that comprise the extended kinship domain.

The more general principle here is not limited, of course, to kinship concepts. We could extend any of the domain models in each of our previous examples with new concepts learned from language. For example:

\begin{itemize}
    \item \textit{In tug of war, the strongest person on a team is referred to as the ``anchor''.}
    \item \textit{A ``monochrome'' scene is one in which every object is the same color.}
    \item \textit{On ``National Restaurant Day'', all the restaurants in town are guaranteed to be open.}
\end{itemize}

Our proposal in this section is closely related to other work which formalizes the learning of new concepts as the learning of new program components, such as program synthesis systems that bootstrap a growing library of domain-specific concepts constructed out of an initial programming language \citep{dechter2013bootstrap,ellis2020dreamcoder,bowers2023Stitch}; work that formalizes the learning of new concepts from language as the learning of new program primitives \citep{sumers2022talk,wong2021leveraging,shin2018program}; and semantic parsers that bootstrap lexicons of compositional word meanings, defined in a formal logical language, for interpreting new sentences \citep{cai2013large,artzi2014learning,kwiatkowski2011lexical}. 

The framing we describe here showcases the tight integration between language, meanings, and the probabilistic programs that form the formal substrate for modeling the world in our framework. Language that specifies new parts of a world model can be cleanly interpreted as program expressions, which are used to extend the generative world modeling program itself. These generative models in turn provide the basis for reasoning about new observations that build on these learned and structured bodies of conceptual knowledge. Returning to the themes of our introduction, human-like thinking, under the broader computational approach we take throughout this paper, is formalized as probabilistic programming and inference over probabilistic programs. This is how we construct models of and reason about the world. Language, then, is an especially powerful tool for constructing programs of all kinds---ones that condition and query existing world models, and ones that actually construct and extend the flexible domain models themselves that undergird linguistic meaning and thought.

\subsection{Constructing new world models from language}\label{sec:constructing-domain-theories}

So far, we have assumed that language understanding happens in the context of  a particular world model appropriate for the situation at hand, containing definitions of key concepts like \textit{sibling} for kinship reasoning, or \textit{strength} for reasoning about playground games. We have now seen how these models can be extended with new lexical definitions on the fly, but the question remains of where these background world models come from in the first place. The full answer to this question is likely complex: people learn about the world in all sorts of ways. But in some settings, people do seem to acquire new world models largely \textit{through} language: we read the rules of new games, are taught the workings of machines, and take classes on the causal structure of many other complex systems (the human body, the solar system, the government). In this section, we broaden our scope beyond language that conveys new concepts that extend an existing domain model to consider how language can define entire new domain models from scratch.

As a concrete example, let's return to the scenario from \cref{sec:models-probabilistic-reasoning}. Suppose your friend is telling you about a tug-of-war tournament that took place the prior weekend---only this time, you've never heard of tug-of-war before and don't know how it's played. Your  friend might explain the scenario to you using language---indeed, their description might sound similar to the one our paper itself uses to convey the concepts of this particular situation:

\begin{quote}
\textit{Tug-of-war is a game played between teams of players. First, strength levels vary widely from person to person. Furthermore, each person has a percentage of the time that they are lazy. The strength of a team is the combined strength of its members, except that in any given match, each player may decide to be lazy, and thus contribute only half of their strength. Whether one team beats another just depends on which team pulls stronger that match.}
\end{quote}

\noindent Given this language, you can learn the underlying domain model necessary to reason about  future observations (\textit{Even working as a team, Lio and Alex could not beat Josh}) and answer questions (\textit{How strong is Josh?}). In this section, we explore how the components of our framework can be used to construct an entire domain model as it is communicated in language, using the tug-of-war domain as an illustrative example.

\paragraph{Linguistic concepts as program expressions.} Considering the vignette above, we might distinguish between two kinds of statement in your friend's description of tug-of-war:
\begin{itemize}
    \item Some statements introduce new concepts solely in terms of previously introduced concepts (e.g., \textit{Whether one team beats another just depends on which team pulls stronger that match}).
    \item Other statements posit the existence of new primitive concepts, like \textit{strength} and \textit{laziness}, that have certain properties (e.g., \textit{Strength levels vary widely from person to person}).
\end{itemize}
The first case is similar to the sentences we saw in \cref{sec:growing-world-model}, and we can interpret them as language-of-thought definitions. The second case, however, is genuinely new: these sentences neither define new words in terms of an existing domain theory, nor encode predicates over possible worlds. Rather, they define \textit{random variables} that we expect to have different values in each possible world.\footnote{\label{foot:hierarchical}
    An alternative perspective is that the sentences we consider in this section\textemdash both straightforward definitions, and sentences introducing new primitive concepts\textemdash \textit{do} still encode predicates on possible worlds. According to this viewpoint, a sentence like \textit{``The term `uncle' refers to the brother of one's parent, or the husband of one's aunt''} is an assertion that can be true or false; maybe \textit{uncle} means something different in another possible world. To understand this viewpoint within our framework, we need to imagine that there is a background world model that models uncertainty about the \textit{code} of a first-order world model (which definitions exist, and how they are defined). If we had such a \textit{model over world models}, then sentences like \textit{``Players have different strength levels''} could be interpreted as conditioning statements, observing that \textit{strength} exists as a variable and that its value should vary from person to person. Conditioning on this constraint, we could then sample from the posterior over world models that satisfy this property. In this posterior, there would be some uncertainty over exactly how strength is modeled: e.g., does it vary according to a Gaussian distribution, and if so, with what parameters? 
    
    We find this view appealing, and believe that making it practical would be an intriguing technical challenge, requiring new developments in the field of \textit{Bayesian probabilistic program synthesis}~\citep{saad2019bayesian}.
    In this section, we take a shortcut, of assuming that the meaning distribution induced
    by a sentence like \textit{``Players have different strength levels''} directly 
    samples model fragments consistent with the statement. That is, we ask our meaning
    function to \textit{amortize} inference in the hierarchical model, directly proposing
    code defining \textit{strength}, rather than first translating to a conditioning statement
    about \textit{strength} existing, and then using a slower inference algorithm to 
    infer its definition.
}

In Church, such variables can be defined using \lstinline{mem}: for example,

\begin{center}
\noindent\lstinline[breaklines]{(define strength (mem (lambda (person) (normal 100 20))))}
\end{center}

\noindent declares that expressions of the form \lstinline{(strength person)} are well-formed and evaluate to a number in each possible world, and that our prior distribution for a new person's strength is a Gaussian centered at 100. (The \lstinline{mem} construct \textit{memoizes} the defined function, so that repeatedly evaluating \lstinline{(strength 'lio)} in the same world will always give the same result.) It might seem strange to claim that the \textit{meaning} of the sentence \textit{``Players have different strength levels''} includes a specific prior over player strengths, like \lstinline{(normal 100 20)}. We do not make this claim: rather, the meaning function induces a \textit{distribution} over possible definitions of \lstinline{strength}, each of which uses a different prior. What the different possible translations have in common is that they model strength as a continuous variable assigned on a per-player basis, with some population-level variation. See Footnote~\ref{foot:hierarchical} for further discussion of this distribution, and how it might arise from the literal meaning of the sentence being translated.

\begin{figure}[htbp!]
    \centering
    \includegraphics[width=\textwidth]{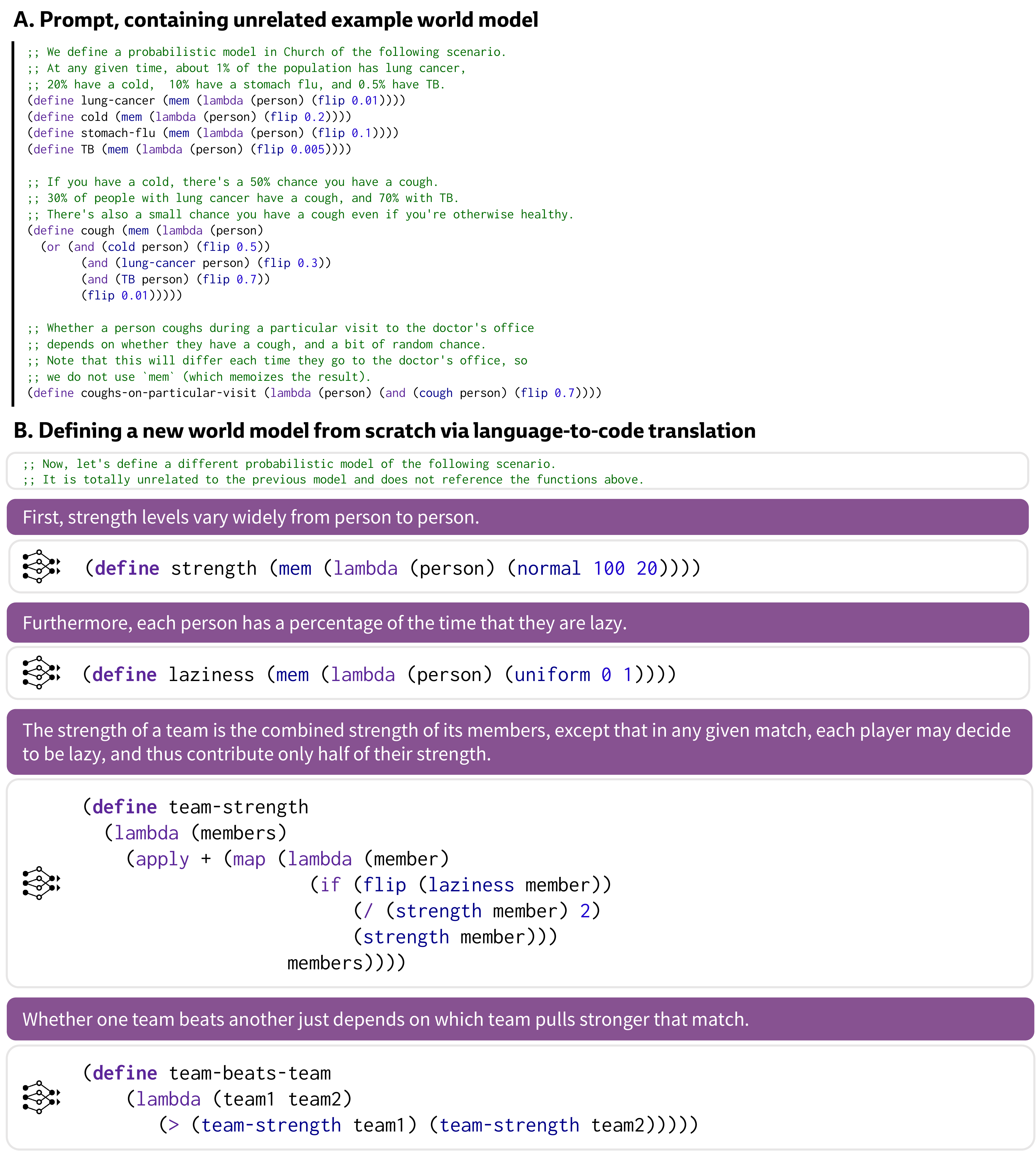}
    \caption{Constructing the tug-of-war model from scratch. This can be accomplished with the same overarching language-to-code approach. (A) We provide a prompt containing one or more unrelated world models as examples. (In this case, the world model defines a medical diagnosis domain.) (B) Prompted line-by-line with language explaining the tug-of-war, \texttt{Codex} constructs a generative model from scratch that is semantically equivalent to the one from \cref{sec:models-probabilistic-reasoning} (modulo some superficial naming and parameter choices).}
    \label{fig:constructing-translations}
\end{figure}

\paragraph{Translating new concepts from language.} As before, because each sentence \textit{means} some distribution over program fragments in a probabilistic language, we can use probabilistic language-to-code translation models like Codex as models of the meaning function. In Fig.~\ref{fig:constructing-translations}, we prompt Codex with an unrelated example world model in domain about diseases and symptoms, and then ask it to translate sentences defining the tug-of-war domain.

\paragraph{Constructing the domain model from new program expressions.} By translating
each sentence in a domain description in sequence, we can\textemdash starting with \textit{no} definitions beyond those built into Church\textemdash build a domain model just as rich as the ones we hand-coded in earlier sections. In Fig.~\ref{fig:constructing-translations}, although the specific priors may vary slightly, Codex recovers all the essential structure of our hand-coded tug-of-war model. Once we have a new domain model, we can immediately begin interpreting observations and queries, like those in \cref{sec:models-probabilistic-reasoning}, or continue to extend the domain model with new definitions.

\paragraph{Putting it together: Growing and constructing world models from language.} In this section, we've illustrated how the same basic building blocks used in the rest of the paper --- language-to-code translation and probabilistic programs --- can be used to extend and construct new world models. Hopefully, these simple sketches highlight a much deeper point: systems that have the ability to author world models in a universal programming language like Church can take advantage of the infinite expressivity of code to generalize to new kinds of language and thinking. 

Nevertheless, the examples presented in \cref{sec:growing-and-constructing} were limited to cases where there was an explicit connection between linguistic instructions and the resulting probabilistic programming expressions. In reality, this relationship is often indirect; language typically only provides us clues about how to think about a situation. In still other instances, we assemble world models in the absence of language, drawing instead on prior experience of similar situations. How can we build systems that \textit{learn} to build world models on-the-fly? How can such systems \textit{remember} and expand on prior world models to understand new situations? And how can they incorporate not just language, but the full spectrum of experiences in the world? In \cref{sec:open-questions}, we consider these questions as part of a discussion of the many future research directions needed to scale our framework to a general model of cognition.

\clearpage
\section[Future directions]{Open questions and future directions}\label{sec:open-questions}
By using neural models to translate sentences into probabilistic programs, the sections above demonstrated how LLMs could extract meaning from\textemdash and inference engines could reason about\textemdash language describing uncertain situations, relational structures, embodied situations and goal-directed reasoning. However, these vignettes also leave open many questions about how to scale this framework to more complex language, and how to automate the process of building meaning representations for new domains.
Together, these questions offer a roadmap for progress on central challenges in modeling language, reasoning, and their interaction, across many sub-fields of artificial intelligence and cognitive science.

\subsection{Scaling models of rational meaning construction}
We begin by describing several of the most important research directions necessary for scaling the framework we have articulated throughout this paper towards a more complete model of integrated cognition and language understanding.

\subsubsection{Building new world models on the fly}\label{building-new-world-models}
A key aspect of our proposed architecture is that language is interpreted relative to a probabilistic model of a domain, capturing just enough structure to represent the situation at hand. In \cref{sec:constructing-domain-theories}, we saw that LLMs could generate these programmatic world models, assuming the model was communicated via a sequence of natural language definitions. But people rarely need such elaborate scene-setting: we can understand language about the world even if no teacher has carefully drawn our attention to the relevant concepts beforehand. A key question is how to model this capability. How do minds craft bespoke world models on the fly, drawing in just enough of our knowledge about the world to answer the questions of interest? How does this process balance competing priorities, such as fidelity to what we know about the world, relevance to the problem at hand, and the efficiency and robustness of inference? These tradeoffs can sometimes seem to evolve during the course of a single chain of human thought. These questions are related to the classic \textit{frame problem}~\citep{mccarthy1980circumscription} in artificial intelligence and cognitive science, and to recent proposals for addressing it in the setting of causal, probabilistic reasoning~\citep{icard2015resource}. These approaches view the problem as one of retrieval: from a vast array of knowledge we have about the world, how can we select just the relevant parts for reasoning about a particular problem? It remains unclear, however, whether the sequences of bespoke models and approximate inferences produced by our minds can be understood as resource-rational approximations to coherent reasoning and planning in some larger, unifying world model, even in principle.

Most probabilistic programming languages were designed for inference in a single, unifying world model~\citep{goodman2012church,bingham2019pyro,carpenter2017stan,milch20071} that was written by an external mechanism, not to dynamically explore a sequence of probabilistic programs that are being synthesized, learned, and/or edited on the fly. But some progress in language-level support for dynamic world modeling has already been made. Probabilistic programs in Gen~\citep{cusumano2019gen} have been used to synthesize and edit other probabilistic programs ~\citep{saad2019bayesian,witty2019bayesian}, and to approximate globally coherent inferences by bridging across sequences of probabilistic programs describing translations among only partially-overlapping worlds~\citep{mansinghka2018probabilistic,cusumano2018incremental,cusumano2020automating,lew2023smcp3}. Analogous language-level support for dynamic abstraction for planning with symbolic world models has also been developed~\citep{zhi2022pddl}. It remains to be seen to what extent these new degrees of freedom can be exploited by language-to-code models targeting these newer probabilistic programming platforms.

How could the common-sense background knowledge needed for dynamic world model synthesis be represented, even in principle? Modern game engines may provide important clues. They can be reconfigured and scripted to simulate diverse imaginary worlds and narratives, featuring interactions between physical objects and goal directed agents in both realistic and physically impossible environments. They routinely combine simulations of the same environment at multiple levels of detail, making computational tradeoffs that are in some ways analogous to the tradeoffs faced by human thinking. The level of scale, coherence, realism, and computational efficiency that they achieve still vastly outstrips the best multi-modal neural models. Although some progress is already being made by synthesizing lightweight, probabilistic game engine scripts using language-to-code models~\citep{zhang2023grounded}, many fundamental challenges remain. Game engines lack crucial affordances for robustly fitting world models to sparse data, simulating rare events, and planning under uncertainty. And despite promising progress in neurally-guided program learning~\citep{ellis2020dreamcoder}, showing that libraries and DSLs can be learned from sparse data, there seems to be a long way to go before we can learn game-engine like rules that are sufficient to robustly model common sense. Flexible synthesis and learning mechanisms that can hope to scale across the vast scope of human thought thus seems to require new ideas that span and integrate probabilistic programming, cognitive architecture, and hierarchical program learning.

\subsubsection{Scaling probabilistic inference in dynamically synthesized world models}\label{scaling-probabilistic-inference}
A central challenge not addressed by this paper is how to scale probabilistic inference to begin to approach the robustness, speed, efficiency, and flexibility of human thought. Consider that the rejection sampling algorithm used in \cref{sec:language-and-world-models,sec:growing-and-constructing} requires an exponentially-growing number of proposal attempts as the scenario becomes less likely under the prior. Although many exact inference methods for probabilistic programs are much faster and more reliable, they are too restrictive to support many of the world models in this paper~\citep{saad2021sppl,holtzen2020scaling,shan2017exact,gehr2020lambdapsi,gehr2016psi}. And although there are many approaches to generic approximate inference in probabilistic programs, drawing on MCMC~\citep{goodman2012church,wingate2011lightweight,carpenter2017stan}, sequential Monte Carlo~\citep{mansinghka2014venture,tolpin2016design}, and variational methods~\citep{hoffman2013stochastic,bingham2019pyro,ranganath2014black,mansinghka2014venture}, they all routinely struggle to solve simple problems that can be solved by custom algorithms.

One potential way forward is to explicitly generate models of thinking processes that augment the world models with which they are thinking, by synthesizing inference programs~\citep{mansinghka2018probabilistic,cusumano2019gen} tailored to specific problems. For example, Venture’s inference meta-programming language is designed to enable concise specification of sequential inference processes that combine SMC, dynamic programming, MCMC, gradient-based optimization, and variational inference to perform inference in a sequence of world models and queries that grows dynamically. Data-driven proposals for use with these thinking strategies can also be generated in real-time, without any offline learning, using dynamic programming over blocks of highly coupled variables. This approach has recently outperformed machine learning methods on hard common-sense reasoning problems in databases with millions of records~\citep{lew2021pclean}. Scaling this approach will require not just synthesizing world models but automatically analyzing and decomposing them, analogously to how inference algorithm designers decompose large inference problems into sequences of more tractable subproblems.

Another promising approach is to train neural networks to make data-driven proposals via amortized inference, potentially using synthetic data from an open-ended simulator of world models and queries~\citep{wu2022foundation}. This can be seen as an alternative to inference programming, avoiding the need for explicit symbolic analysis of the process of thought. It can also be seen as a potential technique by which inference programs might eventually be synthesized, once a suitable training corpus can be generated synthetically — as well as a source of data-driven proposals that can be recombined by inference programs.

\subsubsection{Resource rational amortization in meaning construction and problem solving}\label{amortizing-translation}
In some of our examples, a sentence (e.g., ``Gabe is stronger than Josh'') is translated to a meaning representation that looks very much like its classical formal semantics, composing the literal meanings of each word in the sentence. But in other examples (e.g., ``several of the faculty are real slackers''), the translations appear to incorporate complex contextual and pragmatic judgments, judgments that might otherwise have been arrived at via probabilistic inference in a model of speakers, listeners, and their intents~\citep{goodman2016pragmatic}. This raises the question of where to draw the line between translation and inference. Versions of this question have been extensively studied (e.g., does a word like ``some'' imply ``not all'' as part of its meaning, or does this implicature arise via after-the-fact pragmatic reasoning~\citep{tessler2022logic}?), and some past work has offered a unifying view via theories of \emph{amortized pragmatics}~\citep{white2020learning}, whereby RSA-style inferences are ``compiled down'' into new word meanings.

A key feature of our architecture is that it is largely \emph{agnostic} to where exactly the boundary should lie, and as such could help to model and extend this process of amortized inference in language understanding. For example, as expanded on below, we could extend our symbolic world models to include aspects of the language understanding process itself (such as those described in symbolic derivations of \textit{semantics} \citep{kratzer1998semantics,montague1970english,pollard1994head,steedman2001syntactic,steedman2011combinatory}, and those used explicitly to compute its \textit{pragmatic} interpretations \citep{fox2007free,goodman2016pragmatic}). Symbolic inferences about meanings could then be used to train the language understanding module to directly generate the results of this symbolic inference process\textemdash for use either as a fully amortized pragmatic translator, or as a proposal distribution within a larger Monte Carlo algorithm that could score and reject inaccurate translations. 

In addition to making aspects of translation symbolic, we could consider approaches to amortizing the  more general probabilistic inferences required to answer queries. By supervising ``translation models'' directly with the final outputs of symbolic inference, across a wide variety of tasks, we could enable a pure neural inference mode for these systems that may overcome some limitations of models trained only on language and code. As described above, such supervised models could also be incorporated as proposal distributions in posterior sampling algorithms, leading to improved efficiency without sacrificing the ability to correct for learned biases that may be inapplicable when tackling novel problems.

Ultimately, we envision a new kind of neurosymbolic model in which, rather than pre-assigning responsibilities to the neural or symbolic program, models may flexibly perform any part of the language understanding via explicit probabilistic inference or learned, amortized prediction, with tradeoffs in speed and accuracy for any allocation of responsibilities to modules. The research question is how to do this automatically\textemdash how do we identify pieces of a computation that can reliably be emulated by a neural model, how do we train this neural model efficiently, and how do we decide at runtime which inference mode to use? As above, these questions raise many opportunities to take inspiration from our scientific understanding of the separation of responsibilities in language and thought, and work on learning for inference in more general probabilistic models.

\subsubsection{Language generation}

The preceding discussion has focused largely in problems of language \emph{understanding}---mapping from utterances to inferences about the state of the world that those utterances describe. But effective models of language use should also be able to explain \emph{generation}, making it possible to translate the results of inference back to language. As with the problem of language-informed thinking that we focus on this paper, it is useful to model language generation as two distinct processes: choosing \emph{what to say}, then \emph{how to say it}
\citep{duboue2003statistical}. And as with understanding, the first phase requires a model of the world, and of the speaker's goals within it. What additional work is needed to adapt our models of rational meaning construction for generation?

One possibility, alluded to in the discussion of amortization above, is to interpret the language understanding machinery described above as a \emph{model of a listener}, then perform language generation by selecting utterances that cause this model listener to form correct beliefs or take appropriate actions \citep{goodman2016pragmatic,fox2007free}. This extra layer of reasoning introduces major inferential challenges: the generation model most now reason both about the set of possible utterances and the effect of each utterance on the distribution over possible worlds inferred by a listener. Here it is once again possible to leverage large-scale statistical learning---for example, using LLMs to directly translate candidate communicative intentions back to natural language strings, which may then be used as candidate utterances to be scored using a formal model of language understanding.
Such a hybrid neuro-symbolic generation model 
\citep{langkilde1998generation,fang2022whole}
offers a path towards language generation that is expressive and fluent, but avoids the truthfulness and hallucination problems that plague all purely neural language generation models that exist today
\citep{wiseman2017challenges,maynez2020faithfulness}.

\subsection{Implications for cognitive science}
In this section, we describe several research directions for other closely related disciplines that study language and thought in natural minds, brains, and behavior, focusing on productive intersections in relation to this framework.

\subsubsection{Connections to cognitive and formal models of linguistic structure}\label{sec-open-question-formal-linguistics}

In all the examples described above, the process of translating utterances into formal meaning representations was performed with a black-box statistical model, while reasoning about those meaning representations leveraged an explicit symbolic inferential process. 
However, an enormous body of work in linguistics has argued that the process of mapping from utterances to meaning representations can itself be described (at least approximately) in terms of symbol processing operations
\citep[][\emph{inter alia}]{montague1970english,pollard1994head,steedman2001syntactic}. 
By design, most of our ``meaning representations'' are designed to support efficient reasoning about domain-specific world models, and bear only a vague resemblance to formal and domain-general linguistic representational theories. But can the symbolic models of linguistic meaning posited by these theories (as opposed to the symbolic models of reasoning we already draw on) be incorporated into our framework?

As noted in \cref{amortizing-translation}, a fully realized model of rational meaning construction should be able to flexibly move computation across the statistical--symbolic boundary, ``compiling'' results of symbolic inference into amortized computation, or retrieving symbolic descriptions of amortized processes for explicit verification. In this view, the vignettes above treat the meaning representation process as culminating in domain-specific representations and amortized by default. But probabilistic symbolic models of meaning \citep[e.g.,][]{kwiatkowksi2010inducing}, or Bayesian and game-theoretic models of semantics \citep[e.g.,][]{goodman2016pragmatic} can themselves be implemented as probabilistic programs and composed with domain-specific inferential computations, resulting in an almost purely symbolic (but amortizable) language understanding process similar to the one described by \citet{goodman2015probabilistic}.

Such a model would also offer an appealing path toward \emph{learning} language in a more sample-efficient (and perhaps human-like) ways. Today's neural sequence models require orders of magnitude more data than human learners to discover the structural regularities underlying human languages \citep{linzen2020can}. Explicit probabilistic symbolic models, by contrast, can discover this structure extremely sample-efficiently \citep{yang2022one}.
A model that could automatically infer symbolic meaning representation rules from data, then amortize this representation system into a statistical translation model \citep{liang2008structure}, would be capable of both efficient learning of language, and efficient modeling of other domains using language.
It would also offer a framework for modeling other key aspects of language acquisition, including explicit linguistic instruction (of word meanings, rules of grammar, etc), tradeoffs between different formal representational schemes, and the relationship between linguistic \emph{competence} (understood as symbol-side language processing) and linguistic performance (understood as statistical-side processing).

The semantic framework in this paper is most closely related to other \textit{cognitive semantic frameworks} (eg. \cite{pinker1984language,jackendoff1985semantics,lakoff1988cognitive,pietroski2018conjoining}) that explicitly propose that human language constructs meanings from conceptual and cognitive primitives, including those for causal reasoning, or core knowledge representations of physics and agents. Related information-theoretic proposals have proposed that languages are effectively designed to be efficiently communicable externalizations of underlying thoughts---that the structure of human languages derives from underlying structure in the semantic representations we wish to communicate, and indeed may be driven by environmental and domain-specific pressures (eg. \cite{zaslavsky2018efficient,mollica2021forms,gibson2019efficiency}).

Other related acquisition theories posit that these structural relationships between the representations of thought and externalizable language play an important role in language acquisition. Under these theories, humans can so efficiently learn or hypothesize the meanings of sentences because they ``map cleanly" onto the cognitive structures already present in the minds of the language learner \citep{snedeker_clean}; language learning is \textit{bootstrapped} by these predictable, structured mappings between the underlying space of meanings and the syntax of language \citep{gleitman2005hard,hartshorne2016psych,pinker1987bootstrapping}. In preliminary experiments, we find intriguing evidence that large language-to-code models can extract and generalize syntactic patterns between language and code, including to bootstrap hypotheses about the semantics of novel words expressed as probabilistic programs based on contextual, syntactic usage (see \textit{Syntactic Bootstrapping}, \cref{appendix:open-questions:fig:bootstrapping-translations}). Future work can explore therefore whether these statistical distributional models might be used to implement cognitive models of bootstrapped language acquisition.

\subsubsection{Modeling the mechanisms of human thought}

Using tools for adaptive Bayesian inference over flexibly structured symbolic representations---including not only probabilistic programs but more generally hierarchical Bayesian models \citep{tenenbaum2011grow, griffiths2010probabilistic}, resource-rational modeling \citep{lieder2020resource,gershman2015computational}, and program induction \citep{lake2017building,piantadosi2012bootstrapping}---computational cognitive scientists 
have built quantitatively predictive and functionally explanatory models of human behavior in almost every domain of cognition.  This range spans from models of perception, concept learning and categorization, causal reasoning, decision-making and planning, to intuitive physics, theory of mind, sentence processing, and cognitive and language development \citep{griffiths2006optimal,lake2017building,goodman2014concepts,ho2022planning,baker2011bayesian,jara2020naive,goodman2016pragmatic,perfors2011learnability}.  
However, in almost every one of these cases, the models are not fully ``stimulus-computable'': Behavioral experiments in cognitive psychology almost always use natural language to present participants with some situation for thinking about (in addition to perhaps perceptual stimuli); language is also almost invariably used to pose some question or goal as the end for thinking. Put another way, almost all our behavioral experiments---like so many instances of cognition in the wild---follow the language-informed thinking paradigm of this paper. But our cognitive models traditionally do not; they are created by hand from the modeler's understanding of the natural language task description, rather than synthesized automatically from the linguistic stimuli presented to participants. To what extent can the rational meaning construction framework presented here reduce the need for computational cognitive scientists to manually create Bayesian models that match the natural-language prompts given to humans in behavioral experiments? Can we build ``language-computable’’ models of human thought, that are much easier to test and vary via large-scale online experiments?  

We have already begun to explore these possibilities and shown promising preliminary results in several domains, including to model how language implicates commonsense physical reasoning about linguistic scenes \citep{zhang2023grounded}, social reasoning about goal-directed agents \citep{ying2023nipe}; as well as to test the claim that the LLM-based meaning function we implement in this paper can compute amortized pragmatic judgments of scalar implicatures that accord with human interpretations \citep{lipkin2023evaluating}.

There is also a growing body of research in computational cognitive science showing that salient dynamics of thought, including well-known departures from Bayesian norms, can be explained via Monte Carlo inference approximations that aim to rationally use limited computational resources~\citep{lieder2020resource,lieder2014high,sanborn2017sampling,chater2020probabilistic,gershman2015computational}. In some cases, human inferences seem to rest on just a single, highly approximate sample~\citep{vul2014one}, or perhaps just a few of them~\citep{vul2008measuring}. If we extend our proposed architecture for rational meaning construction to incorporate these kinds of Monte Carlo mechanisms, could we build models of language-guided thinking that can be directly compared at a more mechanistic level to human behavior?  How will processes of language understanding and reasoning interact mechanistically, and can we build resource-rational approximate inference models that capture this interaction? 

\subsubsection{Language and thought in the brain}\label{open-questions-neuroscience}
Evidence from cognitive neuroscience suggests a number of parallels between the framework we describe in this paper, and how language relates to systems for general cognition in the human brain. Over decades, cognitive neuroscientists have mapped out a series of interconnected areas in the frontal and temporal lobes that are implicated in \textit{human language processing}. This ``language network'' is activated in both linguistic comprehension \citep{deniz_representation_2019, fedorenko_new_2010, macsweeney_neural_2002, regev_selective_2013, scott_new_2017} and production \citep{menenti_shared_2011,hu_language_2021}. It is sensitive to regularities in all levels of linguistic structure---from phonology, to words, to phrases and sentences \citep{wilson_beyond_2008, lerner_topographic_2011, silbert_coupled_2014, blank_domain-general_2017} and is implicated in combinatorial semantic and syntactic processing \citep{fedorenko_lack_2020,hu_language_2021}.

Convergent evidence suggests that the language network is distinct from these systems, and that it is\textit{not} activated in more general, non-linguistic cognition. Aphasic individuals with damage to the language network exhibit impaired language production and comprehension, but retain the ability to solve arithmetic and logic puzzles, reason about causality and social situations, and perform many other non-linguistic tasks \citep[e.g., ][]{fedorenko_language_2016,basso_spared_1985, bek_language_2010, klessinger_algebra_2007, luria_aphasia_1965, lecours_linguistic_1980, varley_aphasic_1998}. Functional neuroimaging studies provide further evidence that the language network is \textit{not} activated in a variety of non-linguistic tasks including reasoning about arithmetic, logic, actions, or events \citep{amalric_origins_2016, amalric_distinct_2019, blank_functional_2014, deen_functional_2015, fedorenko_functional_2011, monti_functional_2007, monti_thought_2012, paunov_functionally_2019, paunov_differential_2022, shain_no_2022}. 

In tandem, a broader line of cognitive neuroscience work has located non-linguistic networks that \textit{are} activated in processing many of the core cognitive domains we model throughout this paper, including logic, mathematical reasoning (eg. \cite{amalric_distinct_2019,monti_functional_2007}), 
social reasoning and planning \citep{saxe2006s,saxe2006overlapping,adolphs2009social}; and physical reasoning and simulation \citep{schwettmann2019invariant,pramod2022invariant}. More recent work suggests the existence of an ``amodal semantics network" \citep{ivanova2021language,ivanova2022role}, a network that appears proximal to the language networks activated in processing linguistic structures, interfaces between the language network and the more general multiple demand networks involved in complex non-linguisitc cognition, and that appears to be activated specifically in processing semantically meaningful sentences (as opposed to scrambled tokens or syntactically correct but semantically incoherent strings.)

Recently, neuroscientists who study language cognition have begun to draw explicit parallels between the language network and LLMs \citep[see][for a review]{mahowald2023dissociating}. Several recent studies have observed that smaller LLMs trained specifically on the distriutional statistics of language (generally focusing on the GPT-2 model) can predict brain activity in humans processing sentence input \citep{caucheteux_brains_2022, goldstein_shared_2022, schrimpf_neural_2021} and may share representational characteristics of the human language network \citep{fedorenko_lack_2020,shain_fmri_2020}. These accounts, however, align LLMs with the modular role we propose for neural models in our framework---\textit{not} as end-to-end models of language and reasoning, but instead as robust, context-aware mappings between language and meanings. As a ground for future work, our framework can inform evaluations of LLMs with respect to human language understanding. For instance, our proposal suggests that code-trained LLMs might better capture latent semantic and syntactic structure than language-only LLMs. Ideas from neuroscience, in turn, can help us figure out which kinds of computations can be neurally amortized and where our model's boundary between language and thought should lie.

\subsection{Implications for AI}
\subsubsection{Structured hybrid models of language and thought}
Growing awareness of the limitations of LLM-based reasoning has motivated several recent proposals for interfacing language models with external symbolic plug-ins or toolkits \citep{karpasMRKLSystemsModular2022, schick2023toolformer, openai2023gpt4, wolfram2023chatgpt}. At face value, one perspective is to view rational meaning construction as an argument to add probablistic programs to the growing ``swiss army knife'' of LLM plug-ins. However, we see this notion as inverted: \textit{thought} should not simply be a plug-in on top of language models. Rather, we believe that future AI systems should be architected around \textit{thought}---general-purpose computing systems that provide a principled framework for expressing world models, conditioning them on observations from sources including language and perceptual input, and drawing principled inferences and decisions with respect to the goals of an intelligent system.\footnote{A similar argument has been expressed by Stephen Wolfram in a compelling series of writings on integrating ChatGPT with the Wolfram Language and its suit of symbolic computational tools \citep{wolfram2023chatgpt}.} As we show throughout this paper, many core domains of cognition can be expressed as forms of probabilistic inference. A probabilistic language of thought, in turn, provides a unifying language for world modeling that can nest calls to other cognitively-motivated modules. In this sense, all of these plug-ins and modules would become plug-ins to the substrate of thought, \textit{including} graphics engines, physics simulators, planning algorithms, \textit{and}, in fact, language models themselves. As we discuss in the future directions of each section, scaling any of our toy implementations towards robust, human-like reasoning and language-understanding systems will almost certainly require more sophisticated implementations of each reasoning module. We therefore hope this general probabilistic framework suggests a symbolic substrate that might in turn incorporate many of the specific modules and plug-ins in this recent work.

To this end, another important near-term AI research direction will involve building probabilistic programming frameworks that natively incorporate LLMs. Important steps in this direction are already being taken through work leveraging LLMs to approximate prior probabilities over strings \citep{lew2020leveraging} and amortize complex posterior inferences \citep{wu2022foundation}. Indeed, many popular LLM  techniques, such as scratchpads \citep{nye2021show}, chain-of-thought prompting \citep{wei2022chain}, selection-inference \citep{creswellSelectionInferenceExploitingLarge2022}, STaR \citep{zelikman2022star}, and others can be viewed as implementations of probabilistic programs over string-valued random variables \citep{dohan2022language}. A maturing theoretical understanding of LLMs as probabilistic entities will afford powerful ways of harnessing and controlling generations. For instance, the sequential Monte Carlo (SMC) steering technique introduced under the LLaMPPL framework \citep{lew2023sequential} enables concise and tractable specification of infilling, prompt intersection, and other constrained LLM generation tasks as \textit{language model probabilistic programs}. Many of these hybrid models can be viewed as instantiations of rational meaning construction that make resource-motivated tradeoffs between inference in the unstructured space of strings (words) and more structured hypothesis spaces (worlds).

\subsubsection{Robustness and trustworthiness in language understanding}
Recent, high-profile attempts to deploy LLMs in production highlight the fundamental robustness challenges of using these models as the backbone of usable AI systems \citep{brereton2023bing,sorkin2023revenge}, even with automated filters and supervised finetuning to human preferences. While LLMs may reasonably appear to condition on input language or answer queries under some circumstances, it is precisely this combination of linguistic fluency and underlying unpredictability that makes them problematic in situations where verifiable, systematic behavior is paramount. LLMs easily produce syntactically convincing but inaccurate ``hallucinations'' that fabricate facts and inferences \citep{dziri2022origin,ji2022survey}, fail to consistently condition on rules and constraints described in natural language, including rules intended to ensure user safety \citep{zhuo2023exploring,edwards2023bing}, and can generally degrade into nonsensical or highly undesirable language in the vast, easily accessible ``long tail'' of situations that deviate from their training distribution \citep{roose2023bing,tangermann2023microsoft,bender2021dangers}. 

The unevenness of today's LLMs recalls a classic critique of even older neural architectures \citep{fodor1988connectionism}---that neural models trained on predictive objectives do not produce systematic, logical outputs by design. Similarly, while current or future LLMs may be able in principle to recover the latent representations and algorithms necessary to reason over language---or even successfully approximate them in many settings---they do not \textit{need} to produce systematic results by construction. Rather, they often approximate them with unexpected, undesirable outputs, particularly in out-of-distribution settings.

Even if future LLMs do appear to improve with scale without an external reasoning substrate, engineers may find it desirable to distinguish modularly between external symbolic reasoning engines and language-specific systems to enable separate supervision and verification of each. The framework we present here offers one roadmap for language understanding architectures whose robustness guarantees derive from explicit inference over a structured, editable, and formally constrainable programming language. Inferences themselves, and other formalizable reasoning computations including planning and physical simulation, take place in modules constructed explicitly to perform these calculations.

\subsubsection{Interpreting models that use language}
As with verifiability and robustness, the framework we propose here is an architecture for language understanding systems that are also \textit{inherently interpretable}, or \textit{interpretable by design} \citep{rudin2019stop,rudin2022interpretable}---it constructs visible, editable, and constrainable world models and meanings that serve as the formal basis for inference, rather than post-hoc explanations decoded from or produced over hidden internal
computations.

However, a fundamental part of our hypothesis is that \textit{any} system that reasons effectively over language should need to---explicitly or implicitly---represent and implement the kinds of computations we formalize throughout this paper. Implementations of this framework might therefore also be useful for model-guided hypotheses and experiments intended to explain other less transparent language processing systems, both biological (as we suggest in \cref{open-questions-neuroscience}) and artificial. This framework might be incorporated productively into the growing body of work using explicit world models and symbolic languages to formally model the internal computations of deep neural models \citep{mu2020compositional,biggio2021neural} and LLMs specifically \citep{li2021implicit}; as with the related body of work using structured probabilistic models and reasoning engines to interpret human neural activity on social reasoning, physical understanding, and other general inference tasks \citep{schwettmann2018evidence,watters2021modular,ho2022planning}. Explaining how LLMs represent the meanings of language, and perform computations with them, is a pressing open question whose scientific interest only increases if LLMs do appear to become more coherent and robust with scale.

In light of this, inspired by our proposed architecture, it may be interesting to probe, or trace, whether end-to-end LLMs construct context-specific world models \citep{li2021implicit}, maintain belief distributions over uncertain world states \citep{hase2021language}, and implement reasoning algorithms like probabilistic inference, physical simulation, or planning over these representations.

\subsubsection{Learning from human-scale data}
Large language models must be trained with many orders of magnitude more language data than any human learner encounters over a lifetime. How can we engineer systems that not only understand language as we do, but also learn from human-\textit{scale} language data?

Effective, data-efficient language models hold great relevance for both scientific and engineering applications. Complete cognitive models of human language understanding---including models built on the framework we propose here---should account for language acquisition, as well as language use. For engineering purposes, addressing the data-hungry training regime of current LLMs could also address challenges in learning low-resource languages (and the more general problem of accurately learning and deploying the ``long tail'' of knowledge from statistical distributional data) \citep{kandpal2022large}, incorporating more expensive input modalities like videos or embodied trajectories \citep{ahn2022can,reed2022generalist}, finetuning on more targeted, task-specific supervision like instruction following \citep{openai2023chatgpt}, and generally enabling the construction of smaller, more accessible models that can be trained without massive computing resources and prohibitive economic and environmental costs \citep{bender2021dangers,Dickson2020gpt}. While current ``scaling routes'' look to improve language understanding by \textit{increasing} data supervision, our hypothesis strongly suggests that this is an expensive, and highly indirect, approach towards learning the representations and inference procedures necessary to reason about language.

Instead, our framework suggests several alternative directions for improving data efficiency. First, perhaps the most direct consequence of this framework is the suggestion that neural models need only play a much tighter, focused role in language understanding systems---as translation models that parse from language into structured symbolic programs for reasoning. Training a translation model focused on parsing from language into probabilistic programs almost certainly requires much less data for effective performance than required to solve the general token prediction problem. 

Further, several ideas we discuss in \cref{building-new-world-models} and \cref{amortizing-translation} might also be relevant for training simpler translation models, and using them to bootstrap larger and more complex neural language models. 

First, as we discuss in \cref{amortizing-translation}, we might consider a progressively amortized avenue for training even complex translation models like the one in our concrete implementation, which appears to  contextually amortize certain pragmatic inferences (such as those that adjust vague quantifiers to the context of a particular world model) that could be explicitly computed from a more literal initial semantic parse. One possibility, then, would be to train a more limited, literal semantic parser from language to probabilistic programs, but seek to train neural models that progressively amortize more of these inferences by supervising on its outputs.

Other ideas from human language acquisition might offer more avenues for more radically data efficient learning. Human language learners progress through several phases of language mastery \citep{saffran2001acquisition,tomasello2009usage,brown1973first}, appearing to learn initial but highly imperfect grammars and meaning functions that they refine progressively over time, but much more quickly and with much less data than a comparable LLM trained directly on the distribution of language. Framed as a problem of learning a translation model, however, a more data efficient training regime might also draw inspiration from other methods for learning more flexible translation and semantic parsing distributions. Multiple approaches have used simpler models to bootstrap more complex ones, either by using simpler models trained on more constrained translation objectives to directly initialize the parameters of more complex ones \citep{brown1993mathematics,petrov2008coarse,dong2018coarse}, or using simpler grammars as generative data sources to train more complex models, as in general wake-sleep training methods that learn predictive models to amortize the outputs of a generative distribution \citep{jia2016data,hinton1995wake,andreas2019good}. 

Both of these approaches rely, importantly, on the language of thought hypothesis we advance here, which separates the computational problem of learning a translation distribution from the problem of learning the representations and algorithms necessary for general intelligence. This drastically reduces the latent structure and computational complexity we seek to learn from distributional supervision---to learn as efficiently as people, we propose a framework that \textit{begins} with a substrate for thinking and then suggests avenues for amortizing its outputs or refining translation into this substrate, rather than seeking to learn an effective language of thought itself from natural language data.

\section{Conclusion}\label{sec:conclusion}
Language is central to our cognition. A theory of meaning in human language should explain how language relates to our \textit{thoughts}---how it connects to all our faculties for reasoning, and how it can shift our beliefs across nearly every domain of what we now, change how we act or respond across a broad range of situations, even construct new knowledge that we might later marshal towards yet unspoken questions and goals. This vision lies at the heart of a human theory of language and meaning, but the most expansive visions of AI have also long been ones in which computers share our language, able to meaningfully understand us as we expect to be understood by other people. Today’s large language models have made striking advances towards building this reality in many important regards. For the first time, we have built computer systems that can speak fluently back to us, using many more of our own words than ever before. 

Still, much more is needed to capture our own relationship to language. We do not learn language like a large language model does. We think first, and learn from far less input how language maps into our thoughts. Our own world models and beliefs are not the fragile byproduct of what we can glean from language—they are the basis of and core of our cognition, constructed and maintained purposefully towards our intentions and desires. We, of course, are the ones who created the language  on which today’s machine learning models are now trained. That language is the product of and reflection of our own goals and questions, and of conceptual systems of our own invention. We continue to think completely new thoughts, and we continue in turn to produce entirely new language, coining new words and even constructing wholly new languages so that we can build its meaning in the minds of other humans. A cognitive theory of human language must capture and explain these aspects of our language and thought. It might in turn form the basis for AI models that reliably and predictably understand us, and that work in ways that we can interpret, explain, and control. This white paper is simply a sketch towards these ends: an outline of the computational components that could relate human language and a substrate for cognition, and one proposal for how this approach might also incorporate today’s language models without requiring them to learn to reliably model the world, draw inferences, or make decisions. We hope it can offer one step towards cognitive and AI models that share the meaning we make from language, and that bridge from language into the vast expanse of our thoughts.

\clearpage
\section*{Acknowledgements}
We have many people to thank whose comments, critiques, and feedback have influenced this manuscript and shaped it for the better. Among others, we are grateful to Steve Piantadosi, Jesse Snedeker, Kate Davidson, Ellie Pavlick, Paul Pietroski, Thomas Icard, Luca Bonatti, and Susan Carey for their insightful comments on an early version of this manuscript that was presented at the July 2022 McDonnell Network Workshop; as well as for innumerable helpful comments and feedback on developing versions of this manuscript from Joshua Hartshorne, Judy Fan, Robert Hawkins, Katherine Collins, Anna Ivanova, Cedegao Zhang, Hayley Ross, Anna Ivanova, Benjamin Lipkin, Megan Wei, Jiahai Feng, Xuan Tan, Lance Ying, William McCarthy, Laura Schulz and Tyler Brooke-Wilson. Language from all of these collaborators has invaluably and profoundly informed our thoughts.

The authors gratefully acknowledge support from support from the MIT Quest for Intelligence, AFOSR Grant No. FA9550-19-1-0269, the MIT-IBM Watson AI Lab, the DARPA Machine Common Sense Program, the ONR Science of AI Program, and Siegel Family Endowment. This material is based on work supported by the National Science Foundation Graduate Research Fellowship under Grant No. 1745302 and No. 2141064. Additionally, GG was supported by the MIT Presidential Fellowship, and JDA was supported by NSF Grant IIS-2212310.

\bibliographystyle{apacite}
\fancyhead[L]{} 
\bibliography{main_clean}

\clearpage
\fancyhead[L]{\textit{\rightmark}} 
\appendix
\section*{Appendices}\label{sec:appendix}
We include code for reference below to help better interpret the examples in the paper. This code is included (with human-readable comments) for completeness and for reference, but is not guaranteed to be the the most up-to-date version of these examples. Please refer to the GitHub repository for the most complete, corrected, and up-to-date code for all examples in this paper, as well as instructions for execution and reproducibility: \href{https://github.com/gabegrand/world-models}{\texttt{github.com/gabegrand/world-models}}.
\section{Language and world models}
\subsection{Probabilistic reasoning}
\subsubsection{Generative world model for tug-of-war} \label{appendix-sec:probabilistic-generative}
\begin{code}
\begin{lstlisting}[numbers=left,frame=lines]
;; This Church program models a tug-of-war game between teams of players.

;; Each player has a strength, with strength value 50 being about average.
(define strength (mem (lambda (player) (gaussian 50 20))))

;; Each player has an intrinsic laziness frequency.
(define laziness (mem (lambda (player) (uniform 0 1))))

;; The team's strength is the sum of the players' strengths.
;; When a player is lazy in a match, they pull with half their strength.
(define (team-strength team)
(sum
  (map
    (lambda (player) (if (flip (laziness player)) (/ (strength player) 2) (strength player)))
    team)))

;; The winner of the match is the stronger team.
;; Returns true if team-1 won against team-2, else false.
(define (won-against team-1 team-2)
  (> (team-strength team-1) (team-strength team-2)))
\end{lstlisting}
\caption{Generative domain theory for the Bayesian tug-of-war.}
\label{code:apx-probabilistic-generative}
\end{code}

\subsubsection{Translation examples for tug-of-war} \label{appendix-sec:probabilistic-example-translations}
\begin{code}
\begin{lstlisting}[numbers=left,frame=lines]
;; Now, let's translate some user-defined statements.
;; Each statement begins with either `Condition` or `Query`.
;; `Condition` statements provide facts about the scenario.
;; `Query` statements are questions that evaluate quantities of interest.

;; Condition: Alice won against Bob.
(condition (won-against '(alice) '(bob)))

;; Condition: John and Mary won against Tom and Sue.
(condition (won-against '(john mary) '(tom sue)))

;; Query: If Mary played against Tom, who would win?
(query (won-against '(mary) '(tom)))

;; Certain statements are underspecified and require some interpretation. For example:
;; Condition: Sue is very strong.
(condition (> (strength 'sue) 75))

;; We can `define` new constructs that are useful for translation. For example:
;; Condition: Bob is stronger than John.
(define (stronger-than? player-1 player-2)
  (> (strength player-1) (strength player-2)))
(condition (stronger-than? 'bob 'john))

;; Query: Is Sue stronger than Mary?
(query (stronger-than? 'sue 'mary))

;; Condition: A couple of the players are stronger than John.
(condition (>= (count (map (lambda (player) (stronger-than? player 'john) players)) 2)))
\end{lstlisting}
\caption{Prompt examples}
\label{code:apx-tug-of-war-prompt}
\end{code}
\subsection{Relational reasoning}
\subsubsection{Generative world model for kinship} \label{appendix-sec:kinship-generative}
\begin{code}
\begin{lstlisting}[numbers=left,frame=lines]
;; -- KINSHIP GENERATIVE DOMAIN THEORY --

;; All the names that can be used in the conversational context.
(define ALL-NAMES '(avery blake charlie dana))

;; Generates unique person ids of the format (person-0, person-1, ...)
(define PERSON-PREFIX "person-")
(define new-person-id (make-gensym PERSON-PREFIX))
(define (id->idx person-id)
  (string->number (string-slice (stringify person-id) (string-length PERSON-PREFIX))))

;; Randomly assign a gender
(define person->gender (mem (lambda (person-id)
  (uniform-draw '(male female)))))

;; Randomly-ordered list of person names
(define NAMES (shuffle-unique ALL-NAMES))
(define person->name (mem (lambda (person-id)
  (list-ref NAMES (id->idx person-id)))))

;; Person node in tree
(define (person person-id parent-1-id parent-2-id) (list
  (pair 'person-id person-id)
  (pair 'name person-id)
  (pair 'gender (person->gender person-id))
  (pair 'parent-1-id parent-1-id)
  (pair 'parent-2-id parent-2-id)))

;; Generate the full tree
;; Max tree size is 1 + (sum_{n=0}^{n=MAX-DEPTH} 2 * MAX-WIDTH^n)
(define MAX-WIDTH 3)
(define MAX-DEPTH 2)
(define PARTNER-PROBABILITY 0.5)
(define (generate-tree root-primary-id root-secondary-id depth)
  (let* (
         ;; Create the primary parent
         (parent-1-id (new-person-id))
         (parent-1 (person parent-1-id root-primary-id root-secondary-id)))
  (if (flip PARTNER-PROBABILITY)
    ;; Case: parent-1 has partner
    (let* (
      ;; Create the secondary parent
      (parent-2-id (new-person-id))
      (parent-2 (person parent-2-id () ()))

      ;; Link the parents with a partner relation
      (parent-1 (append parent-1 (list (pair 'partner-id parent-2-id))))
      (parent-2 (append parent-2 (list (pair 'partner-id parent-1-id))))

      ;; Generate children
      (n-children (if (>= depth MAX-DEPTH) 0 (bounded-geometric 0.5 0 MAX-WIDTH)))
      (child-trees (repeat n-children (lambda () (generate-tree parent-1-id parent-2-id (+ depth 1)))))

      ;; Update the parents to point to the children
      (child-ids (map (lambda (t) (lookup (first t) 'person-id)) child-trees))
      (parent-1 (append parent-1 (list (pair 'child-ids child-ids))))
      (parent-2 (append parent-2 (list (pair 'child-ids child-ids)))))
    (append (list parent-1) (list parent-2) (shallow-flatten child-trees)))

    ;; Case: parent-1 has no partner
    (list parent-1))))

;; Generate the global tree.
(define T (generate-tree () () 0))

;; Assign names randomly to (some of) the people in the tree.
(define (add-names-to-tree tree names)
  (if (null? tree) ()
  (let*
    ;; Probability of addding a name to the first person
    ((p (min 1.0 (/ (length names) (length tree))))
     (person (first tree)))
    (if (flip p)
        ;; Name the person
        (let
          ((named-person (update-list person 1 (pair 'name (first names)))))
        (cons named-person (add-names-to-tree (rest tree) (rest names))))
        ;; Don't name the person
        (cons person (add-names-to-tree (rest tree) names))))))

;; Update the tree with the name information.
(define T (add-names-to-tree T NAMES))
\end{lstlisting}
\caption{Generative domain theory for family trees.}
\label{code:kinship-generative}
\end{code}

\subsubsection{Kinship tree utilities} 
\begin{code}
\begin{lstlisting}[numbers=left,frame=lines]
;; -- KINSHIP TREE UTILITIES --

;; Returns all instances of person with property `key` equal to `value`
(define filter-by-property
  (mem (lambda (key value)
    (filter (lambda (p) (equal? (lookup p key) value)) T))))

;; Returns the unique instance of person with name.
(define get-person-by-name
  (mem (lambda (name)
    (let
      ((results (filter-by-property 'name name)))
    (if (null? results) () (first results))))))

;; People without a name can be referenced directly by person-id.
(define get-person-by-id
  (mem (lambda (person-id)
    (if (null? person-id)
        ()
        (let ((idx (id->idx person-id)))
          (if (>= idx (length T)) () (list-ref T idx)))))))

;; Get a person object either by name or person-id.
(define get-person
  (mem (lambda (person-ref)
    (cond
      ((null? person-ref) ())
      ((member? person-ref NAMES) (get-person-by-name person-ref))
      (else (get-person-by-id person-ref))))))

;; Get a property of a person.
(define get-property
  (mem (lambda (name key)
    (lookup (get-person name) key))))
    
;; -- TREE OPERATORS --
;; predicate :: name -> boolean

(define (map-tree predicate)
  (map (lambda (x) (predicate (lookup x 'name))) T))

(define (filter-tree predicate)
  (filter (lambda (x) (predicate (lookup x 'name))) T))

(define (exists predicate)
  (some (map-tree predicate)))

\end{lstlisting}
\caption{Utility functions for kinship trees.}
\label{code:kinship-utilities}
\end{code}

\subsubsection{Kinship conceptual system} \label{appendix-sec:kinship-conceptual-system}
\begin{code}
\begin{lstlisting}[numbers=left,frame=lines]
;; -- KINSHIP CONCEPTUAL SYSTEM --

;; Gets the partner of a person.
(define (partner-of name)
  (get-property (get-property name 'partner-id) 'name))

;; Gets the parents of a person.
(define (parents-of name)
  (let* ((parent-1-id (get-property name 'parent-1-id))
         (parent-1-name (get-property parent-1-id 'name))
         (parent-2-id (get-property name 'parent-2-id))
         (parent-2-name (get-property parent-2-id 'name)))
    (list parent-1-name parent-2-name)))

;; Gets the grandparents of a person.
(define (grandparents-of name)
  (let ((parent-1 (first (parents-of name))))
    (parents-of parent-1)))

;; Gets the children of a person.
(define (children-of name)
  (let ((child-ids (get-property name 'child-ids)))
    (map (lambda (child-id) (get-property child-id 'name)) child-ids)))

;; Gets the siblings of a person.
(define (siblings-of name)
  (let* ((parent-1-id (get-property name 'parent-1-id))
         (child-ids (get-property parent-1-id 'child-ids))
         (child-names (map (lambda (child-id) (get-property child-id 'name)) child-ids)))
    (filter (lambda (child-name) (not (equal? child-name name))) child-names)))

;; -- BOOLEAN RELATIONS --
(define (partner-of? name_a name_b)
  (equal? name_a (partner-of name_b)))

(define (parent-of? name_a name_b)
  (member? name_a (parents-of name_b)))

(define (father-of? name_a name_b)
  (and (equal? (get-property name_a 'gender) 'male)
       (parent-of? name_a name_b)))

(define (mother-of? name_a name_b)
  (and (equal? (get-property name_a 'gender) 'female)
       (parent-of? name_a name_b)))

(define (grandparent-of? name_a name_b)
  (member? name_a (grandparents-of name_b)))

(define (grandfather-of? name_a name_b)
  (and (equal? (get-property name_a 'gender) 'male)
       (grandparent-of? name_a name_b)))

(define (grandmother-of? name_a name_b)
  (and (equal? (get-property name_a 'gender) 'female)
       (grandparent-of? name_a name_b)))

(define (child-of? name_a name_b)
  (member? name_a (children-of name_b)))

(define (son-of? name_a name_b)
  (and (equal? (get-property name_a 'gender) 'male)
       (child-of? name_a name_b)))

(define (daughter-of? name_a name_b)
  (and (equal? (get-property name_a 'gender) 'female)
       (child-of? name_a name_b)))

(define (sibling-of? name_a name_b)
  (member? name_a (siblings-of name_b)))

(define (brother-of? name_a name_b)
  (and (equal? (get-property name_a 'gender) 'male)
       (sibling-of? name_a name_b)))

(define (sister-of? name_a name_b)
  (and (equal? (get-property name_a 'gender) 'female)
       (sibling-of? name_a name_b)))
\end{lstlisting}
\caption{Conceptual system and derived predicates for kinship trees.}
\label{code:kinship-conceptual-system}
\end{code}

\subsubsection{Translation examples for kinship}
\begin{code} \label{appendix-sec:kinship-translation}
\begin{lstlisting}[numbers=left,frame=lines]
;; -- CONDITION AND QUERY STATEMENTS --
;; Now, let's translate some user-defined statements.
;; Each statement begins with either `Condition` or `Query`.
;; `Condition` statements provide facts about the scenario.
;; `Query` statements are questions that evaluate quantities of interest.

;; Condition: Ryan's partner is Taylor.
(condition (partner-of? 'ryan 'taylor))

;; Condition: Taylor is the mother of Sam.
(condition (mother-of? 'taylor 'sam))

;; Condition: Sam's father is Ryan.
(condition (father-of? 'ryan 'sam))

;; Condition: Sam has two siblings.
(condition (= (length (siblings-of 'sam)) 2))

;; Condition: Sam has a brother.
(condition
  (exists (lambda (x)
    (brother-of? x 'sam))))

;; Condition: Payton's partner has a brother named Kyle.
(condition
  (exists (lambda (x) (and
                        (partner-of? x 'payton)
                        (brother-of? 'kyle x)))))

;; Condition: Payton's partner has a sister who has a son named Sam.
(condition
  (exists (lambda (x) (and
                        (partner-of? x 'payton)
                        (exists (lambda (y) (and
                                              (sister-of? y x)
                                              (son-of? 'sam y))))))))

;; Query: Who are Sam's parents?
(query (parents-of 'sam))

;; Query: How many children does Kyle have?
(query (length (children-of 'kyle)))

;; Query: Who is Ryan's grandfather?
(query
  (filter-tree
    (lambda (x) (grandfather-of? x 'ryan))))

;; Query: Does Taylor have a sister?
(query
  (exists (lambda (x)
    (sister-of? x 'taylor))))
\end{lstlisting}
\caption{Translation examples for kinship trees.}
\label{code:kinship-translation-examples}
\end{code}

\subsubsection{Why not Prolog?} \label{apx:why-not-prolog}

Readers who are familiar with the world of logic programming may wonder why we have chosen to model the kinship domain in Church instead of a more standard logic programming language, such as Prolog. Indeed, kinship is often one of the introductory examples in Prolog textbooks \citep{pereira2002prolog} and online tutorials,\footnote{\url{https://swish.swi-prolog.org/p/prolog-family-tree.pl}} from which we drew inspiration when writing this section. Moreover, there are many structural parallels between our framework and the style of declarative programming embodied by Prolog: \lstinline{scheme}{condition} statements in Church are similar to facts in Prolog; derived concepts like \lstinline{scheme}{father-of?} in our Church kinship model are analogous to Prolog rules; and \lstinline{scheme}{query} performs similar functions in both languages (though the algorithms that underlie these queries differ in important ways). And, as discussed in the introduction to \cref{sec:models-relational-reasoning}, Prolog was originally developed as a model of natural language \citep{colmerauer1972systeme} and has deep ties to computational linguistics. So: why not use Prolog?

In short, there is nothing about our approach to semantic parsing that precludes swapping out Church for other programming languages, like Prolog, SMT solvers, or even a general purpose language like Python. In fact, with the right prompting, Codex readily translates natural language utterances like \textit{Avery has a sister named Blake} into \lstinline{sister_of(blake, avery)} in Prolog. On the parsing side, we did not encounter any technical limitations to using LLMs to translate natural language into Prolog.

However, because Prolog is based on definite (Horn) clauses, there are limitations in the kinds of utterances that we can express and the kinds of inferences that we can make when working in Prolog. For instance, a typical Prolog kinship model might have a rule defining the concept of a ``grandfather'' as follows:

\begin{lstlisting}[numbers=none,frame=lines]
grandfather_of(X,Y) :- male(X),
    parent_of(X,Z),
    parent_of(Z,Y).
\end{lstlisting}

\noindent Now, if we learn that \textit{Charlie is the grandfather of Dana}, we might be inclined to translate this into Prolog as a fact: \lstinline{grandfather_of(charlie, dana)}. Given this information, we can make various deductive inferences: e.g., that Charlie is male, and that there exists some person in the family tree who is both the child of Charlie and the parent of Dana. In fact, this is how the  \lstinline{grandfather_of(X,Y)} rule is defined in the first place. 

For this reason, it is especially counterintuitive that these kinds of inferences are not at all straightforward in Prolog. Because logical implication in definite clauses is \textit{unidirectional}, anyone satisfying the right-hand side of the \lstinline{grandfather_of(X,Y)} rule is considered a grandfather. However, our rule says nothing about what being a grandfather entails. Moreover, our above translation \lstinline{grandfather_of(charlie, dana)} is actually quite facile; it simply modifies \lstinline{grandfather_of(X,Y)} such that queries will now return true for anyone satisfying the original definition; or for the special case where \lstinline{X=charlie} and \lstinline{Y=dana}. These are all examples of limitations of the kinds of \textit{deductive} inferences that we can model with Prolog. Additionally, there are many kinds of \textit{inductive} inferences that are not well-captured by Prolog; e.g., because Charlie has at least one child, he is more likely to have multiple children, and is more likely to be married. 

In sum, to get the kinds of mixed deductive and inductive inferences that we would like to see in an expressive language-of-thought, we need to have ways of incorporating and trading off uncertainty in our world model. ProbLog \citep{suster2021mapping, dries2017solving, problog}, a probabilistic extension of Prolog in which deduction rules can be annotated with probabilities, offers one way of integrating uncertainty with deductive reasoning. Church goes a step further by specifying a \textit{generative domain theory} in addition to probabilistic inference rules. We believe that this interplay between probabilistic priors and likelihoods---which is central to Bayesian inference---is also at the heart of human cognition.

\subsection{Perceptual and physical reasoning}
\subsubsection*{Static visual scenes}
\subsubsection{Generative world model for static visual scenes}\label{appendix-sec:models-static-scenes}
\begin{code}
        \begin{lstlisting}[numbers=left,frame=lines]
;; Objects have a shape attribute, which is a choice of cube, sphere, or cylinder shape categories.
(define choose-shape
  (mem (lambda (obj-id) 
         (pair 'shape (uniform-draw '(mug can bowl))))))

;; Objects have a color attribute that is drawn from a predefined set of RGB values.
(define choose-color
  (mem (lambda (obj-id) 
         (pair 'color (uniform-draw (list
                                      (list 255 0 0)
                                      (list 0 0 255)
                                      (list 0 255 0)
                                      (list 255 255 0)
                                    ))))))
;; An object is an object ID, and the object's attribute types and their values.
(define object (mem (lambda (obj-id) (list 
                                      (pair 'object-id obj-id) 
                                      (choose-shape obj-id) 
                                      (choose-color obj-id)))))

;; Scenes can have a maximum of 12 objects.
(define max-objects 12)
;; The number of objects in a scene tends to be not too large, and is capped at the maximum number of objects.
(define choose-num-objects
  (mem (lambda (scene-id) (floor (min max-objects (* max-objects (exponential 1)))))))

;; Then, for each object we intend to generate, generate an object indexical, and associate it with a choice of attributes.
(define obj-id-gensym (make-gensym "obj-"))
(define (generate-n-objects scene-id total-objects)
    (if (= total-objects 0)
     (list (object (obj-id-gensym)))                      
     (cons (object (obj-id-gensym)) (generate-n-objects scene-id (- total-objects 1)))))
(define objects-in-scene (mem (lambda (scene-id) (generate-n-objects scene-id (choose-num-objects scene-id)))))


;; An object is red if it is of this continuous color value.
(define red (list 255 0 0))
;; An object is blue if it is of this continuous color value.
(define blue (list 0 0 255))
;; An object is green if it is of this continuous color value.
(define green (list 0 255 0))
;; An object is yellow if it is of this continuous color value.
(define yellow (list 255 255 0))

;; Check if an object is of a given shape.
(define is-shape?  (lambda (shape) (lambda (object) (equal? (cdr (assoc 'shape object)) shape))))
;; Check if an object is of a given named color.
(define is-color?  (lambda (color) (lambda (object) (equal? (cdr (assoc 'color object)) color))))

;; Select only objects from the scene of a given color.
(define filter-color(lambda (color) (lambda (object-list) (filter (is-color? color) object-list))))

;; Select only objects from the scene of a given shape.
(define filter-shape (lambda (shape) (lambda (object-list) (filter (is-shape? shape) object-list))))
\end{lstlisting}
\caption{Generative domain theory for tabletop scenes. Generates scenes containing a set of objects which vary in shape and color. These scene states are rendered by a separately generated \lstinline{render} function to generate images. Shown with natural language comments, but these are not used in the LLM prompt.}
\label{code:domain-model-ref-grounding-static-scenes}
\end{code}

\subsubsection{Translation examples for static visual scenes}
\begin{code}
\begin{lstlisting}[numbers=left,frame=lines]
;; There's a blue thing.
(condition (> (length ((filter-color blue) (objects-in-scene 'this-scene))) 0))
;; There's at least two blue plates.
(condition  (>=  (length 
        ((filter-color blue)
        ((filter-shape 'plate) 
        (objects-in-scene 'scene)))) 
2))
;; There's many blue plates.
(condition  (>=  (length 
        ((filter-color blue)
        ((filter-shape 'plate) 
        (objects-in-scene 'scene)))) 
5))
;; There's exactly two plates and there's also a yellow thing.
(condition 
    (and (= (length ((filter-shape 'plate) (objects-in-scene 'scene))) 2) 
    (> (length ((filter-color yellow) (objects-in-scene 'scene))) 0)))  

;; Is there a mug?
(query (> (length ((filter-shape 'mug) (objects-in-scene 'this-scene))) 0))
\end{lstlisting}
\caption{Translation examples for the visual domain. These examples are concatenated with the visual scenes generative model to produce the prompt used to generate new translations.\\}
\label{code:ref-grounding-translation-examples}
\end{code}

\subsubsection*{Dynamic physical scenes}
\subsubsection{Generative world model for physical scenes}\label{appendix-sec:models-dynamic-scenes}
\begin{code}
        \begin{lstlisting}[numbers=left,frame=lines]
(define (get_attribute obj key)
    (if (assoc key obj) (rest (assoc key obj)) ()))

  (define (member? a b)
    (if (member a b) true false))
  (define concatenate
    (lambda (list-1 list-2)
      (if (null? list-1)
          list-2
          (cons (car list-1) (concatenate (cdr list-1) list-2)))))

(define (pairs x l)
  (define (aux accu x l)
    (if (null? l)
        accu
        (let ((y (car l))
              (tail (cdr l)))
          (aux (cons (cons x y) accu) x tail))))
  (aux '() x l))

(define (cartesian_product l m)   
  (define (aux accu l)
    (if (null? l) 
        accu
        (let ((x (car l)) 
              (tail (cdr l)))
          (aux (append (pairs x m) accu) tail))))
  (aux '() l))

;;;;;;;;;;;;;;;;;;;;;;;;;;;;;;;;;;;;;;;;;;;;;;;;;;;;;;;;;;;;;;;;;;;;;;;;;;;;;;;;;;;;;;;;;;;;;;;;;;;;;;;;;;;;;;;;;;;;;;;;;;;;;;;; Generative domain theory: dynamic scenes. Collision detection.
(define get_num_objects 2)
(define OBJECT_DEFAULT_RADIUS 1)
(define GRAVITY 9.8)
(define DELTA_T 0.5)

(define get_initial_color
     (lambda (obj_id)
     (if (eq? obj_id 'obj-0) 
         (list 255 0 0)
         (list 0 0 255))))
   
(define choose_mass
     (mem (lambda (obj_id) 
            (abs (gaussian 5 3)))))

(define choose_shapes
     (mem (lambda (scene-id) (uniform-draw (list 'sphere 'block)))))

(define min_x -3)
(define max_x 3)
(define mid_x (+ (/ (- max_x min_x) 2) min_x))
(define get_initial_x
     (lambda (obj_id)
     (if (eq? obj_id 'obj-0) 
         min_x
         mid_x)))
   
(define min_force 0)
   (define max_force 10)
   (define mid_force (+ (/ (- max_force min_force) 2) min_force))
   (define choose_initial_force
     (mem (lambda (obj_id)
            (if (eq? obj_id 'obj-0) 
            (abs (gaussian mid_force 3))
             0
                ))))

(define static_friction_constant (lambda (shape)
                                  (if (eq? shape 'sphere) 
         0.02
         0.05)
                                    ))
(define kinetic_friction_constant (lambda (shape)
                                  (if (eq? shape 'sphere) 
        0.01
        0.02)
                                    ))
(define normal_force (lambda (m) (* m GRAVITY)))
(define force_after_friction (lambda (f v shape m)
        (if (> (abs v) 0)
        (- f (* (kinetic_friction_constant shape) (normal_force m)))
        (if (< f (* (static_friction_constant shape) (normal_force m))) 0 (- f (* (kinetic_friction_constant shape) (normal_force m)))
         ))))

(define newtons_second (lambda (f m) (/ f m)))
(define v_next (lambda (v_prev a_prev delta_t)
                 (let ((v_temp (+ v_prev (* a_prev delta_t))))
                 (if (>= (* v_prev v_temp) 0) v_temp 0)) 
  ))
(define x_next (lambda (x_prev v_prev delta_t) (+ x_prev (* v_prev delta_t))))
(define initial_object_state (mem (lambda (obj_id scene_id) 
                                       (let ((obj_shape (choose_shapes scene_id)))
                                       (let ((obj_mass (choose_mass obj_id)))
                                         (let ((obj_color (get_initial_color obj_id)))
                                           (let ((initial_x (get_initial_x obj_id)))
                                             (let ((initial_push_force (choose_initial_force obj_id)))
                                             (let ((initial_force (force_after_friction initial_push_force 0 obj_shape obj_mass)))
                                               (list 
                                                (pair 'object_id obj_id)
                                                (pair 'object_radius OBJECT_DEFAULT_RADIUS)
                                                (pair 'shape obj_shape)
                                                (pair 'mass obj_mass)
                                                (pair 'color obj_color)
                                                (pair 'x initial_x)
                                                (pair 'initial_push_force initial_push_force)
                                                (pair 'f initial_force)
                                                (pair 't 0) 
                                                (pair 'a_prev (newtons_second initial_force obj_mass))
                                                (pair 'a (newtons_second initial_force obj_mass))
                                                (pair 'v_0 0)
                                                (pair 'v (v_next 0 (newtons_second initial_force obj_mass) DELTA_T))) 
                                               )))))))))
(define obj_id_gensym (make_gensym "obj-"))
(define generate_initial_state 
     (mem (lambda (scene_id total_objects)
            (if (= total_objects 1)
                (list (initial_object_state (obj_id_gensym) scene_id))                      
                (cons (initial_object_state (obj_id_gensym) scene_id) (generate_initial_state scene_id (- total_objects 1)))))))

(define generate_initial_scene_event_state (mem (lambda (scene_id total_objects)
                                                             (pair 0
                                                                   (list 
                                                                    (pair 'scene_states (generate_initial_state scene_id total_objects))
                                                                    (pair 'event_states [])
                                                                    ))
                                                             )
))

(define event_id_gensym (make_gensym "event-"))
(define circle_intersect? (lambda (subject_x subject_radius object_x object_radius)
(let ((square_circle_distance (expt (- subject_x object_x) 2)))
(let ((square_radii (expt (+ subject_radius object_radius) 2)))
(leq square_circle_distance square_radii)))          
))
(define elastic_collision_subject_v (lambda (subject_m subject_v object_m object_v)
      (/ (+ (* 2 (* object_m object_v)) (* subject_v (- subject_m object_m))) (+ subject_m object_m))
))

(define get_collision_events (lambda (time scene_event_state_for_time)
  (let ((scene_event_state (get_attribute scene_event_state_for_time time)))
  (let ((scene_state (get_attribute scene_event_state 'scene_states)))
  (if (= (length scene_state) 1)
      ()
  (fold (lambda (event events) (if (equal? event ()) events (cons event events)))  () 
  (let ((paired_object_states (cartesian_product scene_state scene_state)))
  (map (lambda (paired_objects)
         
  (let ((event_subject (get_attribute (first paired_objects) 'object_id)))
  (let ((event_object (get_attribute (cdr paired_objects) 'object_id)))
  (if (eq? event_subject event_object) ()
  (let ((subject_v (get_attribute (first paired_objects) 'v)))
  (let ((subject_x (get_attribute (first paired_objects) 'x)))
  (let ((subject_m (get_attribute (first paired_objects) 'mass)))
  (let ((subject_radius (get_attribute (first paired_objects) 'object_radius)))
  (let ((object_v (get_attribute (cdr paired_objects) 'v)))
  (let ((object_x (get_attribute (cdr paired_objects) 'x)))
  (let ((object_m (get_attribute (cdr paired_objects) 'mass)))
  (let ((object_radius (get_attribute (cdr paired_objects) 'object_radius)))
  (if (circle_intersect? subject_x subject_radius object_x object_radius)
      (list
                  (pair 'event-id  (event_id_gensym))
                  (pair 'event_time  time)
                  (pair 'event_predicates (list 'is_colliding))
                  (pair 'event_subject event_subject)
                  (pair 'event_object event_object)
                  (pair 'subject_initial_v subject_v)
                  (pair 'subject_final_v (elastic_collision_subject_v subject_m subject_v object_m object_v))
                  (pair 'object_initial_v object_v)
                  )
   ()))))))))))  
  )))
  paired_object_states)))
)))))


(define generate_next_object_state (lambda (current_time event_state) (lambda (prev_object_state)
  (let ((obj_id (cdr (assoc 'object_id prev_object_state))))
  (let ((collision_events (fold (lambda (event events) (if (equal? (get_attribute event 'event_subject) obj_id) (cons event events) events)) () event_state)))
  (if (> (length collision_events) 0)
  (generate_collision_event_state current_time obj_id prev_object_state (car collision_events))
  (generate_no_collision_event_state current_time obj_id prev_object_state)  
  )
  )))))

(define generate_collision_event_state (lambda (current_time obj_id prev_object_state collision_event)
  (let ((obj_radius (cdr (assoc 'object_radius prev_object_state))))
      (let ((obj_mass (cdr (assoc 'mass prev_object_state))))
        (let ((obj_color (cdr (assoc 'color prev_object_state))))
        (let ((obj_shape (cdr (assoc 'shape prev_object_state))))
          (let ((v_prev (cdr (assoc 'v prev_object_state))))
            (let ((a_prev (cdr (assoc 'a_prev prev_object_state))))
              (let ((x_prev (cdr (assoc 'x prev_object_state))))
                (let ((v (get_attribute collision_event 'subject_final_v)))
                  (let ((x (x_next x_prev v 1)))
                    (list 
                    (pair 'object_id obj_id)
                    (pair 'object_radius obj_radius)
                    (pair 'shape obj_shape)
                    (pair 'color obj_color)
                    (pair 'mass obj_mass)
                    (pair 'x x)
                    (pair 'f 0) 
                    (pair 't (* current_time DELTA_T)) 
                    (pair 'a_prev 0)
                    (pair 'a 0) 
                    (pair 'v_0 0)
                    (pair 'v v))
                    )))))
          ))))
))

(define generate_no_collision_event_state (lambda (current_time obj_id prev_object_state)
  (let ((obj_radius (cdr (assoc 'object_radius prev_object_state))))
      (let ((obj_mass (cdr (assoc 'mass prev_object_state))))
        (let ((obj_color (cdr (assoc 'color prev_object_state))))
        (let ((obj_shape (cdr (assoc 'shape prev_object_state))))
          (let ((v_prev (cdr (assoc 'v prev_object_state))))
            (let ((a_prev_no_friction (cdr (assoc 'a_prev prev_object_state))))
            (let ((a_prev (newtons_second (force_after_friction 0 v_prev obj_shape obj_mass) obj_mass)))
              (let ((x_prev (cdr (assoc 'x prev_object_state))))
                (let ((v (v_next v_prev a_prev DELTA_T)))
                  (let ((x (x_next x_prev v_prev DELTA_T)))
                    (list 
                    (pair 'object_id obj_id)
                    (pair 'object_radius obj_radius)
                    (pair 'shape obj_shape)
                    (pair 'color obj_color)
                    (pair 'mass obj_mass)
                    (pair 'x x)
                    (pair 'f (force_after_friction 0 v_prev obj_shape obj_mass)) 
                    (pair 't (* current_time DELTA_T)) 
                    (pair 'a_prev a_prev)
                    (pair 'a 0) 
                    (pair 'v_0 0)
                    (pair 'v v))
                    )))))
          ))))
)))

(define generate_next_scene_state (lambda (prev_scene_state event_state next_time)
        (map (generate_next_object_state next_time event_state) prev_scene_state)))

(define generate_next_scene_event_state_time (lambda (next_time scene_event_state_for_times)
        (let ((prev_scene_event_state (get_attribute scene_event_state_for_times (- next_time 1))))
        (let ((prev_scene_state (get_attribute prev_scene_event_state 'scene_states)))
        (let ((event_state (get_collision_events (- next_time 1) scene_event_state_for_times)))

        (pair next_time (list
           (pair 'scene_states (generate_next_scene_state prev_scene_state event_state next_time))
           (pair 'event_states event_state)
        ))
)))))

(define generate_next_scene_event_states
     (lambda (current_time prev_scene_event_states_for_times)
     (cons (generate_next_scene_event_state_time current_time prev_scene_event_states_for_times) prev_scene_event_states_for_times)
))

(define generate_scene_event_states_for_times (mem (lambda (scene_id total_objects total_time)
                                                        (if (= total_time 0)
                                                            (list 
                                                             (generate_initial_scene_event_state scene_id total_objects)
                                                             )
                                                            (let ((prev_scene_event_states (generate_scene_event_states_for_times scene_id total_objects (- total_time 1)))) 
                                                              (generate_next_scene_event_states total_time prev_scene_event_states)
)))))

(define max_time 9)

(define base_states_for_times  (generate_scene_event_states_for_times 'this_scene get_num_objects max_time))

;;;;;;;;;;;;;;;;;;;;;;;;;;Derived predicates.
(define objects_in_scene (lambda (base_states_for_times) 
        (let ((initial_base_states_at_time (cdr (assoc 0 (cdr base_states_for_times)))))
        (let ((base_state (cdr (assoc 'scene_states initial_base_states_at_time))))
        base_state
      ))
  ))
(define red (list 255 0 0))
(define blue (list 0 0 255))
(define is_color?  (lambda (color) (lambda (object) (equal? (cdr (assoc 'color object)) color))))
(define is_shape?  (lambda (shape) (lambda (object) (equal? (cdr (assoc 'shape object)) shape))))

(define all_objects  (objects_in_scene base_states_for_times))
(define (exists_object predicate)
  (some (map predicate (objects_in_scene base_states_for_times))))

(define (filter_objects predicate)
  (map 
  (lambda (o) (get_attribute o 'object_id))
  (filter predicate (objects_in_scene base_states_for_times))))
;;;;;;;;;;;;;;;;;;;;;;;;;;;;;;;;;;;;;;;;;;;;;;;;;;;;;;;;;;;;;;;;   
(define QUICKLY_THRESHOLD 2)
(define SLOWLY_THRESHOLD 2)

  (define is_moving_events (mem (lambda (base_states_for_times) 
          (fold (lambda (base_state_for_time these_events)
          (let ((current_time (car base_state_for_time)))
          (let ((base_state (cdr (assoc 'scene_states (cdr base_state_for_time)))))
          (fold (lambda (obj_state these_events)
            (let ((obj_id (cdr (assoc 'object_id obj_state))))
            (let ((obj_velocity (cdr (assoc 'v obj_state))))
            (let ((obj_speed (abs obj_velocity)))
            (if (> obj_speed 0)
              ;;
                (let ((event_predicates 
                      (if (> obj_speed QUICKLY_THRESHOLD)
                          (list 'is_moving 'is_quickly)
                          (if (< obj_speed SLOWLY_THRESHOLD)
                              (list 'is_moving 'is_slowly)
                              (list 'is_moving)
                              ))
                       ))
                (cons 
                  (list
                  (pair 'event-id  (event_id_gensym))
                  (pair 'event_time  current_time)
                  (pair 'event_predicates event_predicates)
                  (pair 'event_subject obj_id)
                   (pair 'event_speed obj_speed)
                  )
                these_events)) 
                these_events
                
                )))))
                these_events base_state))))
() base_states_for_times))))

(define is_resting_events (mem (lambda (base_states_for_times) 
          (fold (lambda (base_state_for_time these_events)
          (let ((current_time (car base_state_for_time)))
          (let ((base_state (cdr (assoc 'scene_states (cdr base_state_for_time)))))
          (fold (lambda (obj_state these_events)
            (let ((obj_id (cdr (assoc 'object_id obj_state))))
            (let ((obj_velocity (cdr (assoc 'v obj_state))))
            (let ((obj_speed (abs obj_velocity)))
            (if (= obj_speed 0)
              ;;
                (let ((event_predicates 
                      (list 'is_resting)))
                (cons 
                  (list
                  (pair 'event-id  (event_id_gensym))
                  (pair 'event_time  current_time)
                  (pair 'event_predicates event_predicates)
                  (pair 'event_subject obj_id)
                   (pair 'event_speed obj_speed)
                  )
                these_events)) 
                these_events
                
                )))))
                these_events base_state))))
          () base_states_for_times))))

(define is_colliding_events (mem (lambda (base_states_for_times)
        (fold (lambda (base_state_for_time these_events)
        (let ((current_time (car base_state_for_time)))
        (let ((event_states (cdr (assoc 'event_states (cdr base_state_for_time)))))
        (fold (lambda (event_state these_events)
                (let ((subject_initial_speed (abs (get_attribute event_state 'subject_initial_v))))
                (let ((subject_final_speed (abs (get_attribute event_state 'subject_final_v))))
                (let ((object_initial_speed (abs (get_attribute event_state 'object_initial_v)))) 
                (let ((cause_subject_object_event (and (> subject_initial_speed 0) (= object_initial_speed 0))))
                (let 
                ((event_predicates 
                      (if (and cause_subject_object_event (eq? subject_final_speed 0))
                          (list 'is_launching 'is_hitting 'is_colliding)
                          (if (> subject_initial_speed 0)
                              (list 'is_hitting 'is_colliding)
                              (list 'is_colliding)
                              )
                       )))
                
                (cons (list
                  (pair 'event-id  (get_attribute event_state 'event-id))
                  (pair 'event_time (get_attribute event_state 'event_time))
                  (pair 'event_predicates event_predicates)
                  (pair 'event_subject (get_attribute event_state 'event_subject))
                  (pair 'event_object (get_attribute event_state 'event_object))
                  (pair 'subject_initial_v (get_attribute event_state 'subject_initial_v ))
                  (pair 'subject_final_v (get_attribute event_state 'subject_final_v ))
                  (pair 'object_initial_v (get_attribute event_state 'object_initial_v ))
                  ) these_events))))))
                ) these_events event_states)      
         )))
        () base_states_for_times)
                                   
))) 
                                   


(define events_in_scene (concatenate
                                  (is_colliding_events base_states_for_times)
                                  (concatenate 
                                  (is_moving_events base_states_for_times)
                                  (is_resting_events base_states_for_times))))


(define is_event? (lambda (event_predicate event) (member? event_predicate (get_attribute event 'event_predicates))))

(define is_subject_of_event? (lambda (event object ) (equal? 
    (get_attribute event 'event_subject)
    (get_attribute object 'object_id)
  )))

(define is_object_of_event? (lambda (event object ) (equal? 
    (get_attribute event 'event_object)
    (get_attribute object 'object_id)
  )))
   
(define event_subject_is? (lambda (event predicate) (member? 
    (get_attribute event 'event_subject)
    (filter_objects predicate)
  )))
(define event_object_is? (lambda (event predicate) (member? 
    (get_attribute event 'event_object)
    (filter_objects predicate)
  )))

(define (exists_event predicate)
    (some (map predicate events_in_scene)))

(define (filter_events predicate)
    (filter predicate events_in_scene))
\end{lstlisting}
\caption{Generative domain theory for physical scenes. Generates scenes containing a red object left of a blue object, and a randomly generated force. These scene states are forward simulated using a physics engine which is shown implemented within this Church code. Shown with natural language comments, but these are not used in the LLM prompt.}
\label{code:domain-model-ref-grounding-physics-scenes}
\end{code}

\subsubsection{Translation examples for visual scenes}
\begin{code}
\begin{lstlisting}[numbers=left,frame=lines]
;; The objects are all balls.
(condition (all (map (lambda (o) ((is_shape? 'sphere) o)) all_objects)))
;; Everything is a ball.
(condition (all (map (lambda (o) ((is_shape? 'sphere) o)) all_objects)))
;; Imagine the red thing is a block, and is somewhat heavy.
(condition (exists_object (lambda (object)
            (and
            ((is_color? red) object)
            ((is_shape? 'cube) object) 
            (> (get_attribute object 'mass) 2)
            ))))
;; There is a blue ball, and it is quite heavy.
(condition (exists_object (lambda (object)
            (and
            ((is_color? blue) object)
            ((is_shape? 'sphere) object) 
            (> (get_attribute object 'mass) 3.5)
            ))))
;; Now, the red block is very light.
(condition (exists_object (lambda (object)
            (and
            ((is_color? red) object)
            ((is_shape? 'cube) object) 
            (< (get_attribute object 'mass) 1)
            ))))
;; A blue ball is somewhat light.
(condition (exists_object (lambda (object)
            (and
            ((is_color? red) object)
            ((is_shape? 'cube) object) 
            (< (get_attribute object 'mass) 2)
            ))))
;; Imagine the red block gets pushed lightly to the right.
(condition (exists_object (lambda (object)
            (and
            ((is_color? red) object)
            ((is_shape? 'cube) object) 
            (< (get_attribute object 'initial_push_force) 2)
            ))))
;; Now, imagine a red ball is pushed hard to the right.
(condition (exists_object (lambda (object)
            (and
            ((is_color? red) object)
            ((is_shape? 'sphere) object) 
            (> (get_attribute object 'initial_push_force) 6)
            ))))  
;; A red block hits a blue block.
(condition 
(exists_object (lambda (object_1)
(exists_object (lambda (object_2)                 
(exists_event (lambda (event)
                    (and 
                        ((is_color? red) object_1)  
                        ((is_shape? 'cube) object_1)
                        ((is_color? blue) object_2)  
                        ((is_shape? 'cube) object_2)
                        (is_subject_of_event? event object_1)
                        (is_object_of_event? event object_2)
                        (is_event? 'is_hitting event))
                        )))))))
;; What's the final velocity of the red block after it is hit?
(query (last (map 
(lambda (event) (get_attribute event 'subject_final_v))
(filter_events
(lambda (e)
(and
(is_event? 'is_colliding e)
(event_subject_is? e (lambda (o)
                    (and 
                     ((is_color? red) o) 
                     ((is_shape? 'cube) o))))))))))
\end{lstlisting}
\caption{Translation examples for the physics domain. These examples are concatenated with the physical scenes generative model to produce the prompt used to generate new translations.\\}
\label{code:ref-grounding-translation-examples-physics}
\end{code}

\subsection{Social reasoning}
\subsubsection{Generative world model for social reasoning}\label{appendix-sec:agents}

\begin{code}
        \begin{lstlisting}[numbers=left,frame=lines]
(define gridworld (list 
     (list 'ames 'lawn 'lawn 'lawn 'sushi)
     (list 'ames 'lawn 'lawn 'lawn 'danner)
     (list 'office 'barlow 'barlow 'barlow 'danner)
     (list 'ames 'lawn 'lawn 'lawn 'danner)             
     (list 'ames 'lawn 'lawn 'lawn 'vegetarian)
     (list 'pizza 'carson 'carson 'carson 'danner)                    
))
(define restaurants (list 'sushi 'pizza 'vegetarian))

(define initial_x 1)
(define initial_y 3)

   
(define has_bike (mem (lambda (agent-id) (flip))))
(define available_motions (mem (lambda (agent-id) (if (has_bike agent-id) (list 'is_walking 'is_biking) (list 'is_walking)))))
(define directions (list 'west 'east 'north 'south))
(define available_actions (mem (lambda (agent-id) (cons (pair 'stay 'stay) (cartesian_product (available_motions agent-id) directions)))))

(define is_open (mem (lambda (restaurant_type) (flip))))
(define POSITIVE_UTILITY_MEAN 10)
(define NEGATIVE_UTILITY_MEAN -10)
(define UTILITY_VARIANCE 1)
(define restaurant_utility (mem (lambda (agent-id restaurant_type)
                           (uniform-draw 
                           (list 
                           (gaussian POSITIVE_UTILITY_MEAN UTILITY_VARIANCE)
                           (gaussian NEGATIVE_UTILITY_MEAN UTILITY_VARIANCE)
)))))

(define motion_utility (mem (lambda (agent-id location_type motion_type)
  (case location_type
      (('lawn) (case motion_type
                (('is_biking) -1)
                (('is_walking) -0.2)
                (('is_staying) 0)
                (else 0))
                )
      (else (case motion_type
                (('is_biking) -0.01)
                (('is_walking) -0.2)
                (('is_staying) 0)
                (else 0)))
))))

(define food_utility (mem (lambda (agent-id location_type)
  (case location_type
      (('lawn) 0)
      (('ames) 0)
      (('barlow) 0)
      (('carson) 0)
      (('danner) 0)
      (('office) 0)
      (else 
       (if (is_open location_type) (restaurant_utility agent-id location_type) NEGATIVE_UTILITY_MEAN))
))))
 
(define utility_function (mem (lambda (agent-id gridworld state_x state_y action) 
        (let ((location_type (get_gridworld_at gridworld state_x state_y)))
        (let ((motion_type (car action)))
        (let ((state_food_utility (food_utility agent-id location_type)))
        (let ((state_motion_utility (motion_utility agent-id location_type motion_type)))
        (+ state_food_utility state_motion_utility))))))))

(define get_gridworld_at (lambda (gridworld x y)
   (list-elt (list-elt gridworld y) x)                   
))
(define x_increment (lambda (direction)
  (case direction
      (('west) -1)
      (('east) 1)
      (('north) 0)
      (('south) 0)
      (('stay) 0)
)))
(define y_increment (lambda (direction)
  (case direction
      (('north) -1)
      (('south) 1)
      (('west) 0)
      (('east) 0)
      (('stay) 0)
)))
(define gridworld_max_x (lambda (gridworld) (length (list-elt gridworld 1))))
(define gridworld_max_y (lambda (gridworld) (length gridworld)))
(define gridworld_transition (lambda (gridworld current_x current_y action)
   (let ((direction (cdr action)))         
   (let ((next_x (if (>= current_x (gridworld_max_x gridworld)) current_x (+ (x_increment direction) current_x))))
   (let ((next_x (if (< next_x 1) current_x next_x)))
   (let ((next_y (if (>= current_y (gridworld_max_y gridworld)) current_y (+ (y_increment direction) current_y))))
   (let ((next_y (if (< next_y 1) current_y next_y)))
   (let ((next_state (get_gridworld_at gridworld next_x next_y)))
   (list next_state next_x next_y)
))))))))

(define value_function (mem (lambda (agent-id curr_iteration gridworld state_x state_y)
   (if (equal? curr_iteration -1) 0 
   (let ((prev_optimal_action_value (optimal_action_value agent-id (- curr_iteration 1) gridworld state_x state_y)))
   (cdr prev_optimal_action_value))       
))))

(define available_actions_to_values (mem (lambda (agent-id curr_iteration gridworld state_x state_y)
       (map (lambda (action)
              (let ((utility (utility_function agent-id gridworld state_x state_y action)))
              (let ((next_state (gridworld_transition gridworld state_x state_y action)))
              (let ((next_state_x (second next_state)))
              (let ((next_state_y (third next_state)))
              (let ((next_state_value (value_function agent-id curr_iteration gridworld next_state_x next_state_y)))
              (pair action (+ utility next_state_value))
        )))))) 
        (available_actions agent-id))
)))

(define optimal_action_value (mem (lambda (agent-id curr_iteration gridworld state_x state_y)
      (let ((actions_to_values (available_actions_to_values agent-id curr_iteration gridworld state_x state_y)))
      (max_cdr actions_to_values)
      )
)))

(define MAX_ITERATIONS 20)
(define should_terminate (mem (lambda (agent-id gridworld state_x state_y)
        (if (<= (value_function agent-id MAX_ITERATIONS gridworld initial_x initial_y) 0) true                        
        (let ((location_type (get_gridworld_at gridworld state_x state_y)))
        (let ((state_food_utility (food_utility agent-id location_type)))
                           (> state_food_utility 0)))))))



(define optimal_policy_from_initial_state (mem (lambda (agent-id gridworld state_x state_y)
     (if (should_terminate agent-id gridworld state_x state_y) ()
     (let ((curr_optimal_action_value (optimal_action_value agent-id MAX_ITERATIONS gridworld state_x state_y)))
     (let ((curr_optimal_action (car curr_optimal_action_value)))
     (let ((next_state (gridworld_transition gridworld state_x state_y curr_optimal_action)))
     (let ((next_state_x (second next_state)))
     (let ((next_state_y (third next_state)))
     (let ((remaining_policy (optimal_policy_from_initial_state agent-id gridworld next_state_x next_state_y)))
     (cons curr_optimal_action remaining_policy)
))))))))))

(define trajectory_from_initial_state (mem (lambda (agent-id gridworld state_x state_y)
     (if (should_terminate agent-id gridworld state_x state_y) ()
     (let ((curr_optimal_action_value (optimal_action_value agent-id MAX_ITERATIONS gridworld state_x state_y)))
     (let ((curr_optimal_action (car curr_optimal_action_value)))
     (let ((next_state (gridworld_transition gridworld state_x state_y curr_optimal_action)))
     (let ((next_state_location (first next_state)))
     (let ((next_state_x (second next_state)))
     (let ((next_state_y (third next_state)))
     (let ((remaining_trajectory (trajectory_from_initial_state agent-id gridworld next_state_x next_state_y)))
     (cons next_state_location remaining_trajectory))
))))))))))

(define optimal_policy (mem (lambda (agent-id gridworld initial_state_x initial_state_y)
        (cons (pair 'start 'start) (optimal_policy_from_initial_state agent-id gridworld initial_state_x initial_state_y)))))

(define optimal_trajectory (mem (lambda (agent-id gridworld initial_state_x initial_state_y)
        (cons (get_gridworld_at gridworld initial_state_x initial_state_y) (trajectory_from_initial_state agent-id gridworld initial_state_x initial_state_y))                       
)))

(define optimal_policy_with_trajectory (mem (lambda (agent-id gridworld initial_state_x initial_state_y)
        (zip (optimal_policy agent-id gridworld initial_state_x initial_state_y) (optimal_trajectory agent-id gridworld initial_state_x initial_state_y))                                
)))

(define get_terminal_goal_state (mem (lambda (agent-id gridworld initial_state_x initial_state_y)
        (last (optimal_trajectory agent-id gridworld initial_state_x initial_state_y)))))

(define trajectory_has_location_type? (mem (lambda (agent-id location_type gridworld initial_state_x initial_state_y)
        (member? location_type (optimal_trajectory agent-id gridworld initial_state_x initial_state_y)) 
)))
(define policy_has_motion_type? (mem (lambda (agent-id motion_type gridworld initial_state_x initial_state_y)
      (let ((policy_motions (map (lambda (action) (first action)) (optimal_policy agent-id gridworld initial_state_x initial_state_y))))
      (member? motion_type policy_motions) 
))))
(define policy_and_trajectory_has_motion_at_location? (mem (lambda (agent-id motion_type location_type gridworld initial_state_x initial_state_y)
      (let ((policy_motions (map (lambda (action) (first action)) (optimal_policy agent-id gridworld initial_state_x initial_state_y))))
      (let ((trajectory (optimal_trajectory agent-id gridworld initial_state_x initial_state_y)))
      (let ((motions_at_locations (zip policy_motions trajectory)))
      (member? (list motion_type location_type) motions_at_locations) 
))))))

(define motion_at_location? (mem (lambda (agent-id motion_type location_type gridworld initial_state_x initial_state_y)
      (let ((policy_motions (map (lambda (action) (first action)) (optimal_policy agent-id gridworld initial_state_x initial_state_y))))
      (let ((trajectory (optimal_trajectory agent-id gridworld initial_state_x initial_state_y)))
      (let ((motions_at_locations (zip policy_motions trajectory)))
      motions_at_locations
))))))
;;;;;;;;;;;;;;;;;;;;;;;;;;;;;;;;;;;;;;;;;;;;;;;;;;;;;;;;;;;;;;;;;;;;;;;;;;;;;;;;;;;;;;;;;;;;;;;;;;;;
;; Derived predicates.
;;;;;;;;;;;;;;;;;;;;;;;;;;;;;;;;;;;;;;;;;;;;;;;;;;;;;;;;;;;;;;;;;;;;;;;;;;;;;;;;;;;;;;;;;;;;;;;;;;;;
(define action_id_gensym (make_gensym "action-"))
(define is_going_to_actions (mem (lambda (agent-id)
        (let ((action_states (optimal_policy_with_trajectory agent-id gridworld initial_x initial_y)))
        (let ((final_location (last (last action_states))))
        (list (list
                  (pair 'action_id  (action_id_gensym))
                  (pair 'action_subject agent-id)
                  (pair 'action_predicates (list 'is_going (list 'to final_location)))
                  (pair 'action_preposition 'to)
                  (pair 'action_location final_location)
         )))))))

(define is_going_on_actions (mem (lambda (agent-id)
        (let ((action_states (optimal_policy_with_trajectory agent-id gridworld initial_x initial_y)))
        (fold (lambda (action_state these_actions)
        (let ((action_location (last action_state)))
        (let ((action_manner (first (first action_state))))
        (let ((action_direction (cdr (first action_state))))
        (cons 
        (list
                  (pair 'action_id  (action_id_gensym))
                  (pair 'action_subject agent-id)
                  (pair 'action_predicates (list 'is_going action_manner action_direction (list 'on action_location)))
                  (pair 'action_preposition 'on)
                  (pair 'action_location action_location)
         )
        these_actions)
                ))))
        () action_states) 
))))

(define actions_in_scene (mem (lambda (agent-id) (concatenate (is_going_to_actions agent-id) (is_going_on_actions agent-id)))))
(define is_action? (lambda (action action_predicate) (member? action_predicate (lookup action 'action_predicates))))
(define is_subject_of_action? (lambda (action entity) (eq? 
    (lookup action 'action_subject)
    entity
  )))

(define is_preposition_of_action? (lambda (action preposition) (eq? 
    (lookup action 'action_preposition)
    preposition
  )))
(define is_location_of_action? (lambda (action location) (eq? 
    (lookup action 'action_location)
    location
  )))

(define get_location (lambda (action) 
    (lookup action 'action_location)
  ))

(define (exists_action agent-id predicate)
    (some (map predicate (actions_in_scene agent-id))))

(define (get_actions agent-id predicate)
    (fold (lambda (action these_actions) (if (predicate action) (cons action these_actions) these_actions)) 
          () (actions_in_scene agent-id))
)
\end{lstlisting}
\caption{Generative domain theory for restaurant navigation domain. Generates agents with varying preferences in a gridworld environment. Also implements a value iteration-based planner directly in the Church code.}
\label{code:domain-model-agents}
\end{code}

\subsubsection{Translation examples for social reasoning domain}
\begin{code}
\begin{lstlisting}[numbers=left,frame=lines]
;; Bob likes pizza.
(condition (> (restaurant_utility 'bob 'pizza) 0))
;; Bob really likes pizza.
(condition (> (restaurant_utility 'bob 'pizza) 10))
;; Bob does not like pizza, and he actually despises vegetables.
(condition (and 
    (< (restaurant_utility 'bob 'pizza) 0)
     (< (restaurant_utility 'bob 'vegetarian) 10)
))
;; The pizza place is not open.
(condition (not (is_open 'pizza)))
;; Condition on: Bob walked North on Danner.
(condition (exists_action 'bob (lambda (action)
                      (and 
                      (is_subject_of_action? action 'bob)
                      (is_action? action 'is_walking)
                      (is_action? action 'north)          
                      (is_preposition_of_action? action 'on)
                      (is_location_of_action? action 'danner))))) 
;; Does Bob like vegetarian food?
(query (> (restaurant_utility 'bob 'vegetarian) 0))
;; Where is Bob going?
(query (get_actions 'bob (lambda (action)
   (and
  (is_subject_of_action? action 'bob)
  (is_action? action 'is_going)))))
;; Where will Bob go to for lunch?
(query (get_location (first 
         (get_actions 'bob (lambda (action)
                      (and (and
                      (is_subject_of_action? action 'bob)
                      (is_action? action 'is_going))            
                      (is_preposition_of_action? action 'to))
                       )))))
\end{lstlisting}
\caption{Translation examples for the social reasoning. These examples are concatenated with the social reasoning scenes generative model to produce the prompt used to generate new translations.\\}
\label{code:ref-agents-translation-examples}
\end{code}

\clearpage
\section{Open questions}
\subsection{Syntactic bootstrapping}
\begin{figure}[h!]
    \centering
    \includegraphics[width=\textwidth]{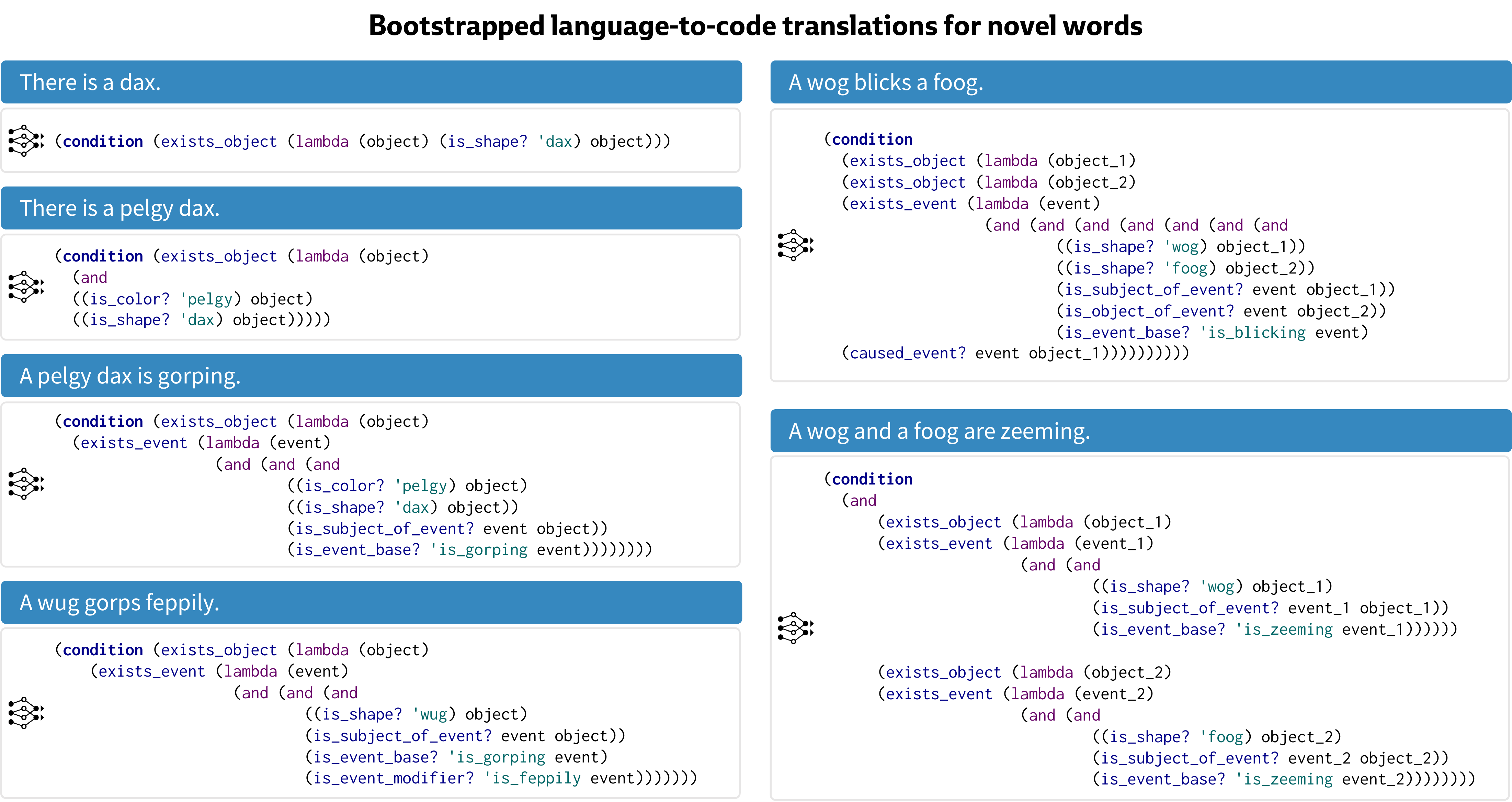}
    \caption{Example translations with novel words, suggesting that language-to-code models can leverage syntax-semantic mappings to inform hypothesized meanings.}
    \label{appendix:open-questions:fig:bootstrapping-translations}
\end{figure}

\subsection{Code editing} \label{appendix:open-questions:code-editing}
While our framework focuses primarily on \textit{generating} code in the PLoT, this view encompasses only part of the broader story of natural language. In particular, in certain  contexts, it might not make sense to write new code, but instead to \textit{modify} the existing domain theory. Consider the following statements, taken in context of the domains we explored in \cref{sec:language-and-world-models}:

\begin{itemize}[noitemsep]
    \item (Tug-of-war) \textit{The team's strength is the strength of the \textbf{strongest} player.}
    \item (Kinship) \textit{Avery has two kids \textbf{from a previous marriage}.}
    \item (Visual scenes) \textit{There's a red mug \textbf{stacked on top of} a yellow can.}
    \item (Navigation) \textit{\textbf{There's a river} separating the North and South sides of town, which \textbf{you can paddle across} in nice weather.}
\end{itemize}

\noindent These utterances bend or break the rules of their respective domain theories. To properly integrate these kinds of language, we'd need to edit pieces of the existing generative models.

While language-guided code editing is still an open area of research, recent advances offer an exciting glimpse of what might be possible in the near-term. \cite{ouyang2022instructgpt} use a combination of finetuning and reinforcement learning to make GPT-3 adhere more closely to human-authored instructions. The resulting InstructGPT models, which OpenAI make available on their API, are capable of editing existing text based on short natural language instructions (e.g., ``Fix the grammar''; ``Turn this into a poem.'').\footnote{\url{https://openai.com/blog/gpt-3-edit-insert/}} Excitingly, this same approach extends to code-based LLMs, meaning that it is possible to prompt GPT models to edit a piece of code according to some instructions. Indeed, we can use OpenAI's editing interface off-the-shelf to handle utterances requiring localized changes to the domain model (see below for a simple example in the tug-of-war domain).

\begin{Dialogue}[colorRedefine]
\Redefine{The team's strength is the strength of the strongest player.}

\begin{minipage}[]{\linewidth}
    \footnotesize
    \begin{minipage}[]{0.48\linewidth}
\begin{lstlisting}[numbers=left]
;; The team's strength is the sum of the players' strengths.
;; When a player is lazy in a match, they pull with half their strength.
(define (team-strength team)
  (sum 
    (map
      (lambda (player) 
        (if (flip (laziness player)) 
            (/ (strength player) 2) 
            (strength player)))
      team)))
\end{lstlisting}
    \end{minipage}
    \hfill
    \begin{minipage}[]{0.48\linewidth}
\begin{lstlisting}[numbers=left]
;; The team's strength is the strength of the strongest player.
;; When a player is lazy in a match, they pull with half their strength.
(define (team-strength team)
  (apply 
    max
      (lambda (player) 
        (if (flip (laziness player)) 
            (/ (strength player) 2) 
            (strength player)))
      team)))
\end{lstlisting}
\end{minipage}
\end{minipage}
\end{Dialogue}

Though questions of scaling and robustness remain, the problem of modeling sequences of code changes is currently gaining traction in the machine learning for code community, which has recently produced multiple language-guided neural models of code editing \citep{Fried2022InCoderAG, Reid2022LearningTM, Chakraborty2022CODITCE, Zhang2022CoditT5PF, Chakraborty2021OnML, panthaplackel2020learning} that draw broadly on contemporary work in automated program repair \citep{Yasunaga2020DrRepair, Li2020DLFixCC, Bai2021JointlyLT}. These advances suggest a broader vision for our framework in which domain theories, expressed in the PLoT, can be iteratively grown and revised to reflect natural language instruction. Moreover, as code LLMs become more general-purpose, the technical gap between generation and editing will continue to narrow, suggesting a point in the near future where defining new components of a domain theory will be a special case of language-guided code editing.

\section{Attributions}
\subsection{Attribution of graphics resources}

\raisebox{-0.75\fontcharht\font`\B}{\includegraphics[height=2.5\fontcharht\font`\B]{figures/nn_icon.png}} 
Artificial neural network icon by \texttt{sachin modgekar} from \url{thenounproject.com}.\\

\noindent\raisebox{-0.75\fontcharht\font`\B}{\includegraphics[height=2.5\fontcharht\font`\B]{figures/cogs_icon.png}} 
Cog icon by \texttt{Rflor} from \url{thenounproject.com}.

\end{document}